\documentclass{article} % For LaTeX2e
\usepackage{microtype}
\usepackage{graphicx}
\usepackage{subfigure}
\usepackage{booktabs} % for professional tables
\usepackage{amssymb}
\usepackage{amsmath}
\usepackage{amsthm}
\usepackage{inconsolata}
\usepackage{xcolor}         % colors
\usepackage{enumitem}
\usepackage{natbib}
\usepackage{multirow}
\usepackage{etoc}
\usepackage{makecell}
\usepackage{adjustbox}
\usepackage{tikz}

\usepackage{hyperref}

\definecolor{royalblue}{rgb}{0.25, 0.41, 0.88}

\hypersetup{
  colorlinks,
  citecolor=blue,
  linkcolor=royalblue,
  urlcolor=royalblue
}

\usepackage{algorithm}
\usepackage{algpseudocode}

\usepackage{mathtools}
\usepackage{amsthm}

\usepackage{cancel}
\usepackage{thm-restate}
\usepackage[mathscr]{euscript}
\usepackage[toc,page,header]{appendix}
\usepackage{dsfont}

\usepackage{authblk}
\usepackage{amssymb}
\usepackage{latexsym}
\usepackage{graphicx}
\usepackage{color}
\usepackage{graphics}

\usepackage[capitalize,noabbrev]{cleveref}

\usepackage[utf8]{inputenc} % allow utf-8 input
\usepackage[T1]{fontenc}    % use 8-bit T1 fonts
\usepackage{hyperref}       % hyperlinks
\usepackage{url}            % simple URL typesetting
\usepackage{booktabs}       % professional-quality tables
\usepackage{amsfonts}       % blackboard math symbols
\usepackage{nicefrac}       % compact symbols for 1/2, etc.
\usepackage{microtype}      % microtypography
\usepackage{graphicx}
\usepackage{amsmath}
\usepackage{cleveref}
\usepackage{soul}
\usepackage{hyphenat}
\usepackage{natbib}
\usepackage{geometry}
\usepackage{mathtools}
\usepackage{wrapfig}
\theoremstyle{plain}

\newtheorem{definition}{Definition}[section]
\newtheorem{theorem}{Theorem}[section]
\newtheorem{proposition}{Proposition}[section]
\newtheorem{lemma}{Lemma}[section]

\newtheorem{corollary}{Corollary}[section]
\usepackage{thmtools}

\algnewcommand{\IfThen}[2]{% \IfThen{<if>}{<then>}
  \State \algorithmicif\ #1\ \algorithmicthen\ #2}

\definecolor{bla}{HTML}{E63946}  % red
\definecolor{blb}{HTML}{2A9D8F}  % teal
\definecolor{blc}{HTML}{E9C46A}  % gold
\definecolor{bld}{HTML}{264653}  % dark blue
\definecolor{ble}{HTML}{F4A261}  % orange
\definecolor{blf}{HTML}{6A4C93}  % purple
\definecolor{blg}{HTML}{1982C4}  % blue
\definecolor{blh}{HTML}{8AC926}  % green

\definecolor{blqs}{HTML}{E63946}   % red
\definecolor{blqsd}{HTML}{F4A261}  % orange
\definecolor{blqsa}{HTML}{E9C46A}  % gold
\definecolor{blms}{HTML}{264653}   % dark blue
\definecolor{blmsa}{HTML}{2A9D8F}  % teal
\definecolor{bluhr}{HTML}{1982C4}  % blue
\definecolor{blshr}{HTML}{6A4C93}  % purple
\definecolor{blfe}{HTML}{8AC926}   % green
\definecolor{blbt}{HTML}{D62828}   % dark red
\definecolor{blhg}{HTML}{457B9D}   % steel blue
\definecolor{bloe}{HTML}{BC6C25}   % brown
\definecolor{blex}{HTML}{606C38}   % olive
\definecolor{ar}{rgb}{0.6, 0.4, 0.8}
\definecolor{atq}{HTML}{8E44AD} % Active Testing w/ QS
\definecolor{atm}{HTML}{16A085} % Active Testing w/ MS

\definecolor{mr}{HTML}{8C564B}

\usepackage[framemethod=TikZ]{mdframed}
\definecolor{OurColor}{HTML}{36aa70}
\definecolor{UserExampleBg}{HTML}{ffffff}
\definecolor{UserExampleTitle}{HTML}{545f7f}
\newmdenv[
    roundcorner=5pt,
    backgroundcolor=UserExampleBg,
    linecolor=UserExampleTitle,
    outerlinewidth=0.5pt,
    frametitlebackgroundcolor=UserExampleTitle,
    frametitlefont={\bfseries\color{white}},
]{user_example}

\AtBeginEnvironment{user_example}{\setlength{\parindent}{0pt}}

\title{Sample Efficient Ranking of Language Models Using \\ Pairwise Comparisons}

\usepackage{bbold}

\author{\normalsize Somnath Basu Roy Chowdhury\textsuperscript{1} \quad Rahul Kidambi\textsuperscript{1} \quad Avinava Dubey\textsuperscript{1} \quad Amr Ahmed\textsuperscript{1} \\\vspace{-10pt} Yeganeh Alimohammadi\textsuperscript{1,2} \\\vspace{5pt}
\textsuperscript{1}Google Research, \textsuperscript{2}University of Southern California \\
\texttt{\small \{somnathbrc,rahulkidambi,avinavadubey,amra,yeganeh\}@google.com}
}

\begin{document}

\date{}
\maketitle

\begin{abstract}
The frontier of large language model (LLM) research is advancing rapidly, with new models being released every few weeks. In this fast-paced environment, rigorous evaluation is crucial to accurately assess a model's quality. Yet evaluation prompts vary in how much evidence they provide about the relative quality of a model pair. In this work, we study how to reduce evaluation cost by adaptively selecting both model pairs and evaluation prompts while controlling comparison errors.
Our contribution is twofold:  First, we formulate pairwise LLM evaluation as a sequential hypothesis-testing problem and develop an e-process-based procedure. We further derive a sampling distribution for actively selecting evaluation prompts.  Second, we construct a complete LLM ranking by using this pairwise evaluation procedure as a black-box comparison function within a sorting algorithm. To further reduce evaluation cost, we adaptively allocate the error budget across pairwise comparisons.   Empirically, across both synthetic and real-world setups, we observe that our approach can efficiently rank LLMs using only $\sim$7-22\% of the total evaluation budget.
\end{abstract}

\section{Introduction}

Evaluating language models on open-ended tasks often relies on human feedback, which is costly to collect. Platforms such as Chatbot Arena~\citep{chiang2024chatbot} rank models by asking users to compare pairs of anonymous  model responses to their prompts. This process is costly: human annotations are expensive, many prompts provide little information for distinguishing models, and the emergence of long-context tasks has made generating model responses substantially more expensive. To reduce these costs, prior work has studied active ranking techniques~\citep{jamieson2011active, maystre2017just, ar}, which adaptively select the item pairs to compare in order to minimize the total number of comparisons. These techniques typically assume that evaluating an item pair yields an immediate noisy comparison outcome. For LLM evaluation, however, a comparison outcome is not directly observed: it must be elicited by selecting a prompt, generating responses from both models, and collecting a human judgment. Therefore, efficient LLM ranking requires adapting not only over model pairs, but also over the prompts used to compare them.

In this work, we bridge this gap with an algorithm that adaptively selects both the pairs of LLMs to compare and the prompts on which to evaluate them. We achieve this by proposing a sample efficient pairwise evaluation algorithm and leverage this algorithm within a sorting routine to construct a full ranking of models. First, we formulate pairwise LLM evaluation as a sequential hypothesis-testing problem and use the \textit{e-process} framework~\citep{testingbybetting}. An e-process sequentially accumulates evidence against a null hypothesis through betting, yielding an anytime-valid stopping rule with Type-I error controlled at level $\alpha$. For each model pair, we construct two e-processes, one for each LLM, to gather evidence that the corresponding model is the superior one. We then derive an active sampling distribution that maximizes the e-process’s expected growth rate. Its structure prioritizes prompts on which human annotators are likely to confidently express a  preference between the two models. Because human preferences are not known in advance, we introduce a practical LLM-driven strategy for approximating this distribution without explicitly modeling them.
Second, we use a comparison-based sorting algorithm to turn the pairwise procedure into a full-ranking algorithm. Each comparison invokes the pairwise evaluator as a black box and returns the superior model with a tunable correctness probability of $1-\alpha$. Finally, to further reduce evaluation cost, we optimize an end-to-end objective, which jointly balances ranking quality and sample complexity to adaptively allocate the error budget across pairwise comparisons within sorting algorithms.

Our proposed ranking algorithm produces high-quality rankings using only a fraction of the evaluations required by existing approaches. Empirically, we show that our ranking algorithm produces accurate results using only $\sim$7-22\% of the total evaluation budget. We further demonstrate that prioritizing discriminative prompts substantially accelerates evidence accumulation during pairwise evaluation. When combined with a comparison-based sorting algorithm, our approach focuses evaluations on the most relevant model pairs while allocating the error budget efficiently. To summarize, our contributions are as follows:
\begin{itemize}[leftmargin=*]
    \itemsep0mm
    \item We frame pairwise evaluation of language models as a sequential hypothesis testing problem and formulate an e-process for this setup. We derive an active sampling distribution that maximizes the growth rate of the e-process, enabling selection of discriminative prompts.
    \item We use a comparison-based sorting algorithm to construct LLM rankings from our pairwise evaluation procedure. We further adaptively allocate the error budget across comparisons by optimizing an objective that balances ranking quality and sample complexity.
    \item We provide rigorous theoretical bounds on the expected sample complexity for both the pairwise evaluation and ranking problem.
    \item Empirically, we show that our proposed approach is able to construct high quality rankings using only $\sim$7-22\% of the evaluation budget across several datasets.
\end{itemize}

\section{Background}

% [\textcolor{red}{Work in Progress, Don't review}]

In this section, we provide an outline of the active testing and active ranking literature and discuss its connection to our LLM ranking setup. We then introduce sequential hypothesis testing, a key component of our proposed pairwise evaluation framework.\\[-0.5em]

\noindent\textbf{Active Testing.}
Active testing~\citep{kossen2021active, berrada2026scaling, sawade2010active} focuses on adaptively selecting examples to label in order to efficiently estimate the performance of a
single model on a given benchmark. Unlike active learning~\citep{houlsby2011bayesian, farquharstatistical}, which aims to reduce uncertainty in model predictions, active testing treats
the model as fixed and prioritizes selecting examples that reduce uncertainty about the performance estimate itself (e.g., the variance of the risk
estimator~\citep{sawade2010active}). Recent work has improved the proposal distribution for active estimation by leveraging predictions from other LLMs~\citep{wu2026efficient}, while a
separate line of work estimates model performance using synthetic labels from LLM verifiers~\citep{zhou2025accelerating, angelopoulos2025cost, cowenmultiple}. We argue that most active
testing techniques are not ideal for pairwise model evaluation, as the primary objective is to reliably identify the better model rather than accurately estimate the win rate.
\citet{sawade2012active} introduced early work on active testing with pairwise predictions, prioritizing examples where two classification models' predictions differ. However, extending
this approach to modern LLMs is non-trivial, as quantifying the difference between two model responses is itself expensive, and for complex tasks, even generating the responses can be costly.
To address this issue, we formulate pairwise evaluation as a hypothesis testing problem and introduce an active selection strategy that accelerates convergence.\\[-0.5em]

\noindent\textbf{Active Ranking.}
Active ranking techniques can be broadly classified into two categories: methods that rank models using a fixed benchmark~\citep{zhou2025speeding, zouhar2026dynamically} and methods that rank models using pairwise evaluation data~\citep{gu2026anytimevalid}.
In our work, we focus on active
 ranking methods using pairwise comparisons, which reduce annotation costs by adaptively selecting informative pairs. In one of the first works in this  setup, \citet{jamieson2011active} 
  perform ranking in a $d$-dimensional Euclidean space by exploiting low-dimensional geometric structure to reduce comparison complexity. \citet{ailon2011active} considers a more general
  setting and obtains query-efficient approximate rankings under potentially non-transitive preferences. Sorting-based approaches include \citet{maystre2017just}, who analyze quicksort
  under Bradley-Terry assumptions, and \citet{pmlr-v70-falahatgar17a}, who combine confidence-based sequential comparisons with merge sort and noisy binary search to obtain PAC ranking
  guarantees. \citet{ar} propose an active ranking algorithm that doesn't require parametric assumptions about win probabilities, which keeps track of individual win rates and stops when the confidence interval of the collected data are sufficiently separated. 
  These methods primarily optimize \textit{which items to compare}, treating observations for each pair as supplied by a noisy comparison oracle. In contrast, our work
  \textit{jointly optimizes both LLM pair selection and prompt selection} for the corresponding pairwise evaluation, improving ranking efficiency along both axes.\\[-0.5em]

\noindent\textbf{Sequential Hypothesis Testing.}
Sequential hypothesis testing has a rich theoretical history rooted in Wald's Sequential Probability Ratio Test (SPRT) \citep{wald1992sequential}, which tests a simple null $H_0: \theta = \theta_0$ against a simple alternative $H_1: \theta = \theta_1$ by accumulating a likelihood ratio statistic $\Lambda_t = \prod_{i=1}^t \frac{p_{\theta_1}(X_i)}{p_{\theta_0}(X_i)}$ and stopping at the first time $\tau$ this ratio crosses a threshold. \citet{wald1948optimum} established that the SPRT minimizes expected sample size among all tests with prescribed Type I and Type II error rates $(\alpha, \beta)$. This result established that sequential methods are superior to fixed-sample tests, which require fixing the sample size in advance. Follow-up works~\citep{robbins1970statistical, lai1988nearly}, extended sequential tests to composite hypotheses and studied asymptotic optimality.

A related line of work reframes sequential testing within a betting or gambling framework. The central object is an \emph{e-value}: a nonnegative random variable $E$ satisfying the bound, $\mathbb{E}_{H_0}[E] \leq 1$, under the null hypothesis. $E$ variable quantifies evidence against the null hypothesis, a larger $E$ indicates stronger evidence \citep{vovk2021values, grunwald2024safe}. A canonical instantiation is the \emph{wealth process} shown below:
\begin{equation}
    E_t = E_{t-1}e_t(X_t), \text{ where } e_t(X_t) = 1 + \lambda_t(X_t - \mu_0) 
\end{equation}
where $\lambda_t \in \mathcal{F}_{t-1}$ is a bet placed using only past observations, $\mu_0$ is the null mean, and $E_0=1$. Under the null hypothesis $H_0$, $\{E_t\}_{t \geq 0}$ is a nonnegative martingale with $\mathbb{E}_{H_0}[E_t] \leq 1$. Using Ville's inequality \citep{ville1939etude}, we can reject the null hypothesis when $E_t \geq 1/\alpha$ and still control the Type I error at $\alpha$ uniformly over all stopping times. 
Intuitively, this implies that under the null hypothesis, it 
is highly unlikely for wealth to grow to a large amount; 
therefore, we can reject the null hypothesis when 
$E_t \geq 1/\alpha$.  
Unlike $p$-values, products of e-values $\prod_{t=1}^{\tau} e_t(X_t)$ remain valid at any data-dependent $\tau$, making them composable by construction. The optimal bet maximizing expected log-wealth, $\lambda^* = \operatorname{arg\,max}_{\lambda}\, \mathbb{E}_{H_1}[\log E_t]$, recovers the SPRT under simple hypotheses and extends naturally to composite settings \citep{testingbybetting, waudby2024estimating}. Unifying these ideas, an \emph{e-process} $\{E_t\}_{t \geq 0}$ \citep{shekhar2024nonparametric} requires $\sup_{\tau \in \mathcal{T}}\,\mathbb{E}_{H_0}[E_\tau] \leq 1$ for \emph{all} stopping times $\tau \in \mathcal{T}$, strictly generalizing test martingales to composite and nonparametric nulls. 
Recently, e-processes have been applied to LLM evaluation in two concurrent works that address different problem settings. \citet{gu2026anytimevalid} combine e-processes with i.i.d.\
prompt sampling to produce an anytime-valid {partial} ordering of LLMs, whereas we target a {full} ranking. \citet{hsu2026efficient} adaptively select prompts to maximize the
growth rate of an e-process, but evaluate a {single} model on a fixed benchmark rather than comparing multiple models. In contrast, our contribution lies in jointly optimizing
prompt selection via active sampling and error allocation via adaptive $\alpha$ within a comparison-based sorting algorithm.
\section{Efficient Ranking Using Pairwise Comparisons}

We introduce a ranking algorithm that makes two decisions: which model pairs to compare and which prompt to use  within each comparison. A comparison-based sorting algorithm makes the first decision. A pairwise evaluator makes the second, collecting human judgments until the evidence supports a clear winner.
We first construct this evaluator and bound its sample complexity, then use these guarantees to guide the ranking algorithm. \\[-0.5em]

\noindent\textbf{Problem Setup.}
Consider $n$ models $\{M_1, M_2, \ldots, M_n\}$. For each model pair $(M_i, M_j)$, let $\mathcal{X}_{ij} = \left\{x_1^{ij}, x_2^{ij}, \ldots \right\}$ be a finite prompt population with a fixed \textbf{target distribution} $p^{ij}(x)>0$.  A human evaluation is a binary label $Y_{ij}\in\{0,1\}$, where $Y_{ij}=1$ indicates a preference for $M_j$ and vice-versa.  Write $\pi_{ij}(x)$ for the probability of this outcome on prompt $x$.
The \textbf{target win rate}, and the  preference gap magnitude are:
\begin{equation}
\mu_{ij}=\sum_{x\in\mathcal X_{ij}}p^{ij}(x)\pi_{ij}(x), \qquad \Delta_{ij}=|2\mu_{ij}-1|.
\label{eq:target-win-rate}
\end{equation}
Thus, $M_j$ is preferred when $\mu_{ij}>1/2$.  The target distribution specifies which comparison we want. For example, two equally weighted prompts with $\pi(x)$: $0.9$ and $0.6$ give a target win rate of $0.75$.

\subsection{Efficient Pairwise Evaluation} \label{sec:pairwise}
Our goal is to select prompts and based on that sample human judgments. In this section, we fix a model pair $(M_i, M_j)$, and omit its subscripts. 
At round $t$, let $\mathcal F_{t-1}$ contain the information available before the next judgment. Using this information, choose a \textbf{sampling distribution} $q_t$ (which we specify later) and draw a prompt $X_t\sim q_t$.  Throughout, we require $q_t(x)>0$ wherever $p(x)>0$. To account for noisy human judgment, we assume that each input $x$ is associated with a latent preference probability, $\pi(x) \in [0, 1]$, and the human labels drawn from:
\begin{equation}
\mathbb P\bigl(Y_t=1\mid\mathcal F_{t-1},X_t=x\bigr)=\pi(x).
\label{eq:sequential-model}
\end{equation}
This lets prompt selection to be history-dependent but human labels are conditionally independent.

\noindent\textbf{E-processes and stopping.}
Since the winner is initially unknown, we test two null hypotheses simultaneously: $H_0^{+}\!: \mu \le 1/2$ ($M_i$ is better than $M_j$) and $H_0^{-}\!: \mu \ge 1/2$ ($M_j$ is better than $M_i$). We do not consider the possibility of a tie as it is unlikely in real-world settings (the boundary case $\mu=1/2$ is the only point at which both nulls hold, and is handled separately in Appendix~\ref{app:validity-proof}). We
construct two e-processes, $E_t^+$ and $E_t^-$, that accumulate evidence against $H_0^+$ and $H_0^-$, respectively:
\begin{align}
e_t^+(y)&=1+\lambda_t(y-1/2), & E_t^+&=E_{t-1}^+\frac{p(X_t)}{q_t(X_t)}e_t^+(Y_t),\nonumber\\
e_t^-(y)&=1-\lambda_t(y-1/2), & E_t^-&=E_{t-1}^-\frac{p(X_t)}{q_t(X_t)}e_t^-(Y_t),
\label{eqn:eprocess}
\end{align}
where $\lambda_t = 4(\nu_t -1/2)$ (following \citet{breiman1961optimal}) and the bet $\nu_t\in[1/2,1)$ controls the strength of each label's contribution.
Both processes start from $E_0^+=E_0^-=1$, and it is easy to see that $E_t^+$ grows as we collect evidence for model $M_j$ (when $y=1$) and $E_t^-$ grows when we collect evidence for $M_i$  (when $y=0$).  We first allow any such bet; next section specifies how to choose it from $\mathcal F_{t-1}$. Both processes use the same prompt, judgment, and realized sampling probability.
For a given error tolerance $\alpha\in(0,1)$,
we stop when either process crosses the threshold:
\begin{equation}
\tau_\alpha=\inf\{t\ge1:\max(E_t^+,E_t^-)\ge1/\alpha\}.
\label{eq:pairwise-stopping}
\end{equation}
The pairwise algorithm returns $M_j$ if $E_t^+$ crosses first and $M_i$ otherwise.

\begin{lemma}\label{lem:valid_eprocess}
 Under  \eqref{eq:sequential-model},   ${E}_t^+$ is a nonnegative supermartingale under $H_0^{+}$, and $E_t^-$ is a nonnegative supermartingale under $H_0^{-}$. For any $\mu\neq.5$, the probability of stopping and returning the inferior model  is at most $\alpha$,  regardless of prediction accuracy of $q_t$. 
\end{lemma} 
In fact, this lemma shows that $E^+$ and $E^-$ are valid e-processes and using Ville's inequality we can bound the crossing probability by $\alpha$~\citep{ville1939etude,ramdas2025hypothesis} (see Appendix~\ref{app:validity-proof}). At a tie where $\mu=.5$, one can use $2/\alpha$ threshold to control the probability of either declaration by $\alpha$.\\[-0.5em]

\noindent\textbf{Choosing prompts.}
To choose prompts, we examine how quickly each process grows on a log-scale.
We first treat  $\pi(x)$ as known; and then replace it by predictions to obtain an implementable rule.
Following the notion of e-power~\citep{ramdas2025hypothesis}, define the contribution of a prompt  as the local e-power for each prompt $x$ and each direction $s\in\{+,-\}$: $  h^s_{{t}}(x) = \mathbb{E}_{{Y_t}} \left[ \log e_t^s(Y_t) \mid \mathcal{F}_{t-1},{X_t =} x \right]$.
Then including the importance weighting, the expected log increment under  proposal $q$ is
\begin{equation}
G_t^s(q)=\sum_xq(x)h_t^s(x)-D_{\mathrm{KL}}(q\|p), \qquad \text{for  } s\in\{+,-\}.
\label{eq:directional-growth}
\end{equation}
The first term rewards  evidence; the second accounts for the cost of changing the prompt distribution.  Here $D_{\mathrm{KL}}$ denotes KL divergence, including between Bernoulli distributions when its arguments are probabilities.
The \textbf{optimal proposal} and its rate are
\begin{equation}
q_t^{s,\star}(x)=\frac{p(x)e^{h_t^s(x)}}{\sum_zp(z)e^{h_t^s(z)}},
\qquad G_t^{s,\star}=\log\sum_xp(x)e^{h_t^s(x)}.
\label{eq:directional-oracle}
\end{equation}
Indeed, $G_t^s(q)=G_t^{s,\star}-D_{\mathrm{KL}}(q\|q_t^{s,\star})$. This identity identifies both the optimizer and the exact loss from using a different proposal. For our factors it gives
\begin{equation}
q_t^{+,\star}(x)\propto p(x)\left(\frac{\nu_t}{1-\nu_t}\right)^{\pi(x)},
\qquad
q_t^{-,\star}(x)\propto p(x)\left(\frac{1-\nu_t}{\nu_t}\right)^{\pi(x)}.
\label{eq:directional-tilts}
\end{equation}
The positive proposal favors prompts likely to favor $M_j$; the negative proposal favors the opposite direction. Their optimality concerns one-step log growth at the current bet. Appendix~\ref{app:oracle-proof} gives the derivation and its local-e-power interpretation.

The two proposals answer different questions: where to seek evidence for $M_j$ versus for $M_i$. Since the winner is unknown, we combine them in proportion to their accumulated evidence:
\begin{equation}
\widetilde q_t(x)
=w_{t-1}q_t^{+,\star}(x)+(1-w_{t-1})q_t^{-,\star}(x),
\qquad
w_t={E_t^+}/({E_t^++E_t^-}).
\label{eq:mixture-proposal}
\end{equation}
 Initially $w_0=1/2$; thereafter, the proposal associated with the larger process receives the larger weight.   We call this \emph{the oracle mixture} when it uses the true $\pi(x)$.

In practice, the preference probabilities $\pi(x)$ are unknown and estimating them is challenging. We explore various online learning based predictors $\hat \pi(x) = f(x)$ using past observations (Sec.~\ref{sec:pair_eval}) but found an LLM-based predictor to be most effective. An LLM predicts $\hat \pi(x)$ from prompt texts and previous  comparisons; we also consider a predictor trained on prompt embeddings. These predictions guide prompt selection, while the human judgment supplies the label used to update evidence.
  Replacing $\pi$ in Eq.~\ref{eq:directional-tilts} with predicted scores $\widehat\pi_{t-1}(x)\in[0,1]$ ((normalized over $\mathcal X$ or over a shortlist $C_t\subseteq\mathcal X$)  gives normalized proposals $\widehat q_t^+$ and $\widehat q_t^-$. We use
\begin{equation}
q_t=\kappa p+(1-\kappa)\bigl[w_{t-1}\widehat q_t^++(1-w_{t-1})\widehat q_t^-\bigr],
\qquad \kappa\in(0,1].
\label{eq:practical-proposal}
\end{equation}
This final $q_t$ is the proposal used both to draw the prompt and to compute the importance weight.\\[-0.5em]

\noindent\textbf{Choosing the bet size.} 
The matched target-sampling bet is $\nu^\star=\max\{\mu,1-\mu\}$, obtained by maximizing the expected Bernoulli log factor~\citep{breiman1961optimal}. We estimate it from past judgments:
\begin{align}
\widehat\mu_{t-1}
&=\operatorname{clip}_{[0,1]}\!\left(\frac1{t-1}\sum_{u<t}\frac{p(X_u)}{q_u(X_u)}Y_u\right),\quad t\ge2,
\label{eq:mean-estimate}\\
\nu_t&=\min\{\bar\nu,\max\{\widehat\mu_{t-1},1-\widehat\mu_{t-1}\}\},
\qquad 1/2<\bar\nu<1.
\label{eq:adaptive-bet}
\end{align}
A prespecified $\nu_1\in[1/2,\bar\nu]$ initializes the procedure. The cap keeps both factors positive. These steps define \texttt{PairEval}$(M_i,M_j,\alpha)$: choose the bet and proposal, draw a prompt and judgment, update both processes, and test Eq.~\ref{eq:pairwise-stopping}.\\[-0.5em]

\noindent\textbf{Pairwise sample complexity.} \label{sec:sample-complexity}
We next relate the evidence rate to the number of human evaluations. Without loss of generality assume $\mu>.5$. To isolate the cost of learning the directions, we first analyze the oracle version of the completed procedure: $\pi(x)$ are known, the bet is fixed  $\nu_t=\mu$. The component proposals are therefore fixed, and only their mixture weights adapt, resulting in the correct proposal's growth rate:
\begin{equation}
G^\star=\log\sum_xp(x)(2\mu)^{\pi(x)}\{2(1-\mu)\}^{1-\pi(x)}.
\label{eq:oracle-rate}
\end{equation}
The resulting bound connects the local growth objective to expected evaluation cost for this benchmark; the learned rule requires additional growth control.

\begin{theorem}[Sample complexity]
\label{thm:sample_complexity}
Suppose $\mu\in(1/2,1)$ and at least one prompt has $0<\pi(x)<1$.   For the oracle mixture with bet $\mu$, a finite constant $C$, independent of $\alpha,\beta\in(0,1)$, satisfies
\begin{equation}
\mathbb E[\tau_\alpha]\le\frac{\log(1/\alpha)}{G^\star}+C,\qquad \mathbb E[\tau_\alpha]\sim\frac{\log(1/\alpha)}{G^\star}
\qquad\text{as }\alpha\downarrow0.
\label{eq:mixture-asymptotic}
\end{equation}
Moreover, with probability at least $1-\beta$,
\begin{equation}
\tau_\alpha\le \frac{\log(1/\alpha)}{G^\star}
+C\left\{\sqrt{\log(1/\alpha)\log(2/\beta)}+\log(2/\beta)+1\right\}.
\label{eq:cost-tail-main}
\end{equation}
\end{theorem}
Learning the direction therefore leaves the leading sample cost unchanged. The proof has a simple structure: importance weights cancel in $E_t^+/E_t^-$; the expected weight on the wrong proposal decreases geometrically; and its cumulative KL loss is finite. The interior-preference assumption supplies the strict drift needed for this argument. Appendices~\ref{app:mixture-proof}--\ref{app:oracle-concentration} give the proof. 
The gain over target sampling comes from prompt heterogeneity: $G^\star\ge D_{\mathrm{KL}}(\mu\|1/2)$, exactly when $\pi(x)$ varies across prompts. Thus the oracle mixture has a smaller leading expected cost in heterogeneous populations (Appendix~\ref{app:heterogeneity-proof}).\\[-0.5em]

\noindent\textbf{When does prompt selection help?}
Our analysis also explains when imperfect  prediction of $\pi$ can reduce evaluation cost. Suppose that the prediction error has population variance $\operatorname{Var}_{X\sim p}(\widehat\pi_{t-1}(X)-\pi(X))\le\sigma^2$.
With the matched bet $\nu=\mu$, replacing the true probabilities by these predictions reduces the expected log growth of each directional proposal by at most \(\frac{\sigma^2}{2}\frac{\mu}{1-\mu}\log^2\!\left(\frac{\mu}{1-\mu}\right)\). 
This bound quantifies the cost of prediction error. In
particular, for any $\kappa \in [0, 1)$, when this penalty is smaller than the 
oracle gain, $G^* - D_{KL}(\mu \| 1/2)$, 
prompt selection strictly improves the leading rate over target sampling ($G_{\mathrm{LLM}} > D_{\mathrm{KL}}(\mu \| 1/2)$), while $\kappa > 0$ ensures full support and concentration of $\nu_t$. Appendices~\ref{app:prediction-proof} state the precise
condition and extend the analysis to $\nu_t$ estimated from past observations 
(both under Eq.~\ref{eq:practical-proposal} and under remaining-prompt exploration on $\mathcal X \setminus C_t$). Regardless of prediction
accuracy, the comparison-error guarantee in Lemma~\ref{lem:valid_eprocess} remains valid.

% This bound quantifies the cost of prediction error.  In particular, when this penalty is smaller than the oracle gain, $G^\star-D_{\mathrm{KL}}(\mu \parallel 1/2)$, prompt selection can reduce the leading expected sample cost. Appendices~\ref{app:prediction-proof} state the precise condition and extend the analysis to $\nu_t$ estimated from past observations. Regardless of prediction accuracy, the comparison-error guarantee in Lemma~\ref{lem:valid_eprocess} remains valid.

In the following section, we treat the proposed pairwise evaluation algorithm as a black box, accessed via a single function call \texttt{PairEval}$(M_i,M_j,\alpha_{ij})$. This function takes as input a model pair $(M_i,M_j)$ and a target error rate $\alpha_{ij}$, and returns the superior model with probability at least $1-\alpha_{ij}$.

\begin{figure}
    \centering
    \includegraphics[width=0.9\linewidth]{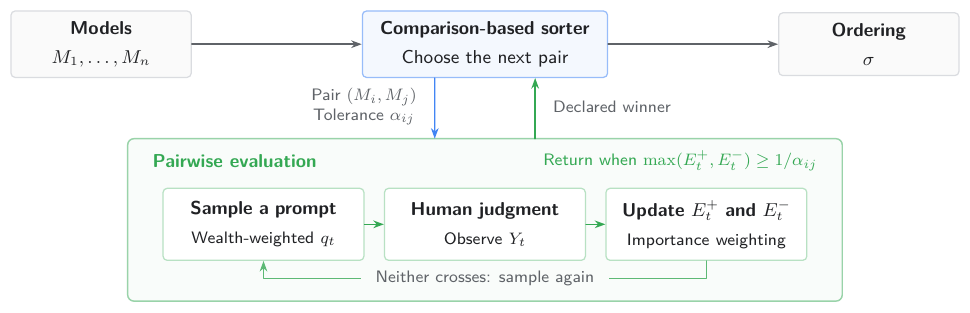}
    \caption{\textbf{Ranking with adaptive pairwise evaluation}. A comparison-based sorter requests model pairs. For each pair, the evaluator adaptively samples prompts and updates two e-processes.}
    \label{fig:highlevel}
    \vspace{-15pt}
\end{figure}

\subsection{The Ranking Algorithm}
\label{sec:ranking}

We now use the pairwise evaluator to construct an ordering of $n$ LLMs. Pairwise preferences may contain cycles, as illustrated in Figures~\ref{fig:dag1} and~\ref{fig:dag2}, so no linear ordering may agree with every pairwise winner. So before  describing the algorithm, we need to formalize the ranking objective. 
For a permutation $\sigma$, where larger $\sigma(i)$ means higher rank for $M_i$, we define the ranking quality as:
\begin{equation}
Q(\sigma)
=
\sum_{1 \leq i < j \leq n}
\mu_{ij}\mathbf{1} [\sigma(j)>\sigma(i)]
+
(1-\mu_{ij})\mathbf{1}[\sigma(i)>\sigma(j)].
\label{eq:ranking-quality}
\end{equation}
Each term measures the expected agreement between ordering \(\sigma\) and the pair's human preference. 
To improve the sample efficiency of ranking, we reduce both (a) the number of calls to \texttt{PairEval}$(\cdot,\cdot,\cdot)$ and (b) the number of samples required within each call. For (a), we use any general comparison based sorting algorithm. For (b), we adapt the error rate \(\alpha_{ij}\) to the pair \((M_i,M_j)\) being compared.\\[-0.5em]

\noindent\textbf{Using a comparison-based sorter.}
Let $\mathcal A$ be a comparison-based ranking algorithm, e.g., mergesort or quicksort. At call $k$, it requests a pair $e_k=(i_k,j_k)$ and receives the result of \texttt{PairEval}$(M_{i_k},M_{j_k},\alpha_k)$. The tolerance $\alpha_k$ is chosen before the call. Each evaluation starts from unit evidence and uses fresh judgments satisfying Eq.~\ref{eq:sequential-model}. $\mathcal A$ sees the returned winner, not the evaluator's internal sample path, and uses these winners to choose subsequent comparisons (see Fig.~\ref{fig:highlevel}).

This interface separates the statistical evaluation from the sorting procedure. Sorting-based active ranking has been studied previously~\citep{maystre2017just,pmlr-v70-falahatgar17a}; here the evaluator also controls how prompts are selected within each comparison.
  Next, we derive a principled way to allocate these pair-specific error rates without compromising the ranking quality.\\[-0.5em]

\noindent\textbf{Allocating error across comparisons.}  We allocate error tolerances across comparisons based on their estimated sampling costs.
Let $\mathcal E$ be the set of unordered pairs that the ranker may request. If the ranker requests a pair that has already been compared, we use the previously declared winner. Thus, each pair incurs an error tolerance only once and requires no additional labels on subsequent requests (Appendix~\ref{app:ranking-analysis} extends the analysis to repeated comparisons). When a pair is first requested,  obtain estimates of  $\widehat G_{ij}$ and  $\widehat\Delta_{ij}$.   Let $N_0$ be total pilot judgments used for the initial estimations over all pairs. These estimates may come from a small pilot sample. They are used only to allocate error budget; the validity of the pairwise test does not depend on their accuracy.

The sample complexity bound in \eqref{eq:mixture-asymptotic} motivates using  $\log(1/\alpha_{ij})/\widehat G_{ij}$ as the cost of evaluating pair $(i,j)$. 
Since the pairs ultimately requested by the ranking algorithm
are not known in advance, we minimize the sum over all the candidate pairs. This favors larger tolerances for comparisons that
are expensive to resolve. 
We also account for the smaller direct effect of errors
on nearly tied pairs. We use $\widehat\Delta_{ij}$ to
estimate the magnitude of the change in a pair's
contribution to ranking quality when the order of  $(i,j)$ is reversed. This motivates the estimated
direct-error budget
$\sum_{\mathcal E}\widehat\Delta_{ij}\alpha_{ij}
\le\varepsilon$.
Together with the global error budget
$\sum_{\mathcal E}\alpha_{ij}\le\delta$
and the bounds $0<\alpha_{ij}\le\bar\alpha<1$,
this defines a convex cost model whose exact optimizer
is given by Eq.~\ref{eq:allocation-policy}.
\begin{equation}
\alpha_{ij}=\min\!\left\{\bar\alpha,
\frac1{\widehat G_{ij}(\xi\widehat\Delta_{ij}+\zeta)}\right\},
\qquad (i,j)\in\mathcal E.
\label{eq:allocation-policy}
\end{equation}
The nonnegative multipliers $\xi,\zeta$ are chosen to satisfy the budgets and complementary slackness; Appendix~\ref{app:ranking-allocation} gives the optimization problems and the derivation. \\[-0.5em]

\noindent\textbf{Sample complexity of ranking.} 
We now translate the pairwise guarantees into a guarantee for the full ranking.
Let $M$ denote an upper bound on the number of pairwise comparisons requested by the ranking algorithm in one run.  For example, merge sort  uses $M=O(n\log n)$ comparisons. Also, let $\widehat{\sigma}_{\mathcal A}$ denote the ranking returned by
Algorithm~\ref{alg:ranking}. We compare it with the ordering
$\sigma_{\mathcal A}^{\star}$ that the same ranking algorithm would return if
every requested pair were answered by its true pairwise winner. Thus,
$\sigma_{\mathcal A}^{\star}$ separates errors due to pairwise evaluation
from the behavior of the ranking algorithm.

\begin{theorem}[Sample complexity of ranking]
\label{thm:ranking-cost}
Let $\mathcal E_{\mathrm{eval}}\subseteq\mathcal E$ be the
set of pairs evaluated during a run. 
Suppose that each pairwise
test satisfies the assumptions of Theorem~\ref{thm:sample_complexity}, and the tolerances satisfy $\sum_{(i,j)\in\mathcal E}\alpha_{ij}\le\delta$. Algorithm~\ref{alg:ranking} then satisfies
\(
    \mathbb P\!\left(
        \widehat{\sigma}_{\mathcal A}
        \neq
        \sigma_{\mathcal A}^{\star}
    \right)
    \le \delta.
\)
Moreover, let $N$ denote the total number of judgments collected during
ranking, and $M$ denote the number of pairwise comparisons requested by the ranking algorithm then
\begin{equation}
    \mathbb E[N]
    \le
    M\mathbb E\Big[\max_{(i,j)\in\mathcal E}
    \left\{
        \frac{\log(1/\alpha_{ij})}{G_{ij}}
        + C_{ij}
    \right\}\Big],
    \label{eq:ranking-cost-main}
\end{equation}
where $G_{ij}$ and $C_{ij}$ are the pairwise quantities in
Theorem~\ref{thm:sample_complexity}.
\end{theorem}
The theorem bounds the total cost by the number of comparison requests times the expected largest
per-pair cost bound encountered during the run. Moreover, under Theorem~\ref{thm:sample_complexity}, Eq.~\ref{eq:ranking-cost-main} holds with the oracle rate $G_{ij}=G_{ij}^\star$; Appendix~\ref{app:ranking-proof} (Corollary~\ref{cor:ranking-practical}) instantiates Eq.~\ref{eq:ranking-cost-main} for   LLM predictions under the learned bet $\nu_t$, and proves a tighter bound along the realized comparison path.

\section{Experiments}
\label{sec:exp}

In this section, we describe our experimental setup and evaluate the proposed pairwise evaluation and sample-efficient ranking algorithm across a range of settings. 
We first describe the experimental setup, then present results for the pairwise evaluation procedure, and the efficient ranking algorithm.

\subsection{Setup}
We perform experiments using pairwise LLM evaluation data available in three datasets: (a) LMArena Human Preference dataset (\texttt{arena-human-preference-140k}), which has 140K human labels across 1200+ model pairs, (b) Preference Proxy Evaluations (PPE)~\citep{frick2024evaluaterewardmodelsrlhf}, a collection of 16K labeled human preference pairs from Chatbot
Arena containing responses using over 121 languages and (c) the binarized version of the Ultrafeedback dataset (\texttt{ultrafeedback-binarized-preferences}), which is a filtered version of Ultrafeedback dataset~\citep{cui2023ultrafeedback} containing 63K pairwise examples from 17 LLMs. 
For pairwise evaluation, we consider all model pairs in PPE and retain only pairs with at least 200 labeled examples in  LMArena and UltraFeedback datasets.  For LLM ranking, we first construct a graph with LLMs as nodes, connecting two nodes with available evaluation data. We then extract the largest clique using
  the algorithm of \citet{boppana1992approximating} and restrict the ranking to models within this clique. This procedure yields 46 LLMs in LMArena, 17 in UltraFeedback, and 19 in PPE.

\subsection{Pairwise Evaluation}\label{sec:pair_eval}
In this section, we discuss the baselines, metrics, and results used for pairwise evaluation.\\[-0.5em]

\noindent\textbf{Baselines}.  We construct baselines to independently evaluate two decoupled components of our
pairwise evaluation approach: estimating a prompt's latent preference probability ($\pi(x)$) and the active sampling distribution, $\tilde{q}_t(x)$
(Eq.~\ref{eq:mixture-proposal}). To evaluate \textit{preference-probability estimation}, we fix
the sampling distribution to the derived optimal proposal and compare five
$\hat{\pi}(x)$ predictors: \textbf{\textcolor{bla}{[LM]}} an \texttt{LLM
estimator}, where an LLM assigns $\pi(x)$ scores to prompts using the raw prompt
text and previously observed comparison outcomes;
three variants of
\textit{embedding-based estimation} using online Bayesian logistic regression (\texttt{BLR})~\citep{bishop2006pattern} model, where we estimate $\hat{\pi}(x) = \sigma(\mu^\top \mathrm{emb}(x))$ and sequentially update
parameters after each observation. These variants differ in how the next example is selected using \textbf{\textcolor{blb}{[BLE]}} \texttt{BLR exploit}, by plugging in $\hat{\pi}(x)$ estimates into the derived sampling distribution (Eq.~\ref{eq:mixture-proposal}),
\textbf{\textcolor{blc}{[BLG]}} \texttt{BLR $\epsilon$-greedy}  that uses the proposal distribution with probability $(1-\epsilon)$ and uniform sampling with probability, $\epsilon$, and \textbf{\textcolor{bld}{[BLT]}} \texttt{BLR Thompson}, which uses Thompson Sampling~\citep{pmlr-v28-agrawal13} and considers the average proposal distribution across posterior parameter draws
$\theta \sim \mathcal{N}(\mu, \Sigma)$. Next, we compare against two different active sampling distributions:  \textbf{\textcolor{blg}{[AT]}} \texttt{Active
Testing}~\citep{kossen2021active}, which targets prompts with maximal preference
uncertainty ($q(x) \propto \sqrt{\hat{\pi}(x)(1 - \hat{\pi}(x))}$), {we use \texttt{BLR $\epsilon$-greedy} to estimate $\hat \pi(x)$ in this setup,} and \textbf{\textcolor{blh}{[RL]}}  \texttt{ROUGE-L divergence}~\citep{lin-2004-rouge} distribution, which completely bypasses preference probability estimation by prioritizing prompts that yield the most dissimilar candidate answers (low ROUGE-L). Unlike other baselines, ROUGE distribution requires access to model answers. 
We use Gemini 3.6 Flash model as the LLM estimator and obtain prompt embeddings from Gemini Embedding 2~\citep{shanbhogue2026gemini}. We report the average results across 5 runs for all baselines. We report additional results using open-sourced LLMs in Appendix~\ref{sec:addl_exps}.\\[-0.5em]

\begin{table}[t!]
\centering
\caption{We report the sample efficiency (win rate against uniform sampling; higher is better) and accuracy of pairwise evaluation for different methods across 3 datasets. ROUGE-L with \textcolor{gray}{greyed} out results has prior access to model answers.}
\label{tab:results}
\resizebox{\linewidth}{!}{
\begin{tabular}{lcc cc cc}
\toprule
 & \multicolumn{2}{c}{\textbf{LMArena}} & \multicolumn{2}{c}{\textbf{PPE}} & \multicolumn{2}{c}{\textbf{UltraFeedback}} \\
 \cmidrule(lr){2-3}\cmidrule(lr){4-5}\cmidrule(lr){6-7}
 \textbf{Method} & Win Rate $(\uparrow)$ & Acc. $(\uparrow)$ & Win Rate $(\uparrow)$ & Acc. $(\uparrow)$ & Win Rate $(\uparrow)$ & Acc. $(\uparrow)$\\
\midrule
\texttt{Active Testing} \textbf{\textcolor{blg}{[AT]}}           & 12.20\footnotesize{\color{gray} $\pm$4.15} & 93.17\footnotesize{\color{gray} $\pm$1.76}  & 38.04\footnotesize{\color{gray} $\pm$2.60} & 86.41\footnotesize{\color{gray} $\pm$1.13}  & 17.37\footnotesize{\color{gray} $\pm$3.19} & 97.46\footnotesize{\color{gray} $\pm$0.65} \\
\texttt{BLR exploit} \textbf{\textcolor{blb}{[BLE]}}              & 53.66\footnotesize{\color{gray} $\pm$7.40} & \textbf{94.63}\footnotesize{\color{gray} $\pm$1.57}  & 48.10\footnotesize{\color{gray} $\pm$2.83} & 85.87\footnotesize{\color{gray} $\pm$1.15}  & 44.07\footnotesize{\color{gray} $\pm$4.37} & 97.97\footnotesize{\color{gray} $\pm$0.58} \\
\texttt{BLR $\epsilon$-greedy} \textbf{\textcolor{blc}{[BLG]}}    & 45.12\footnotesize{\color{gray} $\pm$7.28} & 93.17\footnotesize{\color{gray} $\pm$1.76}  & 53.53\footnotesize{\color{gray} $\pm$2.80} & \textbf{87.39}\footnotesize{\color{gray} $\pm$1.09}  & 44.49\footnotesize{\color{gray} $\pm$4.39} & \textbf{98.31}\footnotesize{\color{gray}
$\pm$0.53} \\
\texttt{BLR Thompson} \textbf{\textcolor{bld}{[BLT]}}             & 26.83\footnotesize{\color{gray} $\pm$6.24} & 93.66\footnotesize{\color{gray} $\pm$1.70}  & 40.49\footnotesize{\color{gray} $\pm$2.64} & 87.28\footnotesize{\color{gray} $\pm$1.10}  & 36.02\footnotesize{\color{gray} $\pm$4.19} & 98.14\footnotesize{\color{gray} $\pm$0.56} \\
\texttt{LLM} \textbf{\textcolor{bla}{[LM]}}                      & \textbf{57.32}\footnotesize{\color{gray} $\pm$7.02} & 91.71\footnotesize{\color{gray} $\pm$1.93}  & \textbf{56.79}\footnotesize{\color{gray} $\pm$2.90} & 85.65\footnotesize{\color{gray} $\pm$1.16}  & \textbf{59.32}\footnotesize{\color{gray} $\pm$4.32} & 97.46\footnotesize{\color{gray} $\pm$0.65}
\\
\midrule
\textcolor{gray}{\texttt{ROUGE-L} \textbf{\textcolor{blh}{[RL]}}}           & \textcolor{gray}{42.68\footnotesize{ $\pm$7.02}} & \textcolor{gray}{94.63\footnotesize{ $\pm$1.57}}  &
\textcolor{gray}{44.29\footnotesize{ $\pm$2.83}} & \textcolor{gray}{85.43\footnotesize{ $\pm$1.16}}  & \textcolor{gray}{41.10\footnotesize{ $\pm$4.39}} & \textcolor{gray}{96.95\footnotesize{ $\pm$0.71}} \\
\bottomrule
\end{tabular}
}
\end{table}

\noindent\textbf{Metrics}. We assess pairwise evaluation methods along two axes: \textit{correctness}, measured as the percentage of evaluations that identified the correct winner; and \textit{sample efficiency}, measured as the
  \textit{win rate against uniform sampling}, i.e., the fraction of trials in which the method required fewer samples than the uniform baseline.

\noindent\textbf{Results}.   We report the pairwise evaluation results in Table~\ref{tab:results}. The \texttt{LLM} estimator
\textbf{\textcolor{bla}{[LM]}} consistently achieves the highest win rate (better sample efficiency) across all three datasets, outperforming uniform sampling by up to $\sim$9\% while maintaining competitive accuracy. Among embedding-based methods, \texttt{BLR exploit} \textbf{\textcolor{blb}{[BLE]}} and \texttt{BLR $\epsilon$-greedy} \textbf{\textcolor{blc}{[BLG]}} show moderate sampling efficiency in a few settings.  \texttt{Active Testing} \textbf{\textcolor{blg}{[AT]}} and \texttt{ROUGE-L} \textbf{\textcolor{blh}{[RL]}} both consistently underperform uniform sampling, indicating that neither uncertainty-based selection nor response divergence is an effective proxy for preference informativeness. Accuracy on PPE is relatively lower across all methods as all LLM pairs have fewer than 100 annotations. We observe that the \texttt{LLM estimator} significantly outperforms the \texttt{ROUGE-L divergence} baseline despite having no access to model answers.  We use the LLM-based predictor to report the results of sorting-based variants in the following section.

\subsection{Efficient Ranking}
In this section, we discuss the results of the ranking algorithm described in Section~\ref{sec:ranking}.\\[-0.5em] 

\noindent\textbf{Baselines}.
We compare our approach against three classes of baselines. The first class    provides stopping criteria for the ranking data acquisition process and consists of variants of our algorithm. We ablate two components---the sorting algorithm and error tolerance allocation, to construct the following baselines: \textbf{\textcolor{blqs}{[QS]}} quicksort with a constant error tolerance ($\alpha$) across all pairs; \textbf{\textcolor{blqsd}{[QSD]}} quicksort with a doubling error tolerance
strategy, where the earliest recursion steps receive the tightest allocation (low $\alpha_0$) that is gradually relaxed; \textbf{\textcolor{blqsa}{[QSA]}} quicksort with adaptive error allocation
(Eq.~\ref{eq:allocation-policy}); \textbf{\textcolor{blms}{[MS]}} merge sort with a constant error tolerance; and
\textbf{\textcolor{blmsa}{[MSA]}} merge sort with adaptive error allocation
(Eq.~\ref{eq:allocation-policy}).  The quicksort baselines with constant error allocation ($\alpha$) are a close variant of~\citet{maystre2017just}, where we plug-in e-process-based pairwise evaluation for noisy comparisons. 
The second class of methods consists of active estimation baselines: (a) Active Ranking \textbf{\textcolor{ar}{[AR]}}~\citep{ar}, which performs sequential ranking, and (b) Active
Testing~\citep{kossen2021active}, a pairwise evaluation method that we plug into quick sort \textbf{\textcolor{atq}{[ATQ]}} and merge sort \textbf{\textcolor{atm}{[ATM]}}.
 The third class comprises offline methods that use the entire dataset to construct a ranking, providing an estimate
of the upper bound on ranking quality. We consider \textbf{\textcolor{blbt}{[BT]}} Bradley-Terry ranking~\citep{bradley1952rank}, which assumes a transitive total ordering by assigning each LLM a scalar strength parameter;
\textbf{\textcolor{blhg}{[HG]}} HodgeRank~\citep{jiang2011statistical}, which relaxes the transitivity assumption and accommodates cyclic preferences in pairwise comparisons;
\textbf{\textcolor{bloe}{[OE]}} oracle estimation of the quality upper bound, $Q(\sigma_o) = \sum_{1 \le i < j \le n} \max\{\mu_{ij}, 1-\mu_{ij}\}$ (note that no linear ordering is guaranteed to achieve this bound); and
\textbf{\textcolor{blex}{[EX]}} expected quality achieved by using 10K random permutations (irrespective of the underlying dataset size).\\[-0.5em]

\begin{table}[t!]
\centering
\caption{We report the quality of ranking (agreement with pairwise data) and fraction of samples used with standard error over 3 runs. A higher quality score and lower fraction of samples is better.}
\label{tab:ranking-results}
\resizebox{\linewidth}{!}{
\begin{tabular}{lcc cc cc}
\toprule
 & \multicolumn{2}{c}{\textbf{LMArena}} & \multicolumn{2}{c}{\textbf{PPE}} & \multicolumn{2}{c}{\textbf{UltraFeedback}} \\
 \cmidrule(lr){2-3}\cmidrule(lr){4-5}\cmidrule(lr){6-7}
 \textbf{Method} & Quality $(\uparrow)$ & Samples $(\downarrow)$ & Quality $(\uparrow)$ & Samples $(\downarrow)$ & Quality $(\uparrow)$ & Samples $(\downarrow)$\\
\midrule
\texttt{Active Ranking} \textbf{\textcolor{ar}{[AR]}} & 59.8\footnotesize{\color{gray} $\pm$0.5} & 72.9\%\footnotesize{\color{gray} $\pm$1.8} & 57.4\footnotesize{\color{gray} $\pm$0.8} & 51.8\%\footnotesize{\color{gray} $\pm$4.4} & 65.4\footnotesize{\color{gray} $\pm$3.3} & ~~64.8\%\footnotesize{\color{gray} $\pm$11.0} \\
\texttt{Active Testing} (w/ QS) \textbf{\textcolor{atq}{[ATQ]}} & 60.6\footnotesize{\color{gray} $\pm$0.5} & 19.0\%\footnotesize{\color{gray} $\pm$1.0} & 59.2\footnotesize{\color{gray} $\pm$0.1} & 30.9\%\footnotesize{\color{gray} $\pm$1.2} & 75.0\footnotesize{\color{gray} $\pm$0.1} & 19.9\%\footnotesize{\color{gray} $\pm$0.7} \\
\texttt{Active Testing} (w/ MS) \textbf{\textcolor{atm}{[ATM]}} & 61.0\footnotesize{\color{gray} $\pm$0.2} & 16.8\%\footnotesize{\color{gray} $\pm$0.5} & 59.0\footnotesize{\color{gray} $\pm$0.5} & 29.8\%\footnotesize{\color{gray} $\pm$1.3} & \textbf{75.1}\footnotesize{\color{gray} $\pm$0.0} & 19.9\%\footnotesize{\color{gray} $\pm$1.2} \\
\texttt{Quick Sort} (const.\ $\alpha$) \textbf{\textcolor{blqs}{[QS]}} & \textbf{61.9}\footnotesize{\color{gray} $\pm$0.1} & 15.0\%\footnotesize{\color{gray} $\pm$0.7} & 58.4\footnotesize{\color{gray} $\pm$0.3} & 29.7\%\footnotesize{\color{gray} $\pm$0.1} & 74.9\footnotesize{\color{gray} $\pm$0.1} & 11.5\%\footnotesize{\color{gray} $\pm$0.8} \\
\texttt{Quick Sort} (doubling $\alpha$) \textbf{\textcolor{blqsd}{[QSD]}} & 61.3\footnotesize{\color{gray} $\pm$0.1} & 15.9\%\footnotesize{\color{gray} $\pm$0.4} & 58.4\footnotesize{\color{gray} $\pm$0.2} & 31.3\%\footnotesize{\color{gray} $\pm$1.1} & 75.0\footnotesize{\color{gray} $\pm$0.1} & 15.4\%\footnotesize{\color{gray} $\pm$0.2} \\
\texttt{Merge Sort} (const.\ $\alpha$) \textbf{\textcolor{blms}{[MS]}} & 60.8\footnotesize{\color{gray} $\pm$0.2} & 14.4\%\footnotesize{\color{gray} $\pm$0.5} & 59.2\footnotesize{\color{gray} $\pm$0.2} & 28.1\%\footnotesize{\color{gray} $\pm$0.8} & 75.0\footnotesize{\color{gray} $\pm$0.1} & 12.2\%\footnotesize{\color{gray} $\pm$0.8} \\
\texttt{Quick Sort} (adaptive $\alpha$) \textbf{\textcolor{blqsa}{[QSA]}} (Ours) & 61.0\footnotesize{\color{gray} $\pm$0.3} & \textbf{9.7\%}\footnotesize{\color{gray} $\pm$0.8} &
\textbf{59.3}\footnotesize{\color{gray} $\pm$0.1} & 21.8\%\footnotesize{\color{gray} $\pm$1.3} & 74.7\footnotesize{\color{gray} $\pm$0.2} & \textbf{6.8\%}\footnotesize{\color{gray}$\pm$0.5} \\
\texttt{Merge Sort} (adaptive $\alpha$) \textbf{\textcolor{blmsa}{[MSA]}} (Ours) & 60.1\footnotesize{\color{gray} $\pm$0.5} & \textbf{9.7\%}\footnotesize{\color{gray} $\pm$0.8} & 58.3\footnotesize{\color{gray} $\pm$0.4} & \textbf{21.7\%}\footnotesize{\color{gray} $\pm$1.3} & 74.5\footnotesize{\color{gray} $\pm$0.3} & 7.6\%\footnotesize{\color{gray} $\pm$0.9} \\
\midrule
\textcolor{gray}{\texttt{Expected (MC)} \textbf{\textcolor{blex}{[EX]}}} & \textcolor{gray}{50.0\footnotesize{ $\pm$0.0}} & \textcolor{gray}{-} &
\textcolor{gray}{50.0\footnotesize{ $\pm$0.0}} & \textcolor{gray}{-} & \textcolor{gray}{50.1\footnotesize{ $\pm$0.1}} & \textcolor{gray}{-} \\
\textcolor{gray}{\texttt{Bradley-Terry} \textbf{\textcolor{blbt}{[BT]}}} & \textcolor{gray}{62.6\footnotesize{ $\pm$0.0}} & \textcolor{gray}{100\%\footnotesize{ $\pm$0.0}} &
\textcolor{gray}{59.8\footnotesize{ $\pm$0.0}} & \textcolor{gray}{100\%\footnotesize{ $\pm$0.0}} & \textcolor{gray}{75.0\footnotesize{ $\pm$0.0}} &
\textcolor{gray}{100\%\footnotesize{$\pm$0.0}} \\
\textcolor{gray}{\texttt{HodgeRank} \textbf{\textcolor{blhg}{[HG]}}} & \textcolor{gray}{62.5\footnotesize{ $\pm$0.0}} & \textcolor{gray}{100\%\footnotesize{ $\pm$0.0}} &
\textcolor{gray}{59.8\footnotesize{ $\pm$0.0}} & \textcolor{gray}{100\%\footnotesize{ $\pm$0.0}} & \textcolor{gray}{75.0\footnotesize{ $\pm$0.0}} &
\textcolor{gray}{100\%\footnotesize{
$\pm$0.0}} \\
\textcolor{gray}{\texttt{Oracle} \textbf{\textcolor{bloe}{[OE]}}} & \textcolor{gray}{63.7\footnotesize{ $\pm$0.0}} & \textcolor{gray}{100\%\footnotesize{ $\pm$0.0}} &
\textcolor{gray}{60.8\footnotesize{ $\pm$0.0}} & \textcolor{gray}{100\%\footnotesize{ $\pm$0.0}} & \textcolor{gray}{75.2\footnotesize{ $\pm$0.0}} &
\textcolor{gray}{100\%\footnotesize{
$\pm$0.0}} \\
\bottomrule
\end{tabular}
}
\end{table}

\noindent\textbf{Metrics}.  Evaluating LLM rankings is challenging because we lack access to a gold ranking and the stochastic nature of pairwise evaluation introduces additional uncertainty. Moreover, when we construct a directed preference graph with LLMs as nodes, adding edges based on pairwise evaluation outcomes, we find many strongly connected components, indicating that a single optimal linear ordering may not exist (Fig.~\ref{fig:dag1}, \ref{fig:dag2}). Therefore, we evaluate the quality of a predicted ranking ($\sigma$) by computing its agreement with the pairwise data, $Q(\sigma)$ (we report the fraction in Eq.~\ref{eq:ranking-quality} as percentages). For sample
efficiency, we report the fraction of total pairwise evaluations used to construct the ranking. We execute all algorithms 3 times and use $\alpha=0.05$ for pairwise comparison calls.\\[-0.5em]

\noindent\textbf{Results}.
Table~\ref{tab:ranking-results} summarizes the ranking results.  All sorting-based methods achieve quality close to the offline full-data baselines (\textbf{\textcolor{blbt}{[BT]}}, \textbf{\textcolor{blhg}{[HG]}}), trailing them by less than 1\% on average, while using only a fraction of the total evaluations. In contrast, \textbf{\textcolor{ar}{[AR]}} requires 52-73\% of the evaluations yet yields the lowest quality on every dataset. Plugging Active Testing into sorting (\textbf{\textcolor{atq}{[ATQ]}}, \textbf{\textcolor{atm}{[ATM]}}) achieves
competitive quality but consumes up to $\sim$3x more samples than the best performing method, \textbf{\textcolor{blmsa}{[MSA]}}. Our adaptive sorting variants (\textbf{\textcolor{blqsa}{[QSA]}}, \textbf{\textcolor{blmsa}{[MSA]}}) are the two most sample-efficient
methods across all settings, reducing the sample budget by up to $\sim$40\% compared to their constant error allocated counterparts while achieving similar quality. The ranking quality obtained by the adaptive variants is also comparable to the offline methods that use all evaluations. These results show that our framework produces high quality rankings using only a fraction of the total evaluation budget.

\section{Conclusion}

We propose a sample-efficient LLM ranking framework that optimizes both LLM-pair selection and prompt selection for evaluation. Our pairwise evaluator combines e-process-based hypothesis testing with active sampling. We theoretically analyze how variation in prompt preferences can accelerate evidence accumulation and reduce evaluation cost. An LLM-driven sampling strategy makes this approach practical without access to latent preference probabilities. We integrate the pairwise procedure into comparison-based sorting algorithm, and derive an error allocation rule that balances evaluation cost against direct comparison errors. Experiments on synthetic and real-world preference data demonstrate the potential of active prompt selection to improve pairwise evaluation efficiency. Future work can analyze how errors in predicted prompt preferences affect evaluation cost and how reliably the final ranking reflects human judgments when pairwise preferences conflict.

\bibliographystyle{plainnat}
\bibliography{main}

\clearpage
\appendix
\clearpage

\section{Analysis of Pairwise Evaluation}
\label{app:pairwise-proofs}

The proofs follow the same progression as the method. We first verify error control and derive the directional proposals. We then prove the general sample-cost bound and show how the oracle mixture satisfies its growth condition. The concentration and central limit results build on this same argument. Finally, we quantify the effects of prediction error, learning the bet, and shortlisting, before discussing other fixed bets and a deterministic-label boundary case.

The central chain of implications is: strict drift of the evidence ratio, geometric decay of the wrong-direction weight, summable KL loss, and then the leading stopping-cost bound. Appendix~\ref{app:mixture-proof} proves this chain. Appendix~\ref{app:adaptive-proof} supplies the general growth-to-cost step rather than an additional assumption about the oracle. The learned-policy arguments replace the summable oracle loss by controlled prediction, coverage, and betting losses.

Throughout, the prompt population is finite, $p(x)>0$, and proposals and bets are predictable. The structure of the proposal distribution~\ref{eq:mixture-proposal} (mixing with uniform distribution) ensures full support for the growth and infinite-horizon cost analysis. All logarithms are natural. In sample-cost proofs we label the models so that $\mu>1/2$, and use
\[
B=\log(1/\alpha),\qquad S_t=\log E_t^+.
\]
The negative-direction case follows by exchanging labels. When an argument uses an infinite observation sequence, we continue the sampling policy beyond its stopping time. This continuation does not change the procedure before termination and lets us compare different thresholds on the same sample path.

\subsection{Validity of the pairwise evaluator}
\label{app:validity-proof}

\begin{proof}[Proof of Lemma~\ref{lem:valid_eprocess}]
The only inputs here are the fixed conditional label law, predictability, and full support. Prediction accuracy is not used. Full support lets the importance weights remove the proposal on every target prompt; predictability lets us hold those weights and the bet fixed inside the conditional expectation. Equation~\ref{eq:sequential-model} gives
\begin{align*}
\mathbb E\!\left[\frac{p(X_t)}{q_t(X_t)}e_t^+(Y_t)\,\middle|\,\mathcal F_{t-1}\right]
&=\sum_xp(x)\{2\nu_t\pi(x)+2(1-\nu_t)(1-\pi(x))\}\\
&=1+4(\nu_t-1/2)(\mu-1/2).
\end{align*}
Under $H_0^+$, the last expression is at most one. Multiplying by the nonnegative, $\mathcal F_{t-1}$-measurable $E_{t-1}^+$ proves the supermartingale property. Positivity follows from $p/q_t>0$ and $1/2\le\nu_t<1$. For the negative direction the same calculation gives
\[
\mathbb E\!\left[\frac{p(X_t)}{q_t(X_t)}e_t^-(Y_t)\,\middle|\,\mathcal F_{t-1}\right]
=1-4(\nu_t-1/2)(\mu-1/2)\le1
\]
under $H_0^-$.

Now suppose $\mu>1/2$. An incorrect declaration requires the negative process to cross $1/\alpha$, because that is the process supporting the inferior model. Ville's inequality~\citep{ville1939etude,ramdas2025hypothesis} therefore yields
\[
\mathbb P\{\text{an incorrect declaration}\}
\le\mathbb P\{\sup_{t\ge0}E_t^-\ge1/\alpha\}\le\alpha.
\]
The event inclusion also holds for simultaneous crossings, regardless of the fixed tie-breaking rule. Exchange the directions when $\mu<1/2$.

At $\mu=1/2$, both processes are supermartingales. At threshold $2/\alpha$, each crossing probability is at most $\alpha/2$, so their union has probability at most $\alpha$. This is an error statement, not a finite-sample termination guarantee at a tie.
\end{proof}

\paragraph{Tie outcomes ($Y_t\in\{0,1/2,1\}$).}
If human judgments include ties scored as $1/2$, or more generally $Y_t\in[0,1]$ satisfying $\mathbb E[Y_t\mid\mathcal F_{t-1},X_t=x]=\pi(x)\in[0,1]$, linearity of $y\mapsto 1\pm\lambda_t(y-1/2)$ still gives $\mathbb E[e_t^+(Y_t)\mid\mathcal F_{t-1},X_t=x]=2\nu_t\pi(x)+2(1-\nu_t)(1-\pi(x))$ (and symmetrically for $e_t^-$), so Lemma~\ref{lem:valid_eprocess} holds. Moreover, when $Y_t\in\{0,1/2,1\}$, concavity of $\log$ on $[2(1-\nu_t),2\nu_t]$ gives $\mathbb E[\log e_t^+(Y_t)\mid\mathcal F_{t-1},X_t=x]\ge\pi(x)\log(2\nu_t)+(1-\pi(x))\log(2(1-\nu_t))$ (the binary-label rate $h_t^+(x)$ in Lemma~\ref{lem:local_epower_process}), while for $\ell_t:=\log(\nu_t/(1-\nu_t))$, $e_t^+(Y_t)/e_t^-(Y_t)=e^{\ell_t(2Y_t-1)}$ holds identically at all three values $Y_t\in\{0,1/2,1\}$ (and $e^{-c(2Y_t-1)}\le Y_t e^{-c}+(1-Y_t)e^c$ for all $c\in\mathbb R$ by convexity, including $c=\theta\ell$ in Appendix~\ref{app:mixture-proof}); applying the bounded-increment martingale argument of Theorem~\ref{thm:adaptive-sample} to $\tilde G_t:=\mathbb E[\log(E_t^+/E_{t-1}^+)\mid\mathcal F_{t-1}]\ge G_t^+(q_t)$ preserves all expected-cost and upper-tail upper bounds in Theorem~\ref{thm:sample_complexity}, Corollary~\ref{cor:llm-sample}, and Proposition~\ref{prop:general-mixture}.

\subsection{Local e-power and the directional oracle}
\label{app:oracle-proof}

\begin{definition}[Local e-power]
\label{def:local}
For direction $s\in\{+,-\}$, the local e-power of prompt $x$ is
\begin{equation}
h_t^s(x)=\mathbb E[\log e_t^s(Y_t)\mid\mathcal F_{t-1},X_t=x].
\end{equation}
\end{definition}

\begin{lemma}
\label{lem:local_epower_process}
For $\nu_t\in[1/2,1)$, the local e-powers of the factors in Eq.~\ref{eqn:eprocess} are
\begin{align*}
h_t^+(x)&=D_{\mathrm{KL}}(\pi(x)\|1/2)-D_{\mathrm{KL}}(\pi(x)\|\nu_t),\\
h_t^-(x)&=D_{\mathrm{KL}}(\pi(x)\|1/2)-D_{\mathrm{KL}}(\pi(x)\|1-\nu_t).
\end{align*}
\end{lemma}

\begin{lemma}[Directional oracle]
\label{lemma:optimal}
For either direction and every proposal $q$,
\[
G_t^s(q)=G_t^{s,\star}-D_{\mathrm{KL}}(q\|q_t^{s,\star}).
\]
Consequently, $q_t^{s,\star}$ uniquely maximizes $G_t^s$.
\end{lemma}

\begin{proof}[Proof of Lemmas~\ref{lem:local_epower_process} and~\ref{lemma:optimal}]
We compute the unweighted contribution first, then include the importance weight. This order explains why the proposal maximizes a reward minus a KL cost rather than local e-power alone.

\noindent\textbf{Step 1: compute a prompt's local contribution.}
For Bernoulli probabilities $a\in[0,1]$ and $b\in(0,1)$, expanding the two divergences gives
\[
D_{\mathrm{KL}}(a\|1/2)-D_{\mathrm{KL}}(a\|b)
=a\log(2b)+(1-a)\log(2(1-b)).
\]
The convention $0\log0=0$ covers the endpoints. Substituting $a=\pi(x)$ and $b=\nu_t$ or $1-\nu_t$ proves
\begin{align}
h_t^+(x)&=D_{\mathrm{KL}}(\pi(x)\|1/2)-D_{\mathrm{KL}}(\pi(x)\|\nu_t),\nonumber\\
h_t^-(x)&=D_{\mathrm{KL}}(\pi(x)\|1/2)-D_{\mathrm{KL}}(\pi(x)\|1-\nu_t).
\label{eq:local-power}
\end{align}

\noindent\textbf{Step 2: include the sampling cost.}
Under proposal $q$, the complete log factor has conditional expectation
\[
\sum_{x:q(x)>0}q(x)\left\{h_t^s(x)+\log\frac{p(x)}{q(x)}\right\}
=\sum_xq(x)h_t^s(x)-D_{\mathrm{KL}}(q\|p).
\]
The order of the KL arguments matters: the expectation is under $q$, the distribution actually used to draw the prompt.

\noindent\textbf{Step 3: identify the optimizer and the exact loss.}
The definition of $q_t^{s,\star}$ implies
\[
\log\frac{q_t^{s,\star}(x)}{p(x)}=h_t^s(x)-G_t^{s,\star}.
\]
Substitution into the divergence gives
\begin{align*}
D_{\mathrm{KL}}(q\|q_t^{s,\star})
&=D_{\mathrm{KL}}(q\|p)-\sum_xq(x)h_t^s(x)+G_t^{s,\star}\\
&=G_t^{s,\star}-G_t^s(q).
\end{align*}
The divergence is nonnegative and equals zero only when $q=q_t^{s,\star}$. This proves both optimality and uniqueness.

Finally,
\[
h_t^+(x)=\log(2(1-\nu_t))+\pi(x)\log\frac{\nu_t}{1-\nu_t},
\]
and the negative expression replaces the constant by $\log(2\nu_t)$ and reverses the sign of the second term. Constants independent of $x$ cancel on normalization, giving Eq.~\ref{eq:directional-tilts}.
\end{proof}

For reference, the matched target-sampling bet and the estimated mixture are
\begin{equation}
\nu^\star=\max\{\mu,1-\mu\},\qquad \mu\in(0,1),
\label{eq:optimal-bet}
\end{equation}
\begin{equation}
\widehat q_t=w_{t-1}\widehat q_t^++(1-w_{t-1})\widehat q_t^-.
\label{eq:estimated-mixture}
\end{equation}
Neither matching the target bet nor knowing the preferences is needed for the validity lemma.

\begin{figure}[t!]
\centering
\includegraphics[width=0.42\linewidth,keepaspectratio]{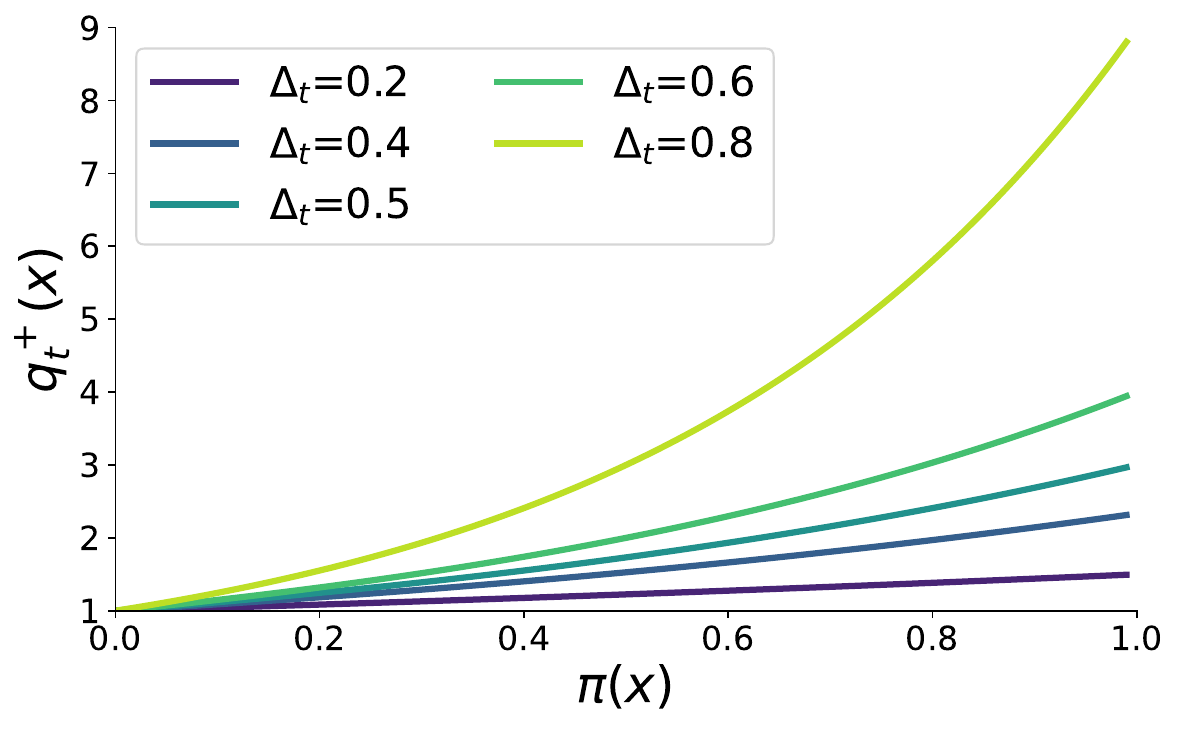}
\includegraphics[width=0.42\linewidth,keepaspectratio]{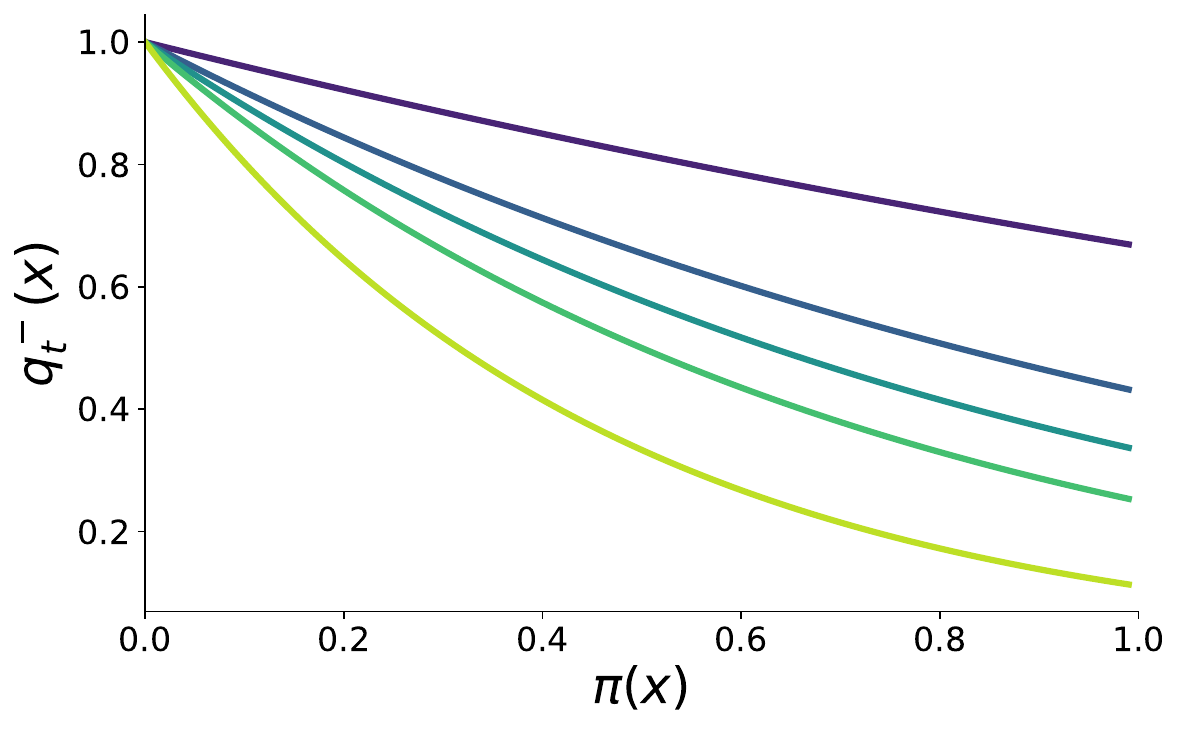}
\caption{\textbf{Directional prompt selection.} Unnormalized tilt factors increase with the preference probability for the positive direction and decrease for the negative direction. The curves are indexed by bet strength $\Delta_t=2\nu_t-1$. The normalized proposals also include the target weight $p(x)$, as in Eq.~\ref{eq:directional-tilts}.}
\label{fig:q}
\end{figure}

\subsection{Sample cost from cumulative evidence growth}
\label{app:adaptive-proof}

The conditional growth defined in Eq.~\ref{eq:directional-growth} obeys
\begin{equation}
\mathbb E[\log(E_t^+/E_{t-1}^+)\mid\mathcal F_{t-1}]
=G_t^+(q_t)=G_t^{+,\star}-D_{\mathrm{KL}}(q_t\|q_t^{+,\star}).
\label{eq:actual-growth}
\end{equation}
The next theorem converts cumulative growth into stopping cost. Its assumptions must be established for the policy under study; they do not follow from error validity alone.

\begin{theorem}[Sample complexity under adaptive sampling]
\label{thm:adaptive-sample}
Suppose $\mu>1/2$, $q_t\ge\kappa p$ for fixed $\kappa>0$, and $1/2\le\nu_t\le\bar\nu<1$. Let $b$ be a finite bound on the absolute log increment,
\[
\left|\log(E_t^+/E_{t-1}^+)\right|\le b.
\]
Such a bound exists under these conditions. Extend the policy beyond termination, and suppose that, for some $G>0$ and nonnegative $R$ with $\mathbb E[R]<\infty$,
\begin{equation}
\sum_{u=1}^tG_u^+(q_u)\ge tG-R
\qquad\text{for every }t\ge1\quad\text{almost surely}.
\label{eq:pathwise-growth}
\end{equation}
Then  for every $\alpha\in(0,1)$ with $B=\log(1/\alpha)$,
\begin{equation}
\mathbb E[\tau_\alpha]\le\frac{B+b+\mathbb E[R]}{G}.
\label{eq:general-expected-cost}
\end{equation}
For every $r\ge0$ and integer $t$ with $tG>B+r$,
\begin{equation}
\mathbb P(\tau_\alpha>t)
\le\mathbb P(R>r)
+\exp\!\left\{-\frac{(tG-B-r)^2}{8b^2t}\right\}.
\label{eq:general-cost-tail}
\end{equation}
The policy and the bounds $b,G,R$ are taken uniformly over $\alpha$.
\end{theorem}

\begin{proof}[Proof of Theorem~\ref{thm:adaptive-sample}]
We separate the proof into the predictable part of log evidence and its random fluctuations. The growth assumption controls the first part; bounded increments control the second. The statement is conditional on the supplied growth bound; no step of this proof establishes that bound for an arbitrary predictor. The subsequent oracle and approximation proofs provide it for the policies they analyze.

\noindent\textbf{Step 1: Bound increments and isolate a martingale.}
Both directional factors lie in $[2(1-\bar\nu),2\bar\nu]$. Since $q_t(x)\le1$ and $q_t(x)\ge\kappa p(x)$,
\begin{equation}
\log\{2(1-\bar\nu)\min_xp(x)\}
\le S_t-S_{t-1}\le\log(2\bar\nu/\kappa).
\label{eq:increment-range}
\end{equation}
Thus one valid choice in the theorem is
\[
b=\max\left\{
\log\frac1{2(1-\bar\nu)\min_xp(x)},
\log\frac{2\bar\nu}{\kappa}
\right\}.
\]
The first term is nonnegative because $\bar\nu>1/2$ and $\min_xp(x)\le1$. By Eq.~\ref{eq:actual-growth},
\begin{equation}
S_t=\sum_{u=1}^tG_u^+(q_u)+M_t,
\qquad
M_t:=\sum_{u=1}^t\{S_u-S_{u-1}-G_u^+(q_u)\}.
\label{eq:log-decomposition}
\end{equation}
The process $M_t$ is a martingale starting at zero. Each increment has absolute value at most $2b$, because both the observed increment and its conditional mean lie in $[-b,b]$.

\noindent\textbf{Step 2: Bound expected cost without assuming it is finite.}
We cannot apply a stopping argument at $\tau_\alpha$ before proving its integrability. Instead, fix an integer $m$ and use the bounded stopping time $T_m=\tau_\alpha\wedge m$. The event $\{T_m\ge t\}$ is known at time $t-1$, so
\begin{align*}
\mathbb E[S_{T_m}]
&=\sum_{t=1}^m\mathbb E[\mathbf 1_{\{T_m\ge t\}}(S_t-S_{t-1})]\\
&=\mathbb E\!\left[\sum_{t=1}^{T_m}G_t^+(q_t)\right]
\ge G\mathbb E[T_m]-\mathbb E[R].
\end{align*}
The last inequality applies the pathwise growth assumption at the random integer $T_m$; the same assumption holds at every integer on a common probability-one event.

Before termination, both log processes are below $B$. At termination the correct process can increase by at most $b$, whether or not it is the process that causes stopping. Therefore $S_{T_m}\le B+b$ in every case. Combining the two bounds gives
\[
G\mathbb E[T_m]\le B+b+\mathbb E[R].
\]
As $m\to\infty$, monotone convergence proves Eq.~\ref{eq:general-expected-cost}. In particular, $\tau_\alpha$ is finite almost surely.

\noindent\textbf{Step 3: Bound the probability of taking too many samples.}
On $\{\tau_\alpha>t,R\le r\}$, the correct log evidence has not reached $B$, while its cumulative conditional growth is at least $tG-r$. Equation~\ref{eq:log-decomposition} therefore implies
\[
M_t<-(tG-B-r).
\]
For $tG>B+r$, the Azuma--Hoeffding inequality for the bounded martingale differences~\citep{azuma1967weighted,hoeffding1963probability} yields
\[
\mathbb P\{M_t\le-(tG-B-r)\}
\le\exp\!\left\{-\frac{(tG-B-r)^2}{8b^2t}\right\}.
\]
Adding the probability of $R>r$ proves Eq.~\ref{eq:general-cost-tail}. Independence between the deficit and the martingale is not required.
\end{proof}

\noindent\textbf{An explicit sample budget.}
Given $\beta\in(0,1)$, choose a deterministic $r\ge0$ satisfying $\mathbb P(R>r)\le\beta/2$. Then a valid upper quantile is
\begin{equation}
U(\alpha,\beta)=\left\lceil
\left\{\frac{b\sqrt{2\log(2/\beta)}
+\sqrt{2b^2\log(2/\beta)+G(B+r)}}{G}\right\}^{\!2}
\right\rceil.
\label{eq:general-cost-quantile}
\end{equation}
Indeed, the expression before rounding solves
$Gt-B-r=2b\sqrt{2t\log(2/\beta)}$. For larger $t$, the martingale term in Eq.~\ref{eq:general-cost-tail} is at most $\beta/2$. The deficit contributes the other $\beta/2$.

A convenient, slightly larger expression is
\begin{equation}
\left\lceil\frac{B+r}{G}
+\frac{2b\sqrt{2(B+r)\log(2/\beta)}}{G^{3/2}}
+\frac{8b^2\log(2/\beta)}{G^2}\right\rceil.
\label{eq:general-cost-simple-quantile}
\end{equation}
It follows from Eq.~\ref{eq:general-cost-quantile} using $\sqrt{x+y}\le\sqrt x+\sqrt y$ after expanding the square. This form makes the leading threshold term, the square-root fluctuation term, and the confidence term explicit.

For sharper constants, the expected-cost numerator can use the upper endpoint $\log(2\bar\nu/\kappa)$ in Eq.~\ref{eq:increment-range} instead of $b$. Likewise, the martingale-increment bound $2b$ can be replaced by the width of that interval. The single constant $b$ in the theorem avoids carrying both endpoints through the statements.

\noindent\textbf{An expectation-only variant.}
A bound on expected cost does not need a pathwise deficit or a tail bound for that deficit. It suffices to control the expected growth accumulated up to bounded stopping times.
\begin{proposition}[Sample complexity from expected cumulative growth]
\label{prop:adaptive-growth}
Under the bounded-increment assumptions above, suppose that for constants $G>0$ and $C\ge0$, independent of $\alpha$,
\begin{equation}
\mathbb E\!\left[\sum_{t=1}^{\tau_\alpha\wedge m}G_t^+(q_t)\right]
\ge G\mathbb E[\tau_\alpha\wedge m]-C
\qquad\text{for every }m\ge1.
\label{eq:growth-deficit}
\end{equation}
Then,
\begin{equation}
\mathbb E[\tau_\alpha]\le\frac{B+b+C}{G}.
\label{eq:adaptive-sample-bound}
\end{equation}
\end{proposition}
\begin{proof}
Step 2 above gives $G\mathbb E[T_m]-C\le\mathbb E[S_{T_m}]\le B+b$. Monotone convergence completes the argument.
\end{proof}
The expectation-only condition does not itself supply an exponential upper tail. For that conclusion, one must also control realized cumulative growth, as in Theorem~\ref{thm:adaptive-sample}.

\subsection{Expected sample cost of the oracle mixture}
\label{app:mixture-proof}

We prove the expected-cost and asymptotic claims in Theorem~\ref{thm:sample_complexity}; the upper-tail claim follows in the next subsection. The key intermediate conclusion is
\begin{equation}
\sum_{t\ge1}\mathbb E[D_{\mathrm{KL}}(q_t\|q^+)]<\infty.
\label{eq:mixture-regret}
\end{equation}

\begin{proof}[Proof of Theorem~\ref{thm:sample_complexity}]
The proof has three ingredients. First, even the opposite proposal favors the correct label on average. Second, this makes the expected weight on that proposal decrease geometrically. Third, the resulting total loss in log growth is finite. This verifies the general sample-cost condition with rate $G^\star$. A final growth argument gives the matching asymptotic lower bound.

For this proof, write
\[
\ell=\log\frac\mu{1-\mu},\qquad
\eta=\mathbb E_{q^-}[\pi(X)].
\]
The fixed matched proposals are mixed as $q_t=w_{t-1}q^++(1-w_{t-1})q^-$.

\noindent\textbf{Step 1: The opposite proposal still favors the correct label.}
Since $q^-(x)\propto p(x)e^{-\ell\pi(x)}$, proving $\eta>1/2$ amounts to proving that the numerator below is positive:
\[
\eta-\frac12
=\frac{\mathbb E_p[(\pi(X)-1/2)e^{-\ell\pi(X)}]}
{\mathbb E_p[e^{-\ell\pi(X)}]}.
\]
For $z\in[0,1]$,
\begin{equation}
(z-1/2)e^{-\ell z}
\ge-\frac12+\frac{1+e^{-\ell}}2z,
\label{eq:opposite-mean-chord}
\end{equation}
with equality only at $z=0$ or $z=1$. To verify it, for $z\le1/2$ use $(z-1/2)e^{-\ell z}\ge z-1/2$. Its difference from the right side is $(1-e^{-\ell})z/2$. For $z\ge1/2$, use $(z-1/2)e^{-\ell z}\ge e^{-\ell}(z-1/2)$; the corresponding difference is $(1-e^{-\ell})(1-z)/2$.

Taking expectations under $p$ and using $e^{-\ell}=(1-\mu)/\mu$ gives
\[
\mathbb E_p[(\pi(X)-1/2)e^{-\ell\pi(X)}]
\ge-\frac12+\frac{1+e^{-\ell}}2\mu=0.
\]
At least one prompt has an interior preference probability and positive target mass, so the inequality is strict. This is where the interior-preference assumption is used.

The decreasing tilt also gives $\eta\le\mu$. One direct verification is
\[
2\operatorname{Cov}_p(\pi(X),e^{-\ell\pi(X)})
=\sum_{x,z}p(x)p(z)(\pi(x)-\pi(z))
(e^{-\ell\pi(x)}-e^{-\ell\pi(z)})\le0.
\]
Using the increasing tilt instead gives $\mathbb E_{q^+}[\pi(X)]\ge\mu$. Consequently,
\begin{equation}
\mathbb E[Y_t\mid\mathcal F_{t-1}]
=w_{t-1}\mathbb E_{q^+}[\pi(X)]+(1-w_{t-1})\eta
\ge\eta>1/2.
\label{eq:mixture-label-drift}
\end{equation}
The mixture therefore receives evidence favoring the correct direction even when it puts most of its weight on $q^-$.

\noindent\textbf{Step 2: Bound the weight on the wrong direction.}
Define the log evidence ratio
\[
H_t=\log(E_t^+/E_t^-),\qquad H_0=0.
\]
The same importance factor appears in both updates and cancels from the ratio. Thus
\begin{equation}
H_t-H_{t-1}=\ell(2Y_t-1),
\qquad 1-w_t=\frac1{1+e^{H_t}}.
\label{eq:ratio-increment}
\end{equation}
Positive drift alone suggests that $H_t$ increases; an exponential-moment calculation gives the quantitative bound we need. Choose
\[
\theta=\frac{\log(\eta/(1-\eta))}{2\ell}\in(0,1/2],
\qquad \rho=2\sqrt{\eta(1-\eta)}<1.
\]
The range of $\theta$ follows from $1/2<\eta\le\mu<1$. Equation~\ref{eq:mixture-label-drift} implies
\begin{align*}
\mathbb E[e^{-\theta(H_t-H_{t-1})}\mid\mathcal F_{t-1}]
&\le\eta e^{-\theta\ell}+(1-\eta)e^{\theta\ell}
=\rho.
\end{align*}
Iterating conditional expectations yields $\mathbb E[e^{-\theta H_t}]\le\rho^t$. For $0\le\theta\le1$ and every real $h$, $(1+e^h)^{-1}\le e^{-\theta h}$. Hence
\begin{equation}
\mathbb E[1-w_t]\le\rho^t,
\qquad
\sum_{t\ge0}\mathbb E[1-w_t]\le\frac1{1-\rho}.
\label{eq:mixture-weight-sum}
\end{equation}

\noindent\textbf{Step 3: Convert the weight bound into a growth bound.}
Let $D=D_{\mathrm{KL}}(q^-\|q^+)$. This is finite because both proposals have full support. By the directional-oracle identity and convexity of KL divergence in its first argument,
\begin{equation}
G_t^+(q_t)=G^\star-D_{\mathrm{KL}}(q_t\|q^+),
\qquad
D_{\mathrm{KL}}(q_t\|q^+)\le D(1-w_{t-1}).
\label{eq:mixture-step-regret}
\end{equation}
Define the total loss
\begin{equation}
R=\sum_{t\ge1}D_{\mathrm{KL}}(q_t\|q^+),
\qquad
\mathbb E[R]\le\frac{D}{1-\rho}<\infty.
\label{eq:mixture-total-regret}
\end{equation}
This proves Eq.~\ref{eq:mixture-regret}, and verifies Eq.~\ref{eq:pathwise-growth} with $G=G^\star$.

For completeness, all increment bounds required by the general theorem hold here. The normalized tilts satisfy $q^\pm\ge e^{-\ell}p$, so their mixture does too. The largest correct log increment is at most $\log(2\mu)+\ell$; a finite lower bound follows from $p(x)>0$ and the finite population. Also, directional optimality gives $G^\star\ge D_{\mathrm{KL}}(\mu\|1/2)>0$. Applying the stopped-expectation argument with this upper increment gives the explicit bound
\begin{equation}
\mathbb E[\tau_\alpha]
\le\frac{B+\log(2\mu)+\ell+D/(1-\rho)}{G^\star}.
\label{eq:mixture-explicit-bound}
\end{equation}
The finite additive constant depends only on $p$ and $\pi$. This calculation applies to the first crossing of either process, not just to the correct process: before that crossing $S_t<B$, and its final increment is still bounded if the other process triggers stopping.

\noindent\textbf{Step 4: Identify the limiting sample cost.}
The martingale in Eq.~\ref{eq:log-decomposition} gives
\[
S_t=tG^\star-\sum_{u=1}^tD_{\mathrm{KL}}(q_u\|q^+)+M_t.
\]
Since $\mathbb E[R]<\infty$, the nonnegative total loss $R$ is finite almost surely. Moreover, bounded martingale increments and Azuma--Hoeffding give a summable bound on $\mathbb P(|M_t|>\varepsilon t)$ for each $\varepsilon>0$. The Borel--Cantelli lemma then gives $M_t/t\to0$ almost surely. Dividing the display by $t$ proves $S_t/t\to G^\star$ almost surely.

We must also rule out an earlier crossing by the opposite process at large thresholds. Ville's inequality gives
\[
\mathbb P\{\sup_{t\ge0}E_t^-\ge u\}\le u^{-1},\qquad u\ge1.
\]
Letting $u\to\infty$ shows that $\sup_t E_t^-<\infty$ almost surely. Therefore, on almost every infinite path, sufficiently large thresholds can be crossed only by $E_t^+$. For any $0<\varepsilon<G^\star$, the growth limit eventually places $S_t$ between $(G^\star-\varepsilon)t$ and $(G^\star+\varepsilon)t$. The finitely many earlier values cannot cross arbitrarily large thresholds. Comparing the hitting time with $B/(G^\star+\varepsilon)$ and $B/(G^\star-\varepsilon)$, then letting $\varepsilon\downarrow0$, yields
\[
\frac{\tau_\alpha}{B}\longrightarrow\frac1{G^\star}
\quad\text{almost surely as }\alpha\downarrow0.
\]
Fatou's lemma gives the corresponding lower bound on the limiting expected ratio. Equation~\ref{eq:mixture-explicit-bound} supplies the matching upper bound. This proves Eq.~\ref{eq:mixture-asymptotic} without assuming uniform integrability in advance.
\end{proof}

\subsection{Concentration for the oracle mixture}
\label{app:oracle-concentration}

\begin{corollary}[An upper bound on realized sample cost]
\label{cor:oracle-tail}
Under the oracle assumptions of Theorem~\ref{thm:sample_complexity}, there are finite constants $C_1,C_2,C_3$, depending only on $(p,\pi)$, such that, for every $\alpha,\beta\in(0,1)$,
\begin{equation}
\mathbb P\!\left\{\tau_\alpha>
\frac{B}{G^\star}+C_1\sqrt{B\log(2/\beta)}
+C_2\log(2/\beta)+C_3\right\}\le\beta.
\label{eq:oracle-cost-tail}
\end{equation}
\end{corollary}

The main-text bound Eq.~\ref{eq:cost-tail-main} follows by taking one constant at least as large as $C_1,C_2,C_3$ and the expectation constant.

The expected-cost proof controls the mean of the total growth loss. For a high-probability sample budget, we also need its tail. Geometric decay of the expected wrong-direction weight is enough because each round's loss is bounded. No independence is needed for this step.

\begin{lemma}[A sum with geometrically decaying means]
\label{lem:geometric-sum}
Suppose $0\le A_t\le C$ and $\mathbb E[A_t]\le CC_0\rho^t$ for $t\ge0$, where $C\ge0$, $C_0\ge1$, and $0<\rho<1$. Then
\[
\mathbb E\!\left[\sum_{t\ge0}A_t\right]\le\frac{CC_0}{1-\rho}<\infty,
\]
and, for every integer $m\ge0$,
\begin{equation}
\mathbb P\!\left\{\sum_{t\ge0}A_t>C(m+1)\right\}
\le\frac{C_0\rho^m}{1-\rho}.
\label{eq:geometric-sum-tail}
\end{equation}
\end{lemma}
\begin{proof}
Summing the nonnegative expectations proves the first assertion and almost-sure finiteness. If $C=0$, the sum is zero. Otherwise split it after its first $m$ terms. Those terms contribute at most $Cm$, while the remaining terms have expectation at most $CC_0\rho^m/(1-\rho)$. For the full sum to exceed $C(m+1)$, this tail must exceed $C$. Markov's inequality gives the stated bound.
\end{proof}

\begin{proof}[Proof of Corollary~\ref{cor:oracle-tail}]
Recall from Appendix~\ref{app:mixture-proof} that
\[
D=D_{\mathrm{KL}}(q^-\|q^+),\qquad
\rho=2\sqrt{\eta(1-\eta)}<1,
\]
and that the total growth loss satisfies
\[
R\le D\sum_{t\ge0}(1-w_t),\qquad\mathbb E[1-w_t]\le\rho^t.
\]
Apply Lemma~\ref{lem:geometric-sum} to $A_t=D(1-w_t)$. For $\beta\in(0,1)$, choose
\begin{equation}
r=D\left[1+\left\lceil
\frac{\log\{2/((1-\rho)\beta)\}}{\log(1/\rho)}
\right\rceil\right].
\label{eq:geometric-sum-quantile}
\end{equation}
Then $\mathbb P(R>r)\le\beta/2$. If $D=0$, simply take $r=0$.

The growth condition holds with $G=G^\star$, and the increment bound depends only on the fixed prompt population. Substituting this $r$ into Eq.~\ref{eq:general-cost-simple-quantile} gives an explicit sample budget. In particular, $r$ is at most a pair-dependent constant times $1+\log(2/\beta)$. Expanding the square-root term with $\sqrt{x+y}\le\sqrt x+\sqrt y$ therefore gives
\[
\frac{B}{G^\star}+C_1\sqrt{B\log(2/\beta)}
+C_2\log(2/\beta)+C_3.
\]
The constants absorb the bounded rounding term and are independent of $\alpha$ and $\beta$. This is Eq.~\ref{eq:oracle-cost-tail}.
\end{proof}

\subsection{A central limit theorem for pairwise sample cost}
\label{app:pairwise-clt}

The preceding bound controls unusually large sample costs at a fixed threshold. The next theorem instead describes typical fluctuations as the threshold increases. The same finite total growth loss is now negligible relative to the square-root scale.

\begin{theorem}[Pairwise stopping-time fluctuations]
\label{thm:pairwise-clt}
Under the matched-oracle assumptions of Theorem~\ref{thm:sample_complexity}, with $\mu>1/2$, let
\begin{equation}
v^\star=\left(\log\frac\mu{1-\mu}\right)^2
\mathbb E_{q^+}[\pi(X)(1-\pi(X))].
\label{eq:oracle-variance}
\end{equation}
Then $v^\star>0$ and, as $\alpha\downarrow0$,
\begin{equation}
\frac{\tau_\alpha-B/G^\star}{\sqrt B}
\ \Rightarrow\ \mathcal N\!\left(0,\frac{v^\star}{(G^\star)^3}\right).
\label{eq:pairwise-clt}
\end{equation}
For $\mu<1/2$, exchange labels and use the correct negative-direction proposal.
\end{theorem}

\begin{proof}
We first identify the limiting variance of log evidence. We then apply a martingale functional central limit theorem and translate the log-evidence limit into a hitting-time limit.

\noindent\textbf{Step 1: Identify the remaining randomness under the oracle.}
Under $q^+$, its normalizing tilt cancels the prompt-specific mean of the log factor:
\begin{equation}
\log\!\left\{\frac{p(X)}{q^+(X)}e^+(Y)\right\}
=G^\star+\ell(Y-\pi(X)),
\qquad\ell=\log\frac\mu{1-\mu}.
\label{eq:oracle-increment-identity}
\end{equation}
Indeed, $\log(p/q^+)=G^\star-h^+(X)$ and $\log e^+(Y)-h^+(X)=\ell(Y-\pi(X))$. Conditional on the prompt, the second term has mean zero and variance $\ell^2\pi(X)(1-\pi(X))$. Averaging over $q^+$ gives $v^\star$. It is positive because $\ell>0$ and some prompt with positive $q^+$ mass has an interior preference probability.

For the actual mixture, the decomposition from Eq.~\ref{eq:log-decomposition} is
\[
S_t=tG^\star-\sum_{u=1}^tD_{\mathrm{KL}}(q_u\|q^+)+M_t.
\]
Appendix~\ref{app:mixture-proof} shows that the entire loss sum is bounded by an almost-surely finite $R$. It also shows $\sum_t\mathbb E[1-w_t]<\infty$, which implies $w_t\to1$ and hence $q_t\to q^+$ almost surely.

\noindent\textbf{Step 2: Obtain a functional limit for the martingale.}
The conditional variance of a log increment (i.e., the conditional second moment of $M_t-M_{t-1}$) is a finite sum of continuous functions of $q_t(x)$. The floor $q_t\ge e^{-\ell}p$ keeps all denominators away from zero. Therefore convergence to $q^+$ implies
\[
\mathbb E[(M_t-M_{t-1})^2\mid\mathcal F_{t-1}]
\longrightarrow v^\star\quad\text{almost surely}.
\]
Averaging these conditional variances gives, for every $u\ge0$ as $T\to\infty$,
\[
\frac1T\sum_{t=1}^{\lfloor Tu\rfloor}
\mathbb E[(M_t-M_{t-1})^2\mid\mathcal F_{t-1}]
\longrightarrow uv^\star\quad\text{almost surely}.
\]
This verifies convergence of the predictable quadratic variation. Since the martingale increments are uniformly bounded ($|M_t-M_{t-1}|\le2b$), the scaled
increments satisfy the conditional Lindeberg condition, since $\mathbf 1_{\{|M_t-M_{t-1}|>\varepsilon\sqrt T\}}=0$ for all $T>4b^2/\varepsilon^2$ (moreover, both the scaled martingale's maximum squared jump and the maximum jump of its predictable quadratic variation are $O(T^{-1})$).
The martingale functional
central limit theorem \cite{billingsley1999convergence}
therefore applies in $D[0,\infty)$. Both bounds are deterministic and therefore also hold in expectation. The martingales are square-integrable. Consequently,
\begin{equation}
\left\{\frac{M_{\lfloor Tu\rfloor}}{\sqrt T}:u\ge0\right\}
\ \Rightarrow\ \{\sqrt{v^\star}W(u):u\ge0\},
\label{eq:pairwise-martingale-fclt}
\end{equation}
where $W$ is standard Brownian motion.

\noindent\textbf{Step 3: Convert log-evidence fluctuations into time fluctuations.}
First consider only the correct-process hitting time
\[
\tau_B^+=\inf\{t\ge1:S_t\ge B\}.
\]
The almost-sure growth limit proved in Appendix~\ref{app:mixture-proof} gives $\tau_B^+/B\to1/G^\star$. Since this limit is deterministic, it holds jointly with the functional limit above. Evaluation at a converging time is continuous at continuous paths~\citep{billingsley1999convergence}; hence
\[
\frac{M_{\tau_B^+}}{\sqrt B}
\ \Rightarrow\ \sqrt{v^\star}W(1/G^\star)
\sim\mathcal N(0,v^\star/G^\star).
\]
The overshoot is bounded: $0\le S_{\tau_B^+}-B\le b$ for a finite increment bound $b$. Rearranging the decomposition of $S_t$ at the hitting time gives
\begin{align*}
\frac{\tau_B^+-B/G^\star}{\sqrt B}
={}&-\frac{M_{\tau_B^+}}{G^\star\sqrt B}\\
&+\frac{\sum_{u=1}^{\tau_B^+}D_{\mathrm{KL}}(q_u\|q^+)
+S_{\tau_B^+}-B}{G^\star\sqrt B}.
\end{align*}
The second line tends to zero almost surely: its numerator is at most $R+b$. The first line converges to a centered normal with variance $v^\star/(G^\star)^3$, proving the desired limit for $\tau_B^+$.

\noindent\textbf{Step 4: Return to the implemented two-process rule.}
The two stopping times can differ only if the negative process also reaches $e^B$. Since it is a null supermartingale, Ville's inequality~\citep{ville1939etude} yields the following:
\[
\mathbb P\{\tau_\alpha\ne\tau_B^+\}
\le\mathbb P\{\sup_tE_t^-\ge e^B\}\le e^{-B}=\alpha.
\]
This probability vanishes, so the implemented stopping time has the same weak limit.
\end{proof}

The variance has two sources of scaling. About $B/G^\star$ judgments contribute martingale variance, and converting log-evidence error to time error divides by $G^\star$. This explains the denominator $(G^\star)^3$. A corresponding theorem for learned proposals would additionally need convergence of their conditional variances and growth errors negligible relative to $\sqrt B$; error validity alone does not imply such a limit.

\subsection{The gain from prompt heterogeneity}
\label{app:heterogeneity-proof}

Under target sampling, the matched bet gives log-growth rate $D_{\mathrm{KL}}(\mu\|1/2)$. The oracle rate decomposes as follows:
\begin{equation}
G^\star=D_{\mathrm{KL}}(\mu\|1/2)
+\log\mathbb E_{X\sim p}\!\left[
\left(\frac\mu{1-\mu}\right)^{\pi(X)-\mu}\right].
\label{eq:heterogeneity-bonus}
\end{equation}
The second term is nonnegative by Jensen's inequality and is strictly positive when preference probabilities vary across prompts. Let $\tau_{p,\alpha}$ denote the stopping time under $q_t=p$, using the same matched bet and two-process stopping rule.

\begin{corollary}[Gain from prompt heterogeneity]
\label{cor:heterogeneity-mixture}
Under the  assumptions of Theorem~\ref{thm:sample_complexity},
\begin{equation}
\lim_{\alpha\downarrow0}\frac{\mathbb E[\tau_\alpha]}{\mathbb E[\tau_{p,\alpha}]}
=\frac{D_{\mathrm{KL}}(\mu\|1/2)}{G^\star}\le1,
\label{eq:asymptotic-comparison}
\end{equation}
with strict inequality exactly when $\pi(X)$ is nonconstant under $p$.
\end{corollary}

The two-prompt population with preferences $0.9$ and $0.6$ therefore admits a strict asymptotic gain. If both probabilities were $0.75$, the target win rate would be unchanged, but both directional proposals would equal $p$ and the gain would disappear.

\begin{proof}[Proof of Corollary~\ref{cor:heterogeneity-mixture}]
We first identify the growth gain and then compare the two stopping costs.\\[-0.5em]

\noindent\textbf{Step 1: Separate the target rate from the gain.}
At the matched bet, the positive local e-power is
\[
h^+(x)=\log(2(1-\mu))+\ell\pi(x)
=D_{\mathrm{KL}}(\mu\|1/2)+\ell(\pi(x)-\mu).
\]
Taking its exponential moment under $p$ gives
\[
G^\star=D_{\mathrm{KL}}(\mu\|1/2)
+\log\mathbb E_p[e^{\ell(\pi(X)-\mu)}],
\]
which is Eq.~\ref{eq:heterogeneity-bonus}. The centered preference has mean zero. Jensen's inequality makes the second term nonnegative, and strict convexity makes it positive exactly when $\pi(X)$ is nonconstant under $p$.\\[-0.5em]

\noindent\textbf{Step 2: Evaluate the target-sampling benchmark.}
Under $q_t=p$, the correct log increment has constant conditional mean $D_{\mathrm{KL}}(\mu\|1/2)>0$ and upper bound $\log(2\mu)$. The bounded-stopping argument therefore gives
\[
\mathbb E[\tau_{p,\alpha}]
\le\frac{B+\log(2\mu)}{D_{\mathrm{KL}}(\mu\|1/2)}.
\]
As in Step 4 of Appendix~\ref{app:mixture-proof}, bounded martingale fluctuations give the almost-sure growth rate, and the wrong-direction supermartingale is bounded over the infinite path almost surely. Thus
\[
\frac{\tau_{p,\alpha}}B\longrightarrow
\frac1{D_{\mathrm{KL}}(\mu\|1/2)}\quad\text{almost surely}.
\]
Fatou's lemma and the displayed expectation bound give the same limit for the expected ratio. Combining this with Theorem~\ref{thm:sample_complexity} proves Eq.~\ref{eq:asymptotic-comparison} and its strictness condition.
\end{proof}

\subsection{Prediction error and proposal loss}
\label{app:prediction-proof}
We now verify the general growth condition for LLM-guided sampling. Let $\ell_t=\log(\nu_t/(1-\nu_t))$ for predictable $\nu_t\in[1/2,1)$ and predictable scores $\widehat\pi_{t-1}(x)\in[0,1]$, and define the population prediction-error variance
\[
\sigma_{t-1}^2:=\operatorname{Var}_{X\sim p}\!\bigl(\widehat\pi_{t-1}(X)-\pi(X)\bigr)
=\sum_{x\in\mathcal X}p(x)\Bigl(\widehat\pi_{t-1}(x)-\pi(x)-\sum_{z\in\mathcal X}p(z)\bigl(\widehat\pi_{t-1}(z)-\pi(z)\bigr)\Bigr)^{\!2}.
\]
Then the full-population estimated proposals satisfy
\begin{equation}
D_{\mathrm{KL}}(\widehat q_t^s\|q_t^{s,\star})\le\frac{e^{\ell_t}\ell_t^2\sigma_{t-1}^2}{2},
\qquad s\in\{+,-\}.
\label{eq:prediction-kl}
\end{equation}
Thus prediction error incurs a quadratic loss in the population variance of the estimated directional proposal (without requiring a uniform per-prompt error bound; constant calibration offsets cancel). The proof below derives this bound from an exact cumulant-interpolation identity. For the final mixture, convexity gives
\begin{align}
D_{\mathrm{KL}}(q_t\|q_t^{+,\star})
\le{}&\kappa D_{\mathrm{KL}}(p\|q_t^{+,\star})\nonumber\\
&+(1-\kappa)\left\{w_{t-1}\frac{e^{\ell_t}\ell_t^2\sigma_{t-1}^2}{2}
+(1-w_{t-1})D_{\mathrm{KL}}(\widehat q_t^-\|q_t^{+,\star})\right\}.
\label{eq:learned-mixture-loss}
\end{align}
The terms have distinct roles: exploration, prediction error in the correct component, and weight assigned to the opposite component. To turn this inequality into a sample bound, we must also show that the last weight decreases.

The estimated proposal is an exponential perturbation of the true proposal. Bounding this perturbation serves two different purposes: a KL bound controls evidence growth, while a label-mean bound ensures that the mixture still learns the correct direction.

\begin{lemma}[Perturbing a directional proposal]
\label{lem:prediction-loss}
Fix $\nu\in[1/2,1)$, and put $\ell=\log(\nu/(1-\nu))$. Let $q^s$ and $\widehat q^s$ be the true and estimated directional proposals at that bet for $\pi(x),\widehat\pi(x)\in[0,1]$, and let $\sigma:=\sqrt{\operatorname{Var}_{X\sim p}(\widehat\pi(X)-\pi(X))}\ge0$. Then
\begin{equation}
D_{\mathrm{KL}}(\widehat q^s\|q^s)\le\frac{e^{\ell}\ell^2\sigma^2}{2},
\qquad
\left|\mathbb E_{\widehat q^s}[\pi(X)]-\mathbb E_{q^s}[\pi(X)]\right|
\le\frac{e^{\ell/2}\ell\sigma}{2}.
\label{eq:prediction-perturbation}
\end{equation}
If the estimated proposal is normalized on a nonempty shortlist $C$, the same bounds hold relative to $q^s(\cdot\mid C)$ with $\sigma_C:=\sqrt{\operatorname{Var}_{X\sim p(\cdot\mid C)}(\widehat\pi(X)-\pi(X))}\ge0$ in place of $\sigma$. In particular,
\begin{equation}
D_{\mathrm{KL}}(\widehat q_C^s\|q^s)
\le\frac{e^{\ell}\ell^2\sigma_C^2}{2}+\log\frac1{q^s(C)}.
\label{eq:shortlist-kl-proof}
\end{equation}
The statements also hold conditionally for predictable bets, predictions, and shortlists.
\end{lemma}

\begin{proof}
We first control the exponential perturbation of the proposal, then its effect on growth and on the mean label. The shortlist result follows by conditioning the reference proposal.

\noindent\textbf{Step 1: Express prediction error as an exponential tilt.}
For the positive direction let $d(x)=\ell(\widehat\pi(x)-\pi(x))$; use its negative for the other direction. For $u\in[0,1]$, define the interpolated proposal
\[
q_u^s(x):=\frac{q^s(x)e^{ud(x)}}{\mathbb E_{q^s}[e^{ud(X)}]},
\qquad
\pi_u(x):=(1-u)\pi(x)+u\widehat\pi(x)\in[0,1],
\]
so $q_0^s=q^s$ and $q_1^s=\widehat q^s$. Since $\ell\ge0$ and $\pi_u(x)\in[0,1]$, for $s=+$ we have $e^{\ell\pi_u(x)}\le e^\ell$ and $\sum_zp(z)e^{\ell\pi_u(z)}\ge1$, so
\[
\frac{q_u^+(x)}{p(x)}
=\frac{e^{\ell\pi_u(x)}}{\sum_zp(z)e^{\ell\pi_u(z)}}
\le e^\ell.
\]
For $s=-$, multiplying numerator and denominator by $e^\ell$ and using $1-\pi_u(x)\in[0,1]$ likewise gives
\[
\frac{q_u^-(x)}{p(x)}
=\frac{e^{\ell(1-\pi_u(x))}}{\sum_zp(z)e^{\ell(1-\pi_u(z))}}
\le e^\ell.
\]
Hence $q_u^s(x)\le e^\ell p(x)$ for all $x\in\mathcal X$, $u\in[0,1]$, and $s\in\{+,-\}$. Consequently, for any function $g:\mathcal X\to\mathbb R$, with $m_p:=\mathbb E_p[g(X)]$,
\begin{align*}
\operatorname{Var}_{q_u^s}(g(X))
&=\min_{c\in\mathbb R}\mathbb E_{q_u^s}[(g(X)-c)^2]
\le\mathbb E_{q_u^s}[(g(X)-m_p)^2]\\
&\le e^\ell\mathbb E_p[(g(X)-m_p)^2]
=e^\ell\operatorname{Var}_p(g(X)).
\end{align*}

\noindent\textbf{Step 2: Bound the KL loss.}
Define $\Lambda(u):=\log\mathbb E_{q^s}[e^{ud(X)}]$ for $u\in[0,1]$. Then $\Lambda(0)=0$, differentiating the finite sum gives $\Lambda'(u)=\mathbb E_{q_u^s}[d(X)]$ and $\Lambda''(u)=\operatorname{Var}_{q_u^s}(d(X))=\ell^2\operatorname{Var}_{q_u^s}(\widehat\pi(X)-\pi(X))$, and $\log(\widehat q^s(x)/q^s(x))=d(x)-\Lambda(1)$. Integrating $\int_0^1u\Lambda''(u)\,du$ by parts and applying Step 1 yields the following:
\begin{align*}
D_{\mathrm{KL}}(\widehat q^s\|q^s)
&=\mathbb E_{\widehat q^s}[d(X)]-\Lambda(1)
=\Lambda'(1)-\Lambda(1)+\Lambda(0)
=\Bigl[u\Lambda'(u)\Bigr]_0^1-\int_0^1\Lambda'(u)\,du\\
&=\int_0^1u\Lambda''(u)\,du
=\ell^2\int_0^1u\operatorname{Var}_{q_u^s}(\widehat\pi(X)-\pi(X))\,du
\le e^\ell\ell^2\sigma^2\int_0^1u\,du
=\frac{e^\ell\ell^2\sigma^2}{2}.
\end{align*}
A constant additive prediction error cancels in $\operatorname{Var}_p(\widehat\pi(X)-\pi(X))$. (If one additionally assumes a uniform error bound $\sup_x|\widehat\pi(x)-\pi(x)|\le\epsilon$, then $\operatorname{Var}_{q_u^s}(\widehat\pi(X)-\pi(X))\le\epsilon^2$ directly drops the $e^\ell$ factor.)\\[-0.5em]

\noindent\textbf{Step 3: Bound the change in the label mean.}
Along the same path $q_u^s$, differentiating the finite normalized sum gives
\[
\frac{d}{du}\mathbb E_{q_u^{s}}[\pi(X)]
=\operatorname{Cov}_{q_u^{s}}(\pi(X),d(X)).
\]
A variable supported on an interval of length $a$ has variance at most $a^2/4$: its squared distance from the interval midpoint is at most that value, and variance minimizes expected squared distance over all centers. Thus $\operatorname{Var}_{q_u^{s}}(\pi(X))\le1/4$, while Step 1 gives $\operatorname{Var}_{q_u^s}(d(X))=\ell^2\operatorname{Var}_{q_u^s}(\widehat\pi(X)-\pi(X))\le e^\ell\ell^2\sigma^2$. Cauchy--Schwarz bounds the absolute covariance by $e^{\ell/2}\ell\sigma/2$. Integrating from $u=0$ to $u=1$ proves the mean bound.\\[-0.5em]

\noindent\textbf{Step 4: Separate prediction error from shortlist coverage.}
For a nonempty shortlist $C\subseteq\mathcal X$, we have $p(C)=\sum_{x\in C}p(x)>0$, and both $q^s(\cdot\mid C)$ and $\widehat q_C^s$ coincide with the true and estimated directional proposals formed under the conditional target distribution $p(\cdot\mid C)=p(\cdot)\mathbf 1_C/p(C)$ on $C$. Applying Steps 1--3 with $(C,p(\cdot\mid C))$ in place of $(\mathcal X,p)$ yields $D_{\mathrm{KL}}(\widehat q_C^s\|q^s(\cdot\mid C))\le e^\ell\ell^2\sigma_C^2/2$ and $|\mathbb E_{\widehat q_C^s}[\pi(X)]-\mathbb E_{q^s(\cdot\mid C)}[\pi(X)]|\le e^{\ell/2}\ell\sigma_C/2$. Since the estimated shortlist proposal is supported on $C$,
\[
D_{\mathrm{KL}}(\widehat q_C^s\|q^s)
=D_{\mathrm{KL}}(\widehat q_C^s\|q^s(\cdot\mid C))
+\log\frac1{q^s(C)}.
\]
The first term is the prediction penalty on the retained prompts; the second is the exact cost of restricting the oracle's support. This proves Eq.~\ref{eq:shortlist-kl-proof}. All calculations condition on choices already made before the draw, so they remain valid for predictable adaptive choices.
\end{proof}

\subsection{Sample cost with predicted preferences and a learned bet}
\label{app:llm-proof}

For an explicit sufficient condition, suppose $\mu\in(1/2,1)$, predictable scores satisfy $\widehat\pi_{t-1}(x)\in[0,1]$, and the high-variance event $A_t:=\{\sigma_{t-1}>\sigma\}\in\mathcal F_{t-1}$ satisfies $\mathbb P(\sigma_{t-1}>\sigma)\le C_A\rho_A^t$ for all $t\ge1$ and some constants $\sigma\ge0$, $C_A\ge0$, and $\rho_A\in(0,1)$ (which includes an every-round bound via $C_A=0$, arbitrary predictions during any finite warmup $t\le t_0$, and exponentially concentrating online predictors). At the matched bet ($\nu=\mu$, with matched directional oracles $q^+,q^-$), write
\[
\ell=\log\frac\mu{1-\mu},\qquad \eta=\mathbb E_{q^-}[\pi(X)].
\]
Assume
\begin{equation}
\eta-\frac{e^{\ell/2}\ell\sigma}{2}>\frac12,
\qquad
G_{\mathrm{LLM}}=\kappa D_{\mathrm{KL}}(\mu\|1/2)
+(1-\kappa)\left(G^\star-\frac{e^{\ell}\ell^2\sigma^2}{2}\right)>0.
\label{eq:llm-rate}
\end{equation}
The first condition keeps both estimated components on the correct side of mean label $1/2$, so the evidence weights can learn the direction. The second leaves a positive growth rate after the prediction penalty. These are sufficient conditions; their failure alone does not imply that the procedure is inefficient. Using population variance $\sigma_{t-1}^2$ and allowing $C_A>0$ avoids requiring a uniform per-prompt error bound or accuracy at every round.

\begin{corollary}[LLM-guided sampling]
\label{cor:llm-sample}
Under these conditions, use the estimated mixture in Eq.~\ref{eq:practical-proposal}.
For a fixed matched bet and $\kappa\in[0,1]$, there is a finite $C$, independent of $\alpha$, such that
\begin{equation}
\mathbb E[\tau_\alpha]\le\frac{B}{G_{\mathrm{LLM}}}+C.
\label{eq:llm-expected-cost}
\end{equation}
An upper-tail bound of the form in Eq.~\ref{eq:oracle-cost-tail} also holds with leading rate $G_{\mathrm{LLM}}$.

For the plug-in bet in Eq.~\ref{eq:adaptive-bet}, suppose $\kappa>0$ and $\bar\nu\ge\mu$. For every fixed $0<G<G_{\mathrm{LLM}}$, the same expected-cost and upper-tail bounds hold with leading rate $G$ and finite constants that may depend on $G$. In particular,
\begin{equation}
\limsup_{\alpha\downarrow0}\frac{\mathbb E[\tau_\alpha]}{B}
\le\frac1{G_{\mathrm{LLM}}}.
\label{eq:llm-adaptive-asymptotic}
\end{equation}
\end{corollary}

The proof below first controls the wrong-direction weight and then the cost of learning the bet. The rate is a lower bound, not an exact rate or a guarantee of any particular LLM's prediction accuracy. These assumptions affect efficiency only; Lemma~\ref{lem:valid_eprocess} continues to hold without them.

\begin{proof}[Proof of Corollary~\ref{cor:llm-sample}]
The proof verifies the same cumulative-growth condition as in the oracle analysis. For a fixed bet, prediction error reduces the useful growth rate, while learning the direction and any high-variance rounds contribute a finite total deficit. For a learned bet, we additionally show that rounds with a poor bet have geometrically decreasing probabilities. The predictor itself may change arbitrarily with the history, as long as the assumed variance tail condition holds.\\[-0.5em]

\noindent\textbf{Step 1: Predicted proposals still reveal the correct direction on good-prediction rounds.}
First fix $\nu_t=\mu$. The matched true proposals satisfy
$\mathbb E_{q^+}[\pi(X)]\ge\mu\ge\mathbb E_{q^-}[\pi(X)]=\eta$.
On $A_t^c=\{\sigma_{t-1}\le\sigma\}$, by the label-mean bound in Lemma~\ref{lem:prediction-loss}, each estimated component has label mean at least $\eta-e^{\ell/2}\ell\sigma/2$. Target sampling has mean $\mu\ge\eta$. Thus, on $A_t^c$,
\begin{equation}
\mathbb E[Y_t\mid\mathcal F_{t-1}]
\ge\eta-\frac{e^{\ell/2}\ell\sigma}{2}>1/2
\label{eq:learned-label-drift}
\end{equation}
under the final proposal, including exploration, so $\mathbb E[H_t-H_{t-1}\mid\mathcal F_{t-1}]=\ell(2\mathbb E[Y_t\mid\mathcal F_{t-1}]-1)\ge c_1:=\ell(2\eta-e^{\ell/2}\ell\sigma-1)>0$.

Since $|H_u-H_{u-1}|=\ell$ for every $u\ge1$, reserve an initial segment $m=\lfloor c_1t/(4(c_1+\ell))\rfloor<t$ for any integer $t\ge1$. Then $H_m\ge-\ell m$, and if $A_u^c$ holds for all $u\in\{m+1,\ldots,t\}$, then $H_m+\sum_{u=m+1}^t\mathbb E[H_u-H_{u-1}\mid\mathcal F_{u-1}]\ge c_1t-(c_1+\ell)m\ge3c_1t/4$. Since centered increments have absolute value at most $2\ell$ and $1\le t-m\le t$, a union bound and Azuma--Hoeffding yield
\[
\mathbb P\{H_t<c_1t/2\}
\le\sum_{u=m+1}^t\mathbb P(A_u)+\exp\{-c_1^2t/(128\ell^2)\}
\le\frac{C_A\rho_A^{m+1}}{1-\rho_A}+\exp\{-c_1^2t/(128\ell^2)\}.
\]
Because $m+1>c_1t/(4(c_1+\ell))$ and $1-w_t=(1+e^{H_t})^{-1}\le e^{-c_1t/2}$ on $\{H_t\ge c_1t/2\}$, there exist finite $C_{0,1}\ge1$ and $\rho_1\in(0,1)$ such that
\begin{equation}
\mathbb E[1-w_t]\le C_{0,1}\rho_1^t,\qquad t\ge0.
\label{eq:learned-weight-decay}
\end{equation}
This step explains the first condition in Eq.~\ref{eq:llm-rate}: it prevents prediction error from reversing the direction learned by the evidence ratio.

\noindent\textbf{Step 2: Identify the fixed-bet growth rate.}
Since $\widehat\pi_{t-1}(x)\in[0,1]$, full-population estimated tilts satisfy $\widehat q_t^s\ge e^{-\ell}p$, so $q_t\ge\{\kappa+(1-\kappa)e^{-\ell}\}p\ge e^{-\ell}p$ on every round (even when $\kappa=0$, reducing Eq.~\ref{eq:practical-proposal} to Eq.~\ref{eq:estimated-mixture}), giving floor parameter $e^{-\ell}>0$ and a finite log-increment bound $|S_t-S_{t-1}|\le b$ in Theorem~\ref{thm:adaptive-sample}. For any distribution $q$ on the population,
\[
D_{\mathrm{KL}}(q\|q^+)\le\log\frac1{\min_xq^+(x)}.
\]
Indeed, $\sum_xq(x)\log q(x)\le0$, and the remaining cross-entropy is bounded by the displayed maximum. Let $C_1:=\max\{\log(1/\min_xq^+(x)),\,G_{\mathrm{LLM}}+b\}$. On $A_t^c$, using the oracle identity, convexity, and the prediction-error bound,
\begin{align*}
G_t^+(q_t)
\ge{}&G^\star-\kappa D_{\mathrm{KL}}(p\|q^+)\\
&-(1-\kappa)\left\{w_{t-1}\frac{e^\ell\ell^2\sigma^2}{2}
+(1-w_{t-1})C_{1}\right\}\\
\ge{}&G_{\mathrm{LLM}}-C_{1}(1-w_{t-1}),
\end{align*}
while on $A_t$ we have $G_t^+(q_t)\ge-b\ge G_{\mathrm{LLM}}-C_1$. Here $G^\star-D_{\mathrm{KL}}(p\|q^+)=D_{\mathrm{KL}}(\mu\|1/2)$, which accounts for the target-sampling part of $G_{\mathrm{LLM}}$.

Thus the cumulative-growth condition holds with rate $G_{\mathrm{LLM}}$ and deficit
\[
R=C_{1}\sum_{t\ge1}\{(1-w_{t-1})+\mathbf 1_{A_t}\}.
\]
Each summand $U_{t-1}:=C_1\{(1-w_{t-1})+\mathbf 1_{A_t}\}$ lies in $[0,2C_1]$ and satisfies $\mathbb E[U_{t-1}]\le C_1(C_{0,1}\rho_1^{t-1}+C_A\rho_A^t)\le(2C_1)\widetilde C_0\widetilde\rho^{t-1}$ with $\widetilde C_0:=\max\{1,(C_{0,1}+C_A\rho_A)/2\}$ and $\widetilde\rho:=\max\{\rho_1,\rho_A\}\in(0,1)$. By Eq.~\ref{eq:learned-weight-decay} and Lemma~\ref{lem:geometric-sum}, this deficit has finite expectation and an exponential tail, so Theorem~\ref{thm:adaptive-sample} applies. Its expectation bound gives Eq.~\ref{eq:llm-expected-cost}; its upper-quantile formula and the exponential deficit tail give the stated concentration bound.\\[-0.5em]

\noindent\textbf{Step 3: A learned bet is eventually close to the matched bet.}
Now use Eq.~\ref{eq:adaptive-bet}, with $\kappa>0$ and $\bar\nu\ge\mu$. Fix a desired rate $0<G<G_{\mathrm{LLM}}$. We will work in a small neighborhood of the matched bet, where both useful growth and directional drift remain positive.

Use a subscript $\nu$ for a true proposal or growth rate evaluated at a fixed bet $\nu$, and write $\ell_\nu=\log(\nu/(1-\nu))$. The two quantities
\begin{align*}
&\kappa G_\nu^+(p)
+(1-\kappa)\{G_\nu^\star-e^{\ell_\nu}\ell_\nu^2\sigma^2/2\},\\
&\min\{\mu,\mathbb E_{q_\nu^-}[\pi(X)]-e^{\ell_\nu/2}\ell_\nu\sigma/2\}
\end{align*}
are continuous in $\nu\in[1/2,\bar\nu]$. At $\nu=\mu$, the first equals $G_{\mathrm{LLM}}>G$ and the second exceeds $1/2$. Choose $\epsilon_0>0$ small enough that these inequalities persist whenever $|\nu-\mu|\le\epsilon_0$, and take $\epsilon_0<(\mu-1/2)/2$. In this neighborhood, $\ell_\nu$ is also bounded away from zero.

All true positive proposals with $\nu\le\bar\nu$ have a common positive floor $q_\nu^+\ge e^{-\bar\ell}p$, where $\bar\ell=\log(\bar\nu/(1-\bar\nu))$. Hence their KL losses against arbitrary distributions admit a uniform finite bound $C$, as in Step 2. Continuity and the mean-perturbation bound now give a constant $c>0$ such that, on every round with $|\nu_t-\mu|\le\epsilon_0$ and $A_t^c=\{\sigma_{t-1}\le\sigma\}$,
\begin{equation}
\begin{split}
G_t^+(q_t)&\ge G-C(1-w_{t-1}),\\
\mathbb E[H_t-H_{t-1}\mid\mathcal F_{t-1}]&\ge c.
\end{split}
\label{eq:good-bet-neighborhood}
\end{equation}
For the second inequality we used
$H_t-H_{t-1}=\ell_t(2Y_t-1)$: both the log odds and the conditional label advantage are uniformly positive in this neighborhood.

It remains to bound the probability of being outside the neighborhood. Each importance-weighted observation $Z_u:=p(X_u)Y_u/q_u(X_u)$ has conditional mean $\mu$ and lies in $[0,1/\kappa]$. Applying Hoeffding's exponential-moment bound ($\mathbb E[e^{\theta(Z_u-\mu)}\mid\mathcal F_{u-1}]\le e^{\theta^2/(8\kappa^2)}$ on $[0,1/\kappa]$) conditionally and iterating over the observations yields the usual bounded-martingale concentration inequality at $\theta=\pm4\kappa^2\epsilon_0$ for $t\ge2$~\citep{hoeffding1963probability,azuma1967weighted}. Clipping the average to $[0,1]$, applying $z\mapsto\max\{z,1-z\}$, and capping at $\bar\nu\ge\mu$ are all $1$-Lipschitz operations; their composition maps $\mu$ to itself. Therefore (and trivially at $t=1$, where the right-hand side equals $2\ge1$)
\begin{equation}
\mathbb P\{|\nu_t-\mu|>\epsilon_0\}
\le2\exp\{-2\kappa^2\epsilon_0^2(t-1)\},\qquad t\ge 1.
\label{eq:bet-learning-tail}
\end{equation}
Thus poor-bet rounds become geometrically unlikely. This is where fixed positive exploration is used to learn the bet: it uniformly bounds the importance-weighted observations.\\[-0.5em]

\noindent\textbf{Step 4: Changing bets still lead to a geometric weight bound.}
With $\bar\ell=\log(\bar\nu/(1-\bar\nu))$, every ratio increment has absolute value at most $\bar\ell$. For any integer $t\ge1$, set
\[
m=\left\lfloor\frac{ct}{4(c+\bar\ell)}\right\rfloor<t.
\]
This reserves a short initial segment for potentially poor behavior. Regardless of what happens in that segment, $H_m\ge-\bar\ell m$. If all bets in rounds $m+1,\ldots,t$ are in the neighborhood and $A_u^c$ holds for all $u\in\{m+1,\ldots,t\}$, Eq.~\ref{eq:good-bet-neighborhood} gives
\begin{align*}
H_t\ge{}&ct-(c+\bar\ell)m\\
&+\sum_{u=m+1}^t\{H_u-H_{u-1}
-\mathbb E[H_u-H_{u-1}\mid\mathcal F_{u-1}]\}.
\end{align*}
The first line is at least $3ct/4$. The centered increments in the second line have absolute value at most $2\bar\ell$. Consequently, the event $H_t<ct/2$ requires either a poor bet or high-variance prediction in the final segment or a martingale fluctuation below $-ct/4$. A union bound and Azuma--Hoeffding yield
\[
\mathbb P\{H_t<ct/2\}
\le\sum_{u=m+1}^t\bigl(\mathbb P\{|\nu_u-\mu|>\epsilon_0\}+\mathbb P(A_u)\bigr)
+\exp\{-c^2t/(128\bar\ell^2)\}.
\]
The first term decreases geometrically by Eq.~\ref{eq:bet-learning-tail} and $\mathbb P(A_u)\le C_A\rho_A^u$, since $m+1>ct/(4(c+\bar\ell))$ for every $t\ge1$. So does the second. This argument uses event inclusion, rather than conditioning the martingale on a favorable future event.

On $\{H_t\ge ct/2\}$, we also have $1-w_t\le e^{-ct/2}$. Combining the two cases and absorbing finitely many early rounds gives finite $C_0\ge1$ and $0<\rho<1$ such that
\begin{equation}
\mathbb E[1-w_t]\le C_0\rho^t,\qquad t\ge0.
\label{eq:adaptive-weight-decay}
\end{equation}

\noindent\textbf{Step 5: include all learning costs in one deficit.}
All log increments are bounded by some $b$ (from $q_t\ge\kappa p$ and $\nu_t\le\bar\nu<1$), so $G_t^+(q_t)\ge-b$ even outside the neighborhood or on $A_t$. Increase the constant $C$ in Eq.~\ref{eq:good-bet-neighborhood}, if necessary, so that $C\ge G+b$. Then every round satisfies
\[
G_t^+(q_t)\ge G-C\{(1-w_{t-1})+
\mathbf 1_{\{|\nu_t-\mu|>\epsilon_0\}}+\mathbf 1_{A_t}\}.
\]
Thus a valid total deficit is
\[
R=C\sum_{t\ge1}\{(1-w_{t-1})+
\mathbf 1_{\{|\nu_t-\mu|>\epsilon_0\}}+\mathbf 1_{A_t}\}.
\]
Its nonnegative summands are bounded by $3C$, and their expectations decrease geometrically by Eqs.~\ref{eq:bet-learning-tail}, \ref{eq:adaptive-weight-decay}, and $\mathbb P(A_t)\le C_A\rho_A^t$ (with parameters $3C$, $\widehat C_0:=\max\{1,(C_0+2+C_A\rho_A)/3\}$, and $\widehat\rho:=\max\{\rho,e^{-2\kappa^2\epsilon_0^2},\rho_A\}\in(0,1)$). Lemma~\ref{lem:geometric-sum} gives finite expectation and an exponential tail. Theorem~\ref{thm:adaptive-sample} now proves both sample-cost bounds at the chosen rate $G$.

Finally, for every fixed $G<G_{\mathrm{LLM}}$, the additive constants disappear after dividing by $B$ and taking $\limsup_{\alpha\downarrow0}$. Letting $G\uparrow G_{\mathrm{LLM}}$ afterward proves Eq.~\ref{eq:llm-adaptive-asymptotic}. This order of limits is important: the finite-threshold constants may grow as $G$ approaches $G_{\mathrm{LLM}}$.
\end{proof}

\paragraph{Randomized uniform exploration ($\epsilon$-greedy sampling).}
Sampling $X_t\sim p$ with probability $\kappa\in(0,1]$ and $X_t\sim\widehat q_t$ with probability $1-\kappa$ via an auxiliary coin $Z_t\sim\operatorname{Bernoulli}(\kappa)$ independent of $\mathcal F_{t-1}$ has exact conditional law $q_t$ in Eq.~\ref{eq:practical-proposal}, so weighting by the marginal proposal $q_t(X_t)$ in Eqs.~\ref{eqn:eprocess} and~\ref{eq:mean-estimate} satisfies Lemma~\ref{lem:valid_eprocess} and Corollary~\ref{cor:llm-sample} verbatim. Even if an $\epsilon$-greedy implementation instead weights by the realized branch density $q_{t,Z_t}:=Z_tp+(1-Z_t)\widehat q_t$ (updating $\mathcal F_t=\sigma(\mathcal F_{t-1},Z_t,X_t,Y_t)$, with $Y_t$ conditionally independent of $Z_t$ given $(\mathcal F_{t-1},X_t)$ as in Eq.~\ref{eq:sequential-model}), full-population tilts satisfy $\widehat q_t^s\ge e^{-\bar\ell}p$ whenever $\nu_t\in[1/2,\bar\nu]$ (with $\bar\ell=\log(\bar\nu/(1-\bar\nu))$, taking $\bar\nu=\mu$ in the fixed-bet case), giving $q_{t,Z_t}\ge e^{-\bar\ell}p$ on both branches. Conditional on $(\mathcal F_{t-1},Z_t)$, each update factor still has null expectation at most one (preserving Lemma~\ref{lem:valid_eprocess}) and $p(X_u)Y_u/q_{u,Z_u}(X_u)\in[0,e^{\bar\ell}]$ has conditional mean $\mu$ (preserving Eq.~\ref{eq:bet-learning-tail} with $e^{-\bar\ell}$ in place of $\kappa$), while applying the bounded-increment martingale argument of Theorem~\ref{thm:adaptive-sample} to $\mathbb E[\log(E_t^+/E_{t-1}^+)\mid\mathcal F_{t-1}]=\kappa G_{\nu_t}^+(p)+(1-\kappa)G_{\nu_t}^+(\widehat q_t)$ yields the same drift and cumulative-growth bounds as Steps~1--5 above and hence the identical leading rate $G_{\mathrm{LLM}}$ (likewise extending the plug-in-bet bound to $\kappa=0$, where $q_t=q_{t,0}=\widehat q_t\ge e^{-\bar\ell}p$).

\subsection{Other fixed bets}
\label{app:general-mixture}

The matched-oracle proof uses two properties of the bet: positive growth of the correct oracle and a label mean above $1/2$ under the opposite proposal. These properties, rather than exact matching itself, are enough for the argument.

Fix $\nu\in(1/2,1)$, let $q_\nu^+$ and $q_\nu^-$ be its true directional proposals, and let $G_\nu^\star$ be the positive-direction oracle growth rate. Use their evidence-weighted mixture with this fixed bet.

\begin{proposition}[A fixed-bet mixture]
\label{prop:general-mixture}
Suppose $\mu>1/2$, $\nu\in(1/2,1)$, $\mathbb E_{q_\nu^-}[\pi(X)]>1/2$, and $G_\nu^\star>0$. Then a finite constant $C=C(p,\pi,\nu)$ satisfies, for every $\alpha\in(0,1)$,
\[
\mathbb E[\tau_\alpha]\le\frac{B}{G_\nu^\star}+C,
\qquad
\mathbb E[\tau_\alpha]\sim\frac{B}{G_\nu^\star}
\quad\text{as }\alpha\downarrow0.
\]
Both growth conditions hold whenever $1/2<\nu<\mu$, including for deterministic prompt preferences.
\end{proposition}

\begin{proof}
\noindent\textbf{Step 1: Apply the mixture argument at the chosen bet.}
Write $\ell=\log(\nu/(1-\nu))$ and $\eta=\mathbb E_{q_\nu^-}[\pi(X)]$. Monotone tilting gives $\mathbb E_{q_\nu^+}[\pi(X)]\ge\eta$, so every mixture has label mean at least $\eta>1/2$. Choose $\theta\in(0,1]$ small enough that
\[
\rho=\eta e^{-\theta\ell}+(1-\eta)e^{\theta\ell}<1.
\]
The derivative at zero is negative ($-(2\eta-1)\ell<0$), so such a choice exists, including when $\eta=1$. As before, $\mathbb E[1-w_t]\le\rho^t$, and the total expected KL loss is at most $D_{\mathrm{KL}}(q_\nu^-\|q_\nu^+)/(1-\rho)$.

The component floor is $e^{-\ell}p$, and the correct log increment is at most $\log(2\nu)+\ell$ (and at least $\log(2(1-\nu)\min_xp(x))>-\infty$). The stopped-expectation argument therefore gives
\begin{equation}
\mathbb E[\tau_\alpha]
\le\frac{B+\log(2\nu)+\ell+
D_{\mathrm{KL}}(q_\nu^-\|q_\nu^+)/(1-\rho)}{G_\nu^\star}.
\label{eq:fixed-mixture-explicit-bound}
\end{equation}
Finite total loss, bounded martingale fluctuations, and the wrong-direction supermartingale give the almost-sure hitting-time limit exactly as in Step 4 of Appendix~\ref{app:mixture-proof}. Fatou's lemma and this upper bound then give convergence of expected costs.

\noindent\textbf{Step 2: Verify the conditions for a conservative bet.}
Suppose $1/2<\nu<\mu$. The pointwise inequality in Eq.~\ref{eq:opposite-mean-chord} applies with the present $\ell$, and gives
\[
\mathbb E_p[(\pi(X)-1/2)e^{-\ell\pi(X)}]
\ge-\frac12+\frac{1+e^{-\ell}}2\mu
=\frac{\mu-\nu}{2\nu}>0.
\]
Dividing by the positive normalizing constant proves $\mathbb E_{q_\nu^-}[\pi(X)]>1/2$, without any interior-preference assumption.

Finally, Jensen's inequality yields
\[
G_\nu^\star\ge\mu\log(2\nu)+(1-\mu)\log(2(1-\nu)).
\]
The right side is zero at $\nu=1/2$ and has derivative $(\mu-\nu)/(\nu(1-\nu))>0$ on $(1/2,\mu)$. It is therefore positive, proving the other growth condition.
\end{proof}

A known lower bound $\underline\mu>1/2$ on $\mu$ permits choosing $\nu\in(1/2,\underline\mu)$. A point estimate alone does not certify that the bet is conservative.

\subsection{A deterministic-label boundary under matched betting}
\label{app:mixture-boundary}

The interior-preference assumption in the matched-oracle theorem cannot simply be removed.
\begin{proposition}
\label{prop:mixture-boundary}
Consider two prompts with
\[
p=(3/4,1/4),\qquad\pi=(1,0),\qquad\mu=\nu=3/4.
\]
Then $q^+=(9/10,1/10)$ and $q^-=(1/2,1/2)$, so the opposite proposal produces the correct label only half the time, and $\mathbb E[\tau_\alpha]=\infty$ for every $\alpha \in(0,4/5)$.
\end{proposition}

\begin{proof}[{Proof of Proposition~\ref{prop:mixture-boundary}}]
\noindent\textbf{Step 1: A wrong initial label forces an excursion.}
Let $J_t=\log_3(E_t^+/E_t^-)$. It starts at zero and moves by $+1$ or $-1$ according to the label (here $\{Y_t=1\}=\{X_t=x_1\}$ and $\{Y_t=0\}=\{X_t=x_2\}$ a.s.). At state $j$, its upward transition probability is
\[
\frac12+\frac25\frac{3^j}{1+3^j}.
\]
The first label is zero with probability $3/10$. On this event,
\[
J_1=-1,\qquad E_1^+=5/12,\qquad E_1^-=5/4.
\]
While $J_{t-1}\le-1$, the weight on $q^+$ is at most $1/4$, so the proposal probability of label one lies in $[1/2,3/5]$. A label-one update multiplies $E_{t-1}^-$ by at most $(3/8)/(1/2)=3/4$. A label-zero update multiplies it by at most $(3/8)/(2/5)=15/16$. Thus $E_t^-$ never rises above $5/4$ during this excursion (including at the return step to $J_t=0$). The positive wealth is smaller while $J_t<0$, and the two are equal upon returning to zero. For a threshold exceeding $5/4$, neither process can stop before (or upon) that return.

\noindent\textbf{Step 2: Return time has infinite mean.}
Conditional on the first label being zero, let $Z_s=-J_{1+s}$ until it hits zero, so $Z_0=1$. At state $k\ge1$, the down and up probabilities are
\[
d_k=\frac12+\frac25\frac{3^{-k}}{1+3^{-k}},
\qquad b_k=1-d_k.
\]
Write $T_n=\inf\{s\ge0:Z_s=n\}$, and use $\mathbb P_k$ ($\mathbb E_k$) for the law (expectation) starting at $Z_0=k\ge0$. For fixed $n\ge 1$, let $h_k=\mathbb P_k(T_n<T_0)$. The nearest-neighbor recurrence is
\[
h_k=d_kh_{k-1}+b_kh_{k+1},\qquad h_0=0,\quad h_n=1.
\]
Rearranging gives $h_{k+1}-h_k=(d_k/b_k)(h_k-h_{k-1})$. Iterating these increments and summing from zero to $n$ gives
\[
\mathbb P_1(T_n<T_0)
=\left(\sum_{j=1}^n\prod_{k=1}^{j-1}\frac{d_k}{b_k}\right)^{-1}.
\]
Here $d_k/b_k-1=\frac{8\cdot 3^{-k}}{5+3^{-k}}=O(3^{-k})$, so the products are bounded above by a finite constant $C$ ($\le e^{4/5}$), uniformly in $j$. For example, their logarithms are bounded by the convergent sum $\sum_k(d_k/b_k-1)$. Consequently,
\[
\mathbb P_1(T_n<T_0)\ge\frac1{Cn}.
\]
Reaching level $n$ from one takes at least $n-1$ steps. Hence $\mathbb P_1(T_0\ge n-1)\ge1/(Cn)$, and summing these tail probabilities shows $\mathbb E_1[T_0]=\infty$. The initial wrong-label event has positive probability, so Step 1 implies $\mathbb E[\tau_\alpha]=\infty$ whenever $1/\alpha>5/4$.
\end{proof}

This is a failure of finite expected stopping cost, not of directional error validity. Proposition~\ref{prop:general-mixture} restores finite expected cost on the same instance for every fixed $\nu\in(1/2,3/4)$.

\section{Analysis of Ranking}
\label{app:ranking-analysis}

The pairwise results control one call to the evaluator. We now explain how those guarantees pass through an adaptive sorter. The finite-sample argument uses conditional bounds and does not require independent calls. The central limit theorem needs a stronger assumption: independent evaluation streams along a fixed reference execution.

\subsection{Conditional error spending and ranking cost}
\label{app:ranking-proof}

\noindent\textbf{The general, call-indexed statement.}
This version allows fresh repeated evaluations and history-dependent budgets. It does not require the caching convention of Algorithm~\ref{alg:ranking}. Each call begins from unit evidence, and the ranker observes only its returned winner when selecting the next pair.
Let $\mathcal H_{k-1}$ be all information available before call $k$. Within that call, $\mathcal F_{k,t}$ contains this information and the first $t$ new judgments, so $\mathcal H_{k-1}\subseteq\mathcal F_{k,0}\subseteq\mathcal F_{k,t}$. The pairwise notation $\mathcal F_t$ suppresses $k$. Assume that all requested pairs are non-tied and that the sorter makes at most $M$ calls on every possible execution. Repeated evaluations count as separate calls. Let $\widehat\sigma_{\mathcal A}$ be its returned ordering, and $\sigma_{\mathcal A}^\star$ the output of the same sorter given exact pairwise winners, with the same initial order and random seed.

Suppose the pairwise results give finite bounds $V_k$ and $U_k$, chosen from $\mathcal H_{k-1}$, such that
\begin{equation}
\mathbb E[\tau_k\mid\mathcal H_{k-1}]\le V_k,
\qquad
\mathbb P(\tau_k>U_k\mid\mathcal H_{k-1})\le\beta_k.
\label{eq:conditional-call-cost}
\end{equation}
Here $\tau_k$ is the cost of call $k$; $V_k$ bounds its conditional expectation and $U_k$ its upper quantile. No independence across calls is needed, but the pairwise assumptions must hold conditional on the history at each call. This is why a pilot may choose tolerances, but its labels cannot simply be reused as fresh observations in a new unit-evidence call.

\begin{theorem}[Ranking with a comparison-based algorithm]
\label{thm:ranking-guarantees}
Under the call-indexed setup above (with conditional Lemma~\ref{lem:valid_eprocess} validity and Eq.~\ref{eq:conditional-call-cost} holding for $\mathcal H_{k-1}$-measurable $\alpha_k\in(0,1)$, $\beta_k\ge0$, $V_k<\infty$, and $U_k$, and writing $(x)_+:=\max\{x,0\}$), let $K\le M$ be the number of executed calls, $N_0<\infty$ the nonnegative pilot cost, and $N=N_0+\sum_{k=1}^K\tau_k$. If the predictable budgets satisfy
\begin{equation}
\sum_{k=1}^K\alpha_k\le\delta,
\qquad
\sum_{k=1}^K\beta_k\le\beta
\quad\text{on every execution},
\label{eq:ranking-spending}
\end{equation}
then
\begin{align}
\mathbb P(\widehat\sigma_{\mathcal A}\ne\sigma_{\mathcal A}^\star)&\le\delta,
\label{eq:ranking-fidelity}\\
\mathbb E\!\left[(Q(\sigma_{\mathcal A}^\star)-Q(\widehat\sigma_{\mathcal A}))_+\right]&\le\delta\sum_{i<j}\Delta_{ij},
\label{eq:ranking-quality-regret}\\
\mathbb E[N]&\le\mathbb E[N_0]+\mathbb E\!\left[\sum_{k=1}^K V_k\right],
\label{eq:ranking-expected-cost}\\
\mathbb P\!\left(N>N_0+\sum_{k=1}^KU_k\right)&\le\beta.
\label{eq:ranking-cost-tail}
\end{align}
Thus, with probability at least $1-\delta-\beta$, the output is $\sigma_{\mathcal A}^\star$ and the sample cost is within the stated bound.
\end{theorem}

\begin{proof}[Proof of Theorem~\ref{thm:ranking-guarantees}]
We first control the probability that any comparison is incorrect. On the complementary event, the actual sorter follows exactly the same path as the sorter given true pairwise winners. We then add the conditional sample-cost bounds along the executed path. The key point is that whether call $k$ is executed is already determined when that call begins.\\[-0.5em]

\noindent\textbf{Step 1: Control comparison errors along the adaptive path.}
Recall that $K\le M$ is the number of executed calls and $\mathcal H_{k-1}$ contains the information available before call $k$. Set the cost and error indicators of unexecuted calls to zero. The event $\{k\le K\}$ is $\mathcal H_{k-1}$-measurable. Conditional pairwise validity (for the inferior-model e-process $E_{k,t}^{\mathrm{inf}}\in\{E_{k,t}^+,E_{k,t}^-\}$, optional stopping at bounded stopping times $T_{k,m}:=\inf\{t\ge0:E_{k,t}^{\mathrm{inf}}\ge1/\alpha_k\}\wedge m$ conditional on $\mathcal H_{k-1}\subseteq\mathcal F_{k,0}$, followed by $m\to\infty$) therefore gives
\[
\mathbb P\{k\le K,\ \text{call }k\text{ is incorrect}\}
\le \mathbb E\!\left[\mathbf 1_{\{k\le K\}}\alpha_k\right].
\]
Summing over the deterministic range $k=1,\ldots,M$ and using the pathwise spending constraint yields
\[
\mathbb P\{\text{some executed comparison is incorrect}\}
\le\mathbb E\!\left[\sum_{k=1}^K\alpha_k\right]\le\delta.
\]
Each executed call terminates almost surely because its conditional expected cost is finite. Thus the sorter also terminates after at most $M$ calls.

Couple the actual and noiseless sorters using the same initial input and internal random seed. Their states and requested pairs agree before the first incorrect return: both algorithms have then received exactly the same sequence of winners. If no return is incorrect, their final outputs agree as well. This proves Eq.~\ref{eq:ranking-fidelity}. Transitivity is not used; with cycles, the reference output is simply the output of this particular sorter on the true pairwise winners.\\[-0.5em]

\noindent\textbf{Step 2: Add the conditional sample-cost bounds.}
The same pre-call measurability justifies applying the tower property to each executed cost:
\begin{align*}
\mathbb E[N-N_0]
&=\sum_{k=1}^M\mathbb E\!\left[
\mathbf 1_{\{k\le K\}}\mathbb E[\tau_k\mid\mathcal H_{k-1}]
\right]\\
&\le\mathbb E\!\left[\sum_{k=1}^KV_k\right].
\end{align*}
For the upper-tail bound, the conditional inequality
$\mathbb P(\tau_k>U_k\mid\mathcal H_{k-1})\le\beta_k$ gives
\[
\mathbb P\{\text{some executed call exceeds its }U_k\}
\le\mathbb E\!\left[\sum_{k=1}^K\beta_k\right]\le\beta.
\]
When no call exceeds its bound, $N\le N_0+\sum_{k=1}^KU_k$. This proves Eq.~\ref{eq:ranking-cost-tail}. A final union bound gives simultaneous agreement with the noiseless output and the sample budget with probability at least $1-\delta-\beta$. These arguments use conditional guarantees, not independence or a comparison path fixed in advance.\\[-0.5em]

\noindent\textbf{Step 3: Translate output agreement into a quality bound.}
For any pair, changing its relative order changes its contribution to $Q$ by at most $\Delta_{ij}=|2\mu_{ij}-1|$. Consequently,
\[
|Q(\sigma)-Q(\sigma')|\le\sum_{i<j}\Delta_{ij}
\]
for any two permutations. The quality difference is zero when the actual and reference outputs agree. Bounding it by this sum on the remaining event proves Eq.~\ref{eq:ranking-quality-regret}:
\[
\mathbb E\!\left[(Q(\sigma_{\mathcal A}^\star)-Q(\widehat\sigma_{\mathcal A}))_+\right]
\le\delta\sum_{i<j}\Delta_{ij}.
\]
This controls the loss caused by statistical comparison errors. It does not compare the noiseless sorter with an optimizer of $Q$.
\end{proof}

\noindent\textbf{Example: A budget fixed before merge sort begins.}
A merge of two lists containing $r$ items uses at most $r-1$ comparisons, whatever winners the evaluator returns. There are at most $\lceil\log_2n\rceil$ merge levels, and list lengths sum to $n$ at each level. Hence $M=n\lceil\log_2n\rceil$ is a valid deterministic bound for $n\ge2$.

For $\delta,\beta\in(0,1)$, choose $\alpha_k=\delta/M$ and $\beta_k=\beta/M$. Under the oracle assumptions, Corollary~\ref{cor:oracle-tail} supplies the bound
\begin{align*}
U_e=\Bigg\lceil\,
\frac{\log(M/\delta)}{G_e^\star}
+C_{1,e}\sqrt{\log(M/\delta)\log(2M/\beta)}
+C_{2,e}\log(2M/\beta)+C_{3,e}\,\Bigg\rceil
\end{align*}
for a call comparing pair $e$. Maximizing over all pairs $e\in\mathcal E$ that may be requested gives a budget available before sorting:
\[
\mathbb P\!\left\{N>N_0+M\max_{e\in\mathcal E} U_e\right\}\le\beta.
\]
The realized-path bound in the theorem is usually smaller because it sums the bounds for the pairs actually requested. If a pairwise bound depends on earlier calls, a precomputed budget must also be uniform over those possible histories. Other sorters enter through their own deterministic call bounds or through a separate high-probability bound on their number of calls; an expected call count alone does not suffice here.

\begin{proof}[Proof of Theorem~\ref{thm:ranking-cost}]
Algorithm~\ref{alg:ranking} caches unordered pairs (under $\mu_{ji}=1-\mu_{ij}$), so repeated requests for an already-evaluated pair use zero additional judgments and incur no new comparison error. Index the newly evaluated pairs by $e_k=(i_k,j_k)$,
$k=1,\ldots,K$, where $\mathcal E_{\mathrm{eval}}:=\{e_1,\ldots,e_K\}\subseteq\mathcal E$ and $K\le M$. Their tolerances $\alpha_{ij}\in(0,1)$ are
chosen after the finite $N_0$-judgment pilot, before sorting begins, and satisfy $\sum_{k=1}^K\alpha_{e_k}=\sum_{(i,j)\in\mathcal E_{\mathrm{eval}}}\alpha_{ij}\le\sum_{(i,j)\in\mathcal E}\alpha_{ij}\le\delta$ almost surely. Let $\mathcal H_{k-1}$ be all information before call $k$ (including the pilot and earlier calls), so $\{k\le K\}$ and $(e_k,\alpha_{e_k})$ are $\mathcal H_{k-1}$-measurable. On $\{k\le K\}$, conditional error probability is at most $\alpha_{e_k}$ and conditional expected cost is at most $V_k:=\log(1/\alpha_{e_k})/G_{e_k}+C_{e_k}<\infty$ (so $\tau_k<\infty$ a.s.). Summing over $k=1,\ldots,M$ and coupling with the noiseless sorter as in Step 1 of Theorem~\ref{thm:ranking-guarantees} gives $\mathbb P(\widehat\sigma_{\mathcal A}\ne\sigma_{\mathcal A}^\star)\le\mathbb E[\sum_{k=1}^K\alpha_{e_k}]\le\delta$. Likewise, Step 2 of Theorem~\ref{thm:ranking-guarantees} and positivity $\log(1/\alpha_{ij})/G_{ij}+C_{ij}>0$ on the nonempty set $\mathcal E$ yield, for every $0\le K\le M$,
\begin{align*}
\mathbb E[N-N_0]
&\le\mathbb E\!\left[\sum_{k=1}^K\left\{\frac{\log(1/\alpha_{e_k})}{G_{e_k}}+C_{e_k}\right\}\right]
\le M\,\mathbb E\!\left[\max_{(i,j)\in\mathcal E}\left\{\frac{\log(1/\alpha_{ij})}{G_{ij}}+C_{ij}\right\}\right],
\end{align*}
which proves Eq.~\ref{eq:ranking-cost-main} without assuming independence across calls.
\end{proof}

\begin{corollary}[Oracle and LLM-guided instantiations]
\label{cor:ranking-practical}
For tolerances $\alpha_{ij}$ as in Theorem~\ref{thm:ranking-cost}, suppose that every pair in $\mathcal E$ satisfies the assumptions of Theorem~\ref{thm:sample_complexity}, or Corollary~\ref{cor:llm-sample}. Then Eq.~\ref{eq:ranking-cost-main} holds (for finite constants $C_{ij}>0$) with:
\begin{enumerate}[label=(\roman*), leftmargin=*, nosep]
    \item $G_{ij}=G_{ij}^\star$ for the oracle mixture (Theorem~\ref{thm:sample_complexity});
    \item any fixed $G_{ij}\in(0,G_{ij,\mathrm{LLM}})$ for full-pool predictions with the learned bet $\nu_t$ (Corollary~\ref{cor:llm-sample}).
\end{enumerate}
\end{corollary}

\begin{proof}[Proof of Corollary~\ref{cor:ranking-practical}]
After exchanging model labels when $\mu_{ij}<1/2$, the pairwise assumptions on every $(i,j)\in\mathcal E$ supply pair constants $(G_{ij},C_{ij})$ with $G_{ij}>0$ from Theorem~\ref{thm:sample_complexity} or Corollary~\ref{cor:llm-sample}; since $\tau_\alpha\ge1$ for all $\alpha\in(0,1)$, letting $\alpha\uparrow1$ shows $C_{ij}\ge1>0$ for all $(i,j)\in\mathcal E$. Conditional on $\mathcal H_{k-1}$ at the pilot-chosen tolerance $\alpha_{e_k}\in(0,1)$, Lemma~\ref{lem:valid_eprocess} (via optional stopping at $T_{k,m}$ as in Theorem~\ref{thm:ranking-guarantees}) bounds each executed test's conditional error probability by $\alpha_{e_k}$, and the conditional pairwise assumptions yield $\mathbb E[\tau_k\mid\mathcal H_{k-1}]\le\log(1/\alpha_{e_k})/G_{e_k}+C_{e_k}$. Theorem~\ref{thm:ranking-cost} therefore yields Eq.~\ref{eq:ranking-cost-main}.
\end{proof}

\noindent\textbf{Corresponding high-probability sample budget.} For any $\beta\in(0,1)$ (with $M\ge1$), let $U$ be the maximum over $(i,j)\in\mathcal E$ of the upper quantile in Eq.~\ref{eq:cost-tail-main} evaluated at $(\alpha_{ij},\beta/M)$. Conditional on the pilot and prior calls, each executed evaluation exceeds $U$ with probability at most $\beta/M$, so Theorem~\ref{thm:ranking-guarantees} (Eq.~\ref{eq:ranking-cost-tail}) gives $\mathbb P(N>N_0+MU)\le\beta$ and $\mathbb P(\widehat\sigma_{\mathcal A}=\sigma_{\mathcal A}^\star\text{ and }N\le N_0+MU)\ge1-\delta-\beta$ unconditionally (keeping any pilot-dependent $U$ inside the probability event).

\subsection{A central limit theorem for total sample cost}
\label{app:ranking-clt}

\begin{theorem}[Ranking sample-cost fluctuations]
\label{thm:ranking-clt}
Fix a finite model collection and a comparison-based sorter satisfying the interface in Section~\ref{sec:ranking}, with at most $M$ calls on every execution. Fix its initial order and internal random seed. Suppose every pair it may request satisfies the matched-oracle assumptions of Theorem~\ref{thm:sample_complexity}, after exchanging model labels when necessary. Each call starts from unit evidence, uses only its own independent fresh evaluation stream, and has tolerance $\alpha_k=c_k\delta$, where $c_k>0$ and $\sum_{k=1}^M c_k\le1$ are fixed.

Let $e_1,\ldots,e_m$ ($m\le M$) be the sequence of comparisons made with exact pairwise winners, including repetitions. With a fixed pilot cost $N_0$ and total sample cost $N_\delta$ at budget $\delta\in(0,1)$, as $\delta\downarrow0$ (writing $\mathcal N(0,0)$ for point mass at $0$ if $m=0$),
\begin{equation}
\frac{N_\delta-N_0-B_\delta\sum_{k=1}^m1/G_{e_k}^\star}{\sqrt{B_\delta}}
\ \Rightarrow\ 
\mathcal N\!\left(0,\sum_{k=1}^m\frac{v_{e_k}^\star}{(G_{e_k}^\star)^3}\right),
\qquad B_\delta=\log(1/\delta),
\label{eq:ranking-clt}
\end{equation}
where $v_e^\star$ is the correct-direction variance in Eq.~\ref{eq:oracle-variance}.
\end{theorem}

\begin{proof}[Proof of Theorem~\ref{thm:ranking-clt}]
The actual comparison sequence is random because later pairs depend on earlier returned winners. We therefore first prove the limit for the fixed noiseless sequence and then show that the actual execution coincides with that reference with probability tending to one.\\[-0.5em]

\noindent\textbf{Step 1: Construct independent costs on the reference path.}
Force the sorter to receive the true winner at each call, but run an independent oracle evaluation to measure the cost of that call. Let $\tau_k^{\mathrm{ref}}$ be this stopping time for pair $e_k$ at tolerance $c_k\delta$, and define
\[
N_\delta^{\mathrm{ref}}=N_0+\sum_{k=1}^m\tau_k^{\mathrm{ref}}.
\]
A fresh stream is assigned to each call, even if a pair appears more than once. The sequence $e_1,\ldots,e_m$ is fixed, so these reference stopping times are independent. This assertion is made for the reference construction, not for the actual data-dependent path.\\[-0.5em]

\noindent\textbf{Step 2: Sum the pairwise limits.}
For $\delta\in(0,1)$, $\alpha_k=c_k\delta\in(0,1)$ and the threshold for call $k$ is
\[
B_{k,\delta}:={}\log\{1/(c_k\delta)\}=B_\delta+\log(1/c_k)\ge B_\delta>0.
\]
The second term is fixed as $\delta$ decreases. It changes the centering by a constant and changes the ratio of the two square-root scales by a quantity tending to zero. Combining Theorem~\ref{thm:pairwise-clt} and Slutsky's theorem results in:
\[
\frac{\tau_k^{\mathrm{ref}}-B_\delta/G_{e_k}^\star}{\sqrt{B_\delta}}
{}=\sqrt{\frac{B_{k,\delta}}{B_\delta}}\,\frac{\tau_k^{\mathrm{ref}}-B_{k,\delta}/G_{e_k}^\star}{\sqrt{B_{k,\delta}}}
+\frac{\log(1/c_k)}{G_{e_k}^\star\sqrt{B_\delta}}
\ \Rightarrow\ 
\mathcal N\!\left(0,\frac{v_{e_k}^\star}{(G_{e_k}^\star)^3}\right).
\]
Independence gives joint convergence to independent normal variables. Since $m$ is fixed, summing these variables proves Eq.~\ref{eq:ranking-clt} for $N_\delta^{\mathrm{ref}}$ by the continuous-mapping theorem~\citep{billingsley1999convergence}.\\[-0.5em]

\noindent\textbf{Step 3: Transfer the limit to the actual execution.}
Run the actual sorter with the same streams as the reference at corresponding calls until the two paths separate. They can separate only when a reference evaluation returns an incorrect winner. By pairwise validity,
\[
\mathbb P\{\text{some reference return is incorrect}\}
\le\sum_{k=1}^m c_k\delta\le\delta.
\]
Otherwise, the requested pairs, stopping times, and final costs are identical. Thus
\[
\mathbb P\{N_\delta\ne N_\delta^{\mathrm{ref}}\}\le\delta\longrightarrow0.
\]
This is enough to transfer the distributional limit: the probabilities of any event involving the two normalized costs differ by at most $\delta$. No bound on the cost after an incorrect return is needed for this step, and we do not infer convergence of expectations from it.
\end{proof}

For a randomized sorter, the theorem conditions on its seed. Different seeds may give different leading sums $\sum_k1/G_{e_k}^\star$, so an unconditional Gaussian limit with deterministic centering need not follow. One may likewise condition on a pilot that fixes the sorter and allocation, provided the reference path and positive weights are then fixed and the evaluation streams remain independent. Sharing LLM learning across pairs is compatible with Theorem~\ref{thm:ranking-guarantees} when its conditional bounds hold, but does not by itself satisfy the independence assumption of this central limit theorem.

\subsection{Derivation of the comparison-error allocation}
\label{app:ranking-allocation}

For a permutation $\sigma$, larger $\sigma(i)$ means a higher rank for $M_i$. Recall that our agreement objective is
\begin{equation*}
Q(\sigma)=\sum_{i<j}\{\mu_{ij}\mathbf1[\sigma(j)>\sigma(i)]+(1-\mu_{ij})\mathbf1[\sigma(i)>\sigma(j)]\}.
\end{equation*}
Reversing a single pair's relative order changes its contribution by $\Delta_{ij}=|2\mu_{ij}-1|$. A sorting error may also change other relative orders; a budget on direct comparison errors is consequently not a bound on $Q$-regret.

\paragraph{The candidate-pair allocation in Algorithm~\ref{alg:ranking}.}
Given a finite nonempty candidate set $\mathcal E$, positive estimates $\widehat G_{ij}>0$, and nonnegative estimates $\widehat\Delta_{ij}\ge0$ fixed before ranking, solve
\begin{equation}
\begin{aligned}
\min_{0<\alpha_{ij}\le\bar\alpha}\quad&
\sum_{(i,j)\in\mathcal E}\frac{\log(1/\alpha_{ij})}{\widehat G_{ij}}
\text{ subject to }
\sum_{(i,j)\in\mathcal E}\widehat\Delta_{ij}\alpha_{ij}\le\varepsilon,
 \sum_{(i,j)\in\mathcal E}\alpha_{ij}\le\delta.
\end{aligned}
\label{eq:candidate-allocation}
\end{equation}
Take $\varepsilon,\delta>0$ and $\bar\alpha\in(0,1)$. Strict feasibility follows by choosing every tolerance sufficiently small (e.g., $\alpha_{ij}^{(0)}=\frac12\min\{\bar\alpha,\delta/|\mathcal E|,\varepsilon/(\sum_{e\in\mathcal E}\widehat\Delta_e+1)\}\in(0,\bar\alpha)$); the objective is strictly convex and diverges at any zero tolerance (and since $\alpha_{e'}\le\bar\alpha$ lower-bounds each other summand by $\widehat G_{e'}^{-1}\log(1/\bar\alpha)$, every nonempty sublevel set is a closed subset of a compact box in $(0,\bar\alpha]^{|\mathcal E|}$, so a unique minimizer is attained; indeed Eq.~\ref{eq:candidate-allocation} is Proposition~\ref{prop:call-list-allocation} specialized to $m=|\mathcal E|$). The derivative of the Lagrangian for pair $(i,j)$ is
$-1/(\widehat G_{ij}\alpha_{ij})+\xi\widehat\Delta_{ij}+\zeta$.
Minimizing on the capped interval gives Eq.~\ref{eq:allocation-policy} (with $1/0=+\infty$). The nonnegative multipliers satisfy
\[
\xi\Bigl(\sum_{\mathcal E}\widehat\Delta_{ij}\alpha_{ij}-\varepsilon\Bigr)=0,
\qquad
\zeta\Bigl(\sum_{\mathcal E}\alpha_{ij}-\delta\Bigr)=0.
\]
These conditions are necessary and sufficient~\citep{vandenberghe2004convex}. If a denominator vanishes, the capped solution is $\bar\alpha$. Rate estimates must be strictly positive, so a practical implementation floors nonpositive estimates before optimization. A convex solver can solve Eq.~\ref{eq:candidate-allocation} directly; no knowledge of the eventual comparison path is needed.

With only the global budget and inactive caps ($\delta\max_{(i,j)\in\mathcal E}\widehat G_{ij}^{-1}/\sum_{e\in\mathcal E}\widehat G_e^{-1}\le\bar\alpha$), $\alpha_{ij}=\delta\widehat G_{ij}^{-1}/\sum_{\mathcal E}\widehat G_{ij}^{-1}$. Equal rates give uniform allocation. Poor estimates can make the allocation inefficient, but $\sum\alpha_{ij}\le\delta$ and caching still control the probability of any incorrect comparison. The gap-weighted constraint involves estimates and cannot be interpreted as a true-gap budget without additional estimation control. The program models the cost of all candidates; it need not minimize actual cost for an adaptive ranker that requests only some of them.\\[-0.5em]

\noindent\textbf{Call-list formulations.}
During the ranking process, a difficult comparison costs more samples, but reversing a nearly tied pair has a smaller direct effect on $Q$. To formalize this tradeoff, first fix a list of $m$ comparison calls $e_1,\ldots,e_m$. Write $G_{e_k}>0$ for the evidence rate used to model call $k$'s cost, and $\Delta_{ij}=|2\mu_{ij}-1|$ for a pair's preference gap. Each occurrence of a repeated pair receives its own tolerance $\alpha_k$.

\begin{proposition}[Optimal call-list allocation]
\label{prop:call-list-allocation}
Fix $m\ge1$ calls with rates $G_{e_k}>0$, gaps $\Delta_{e_k}\ge0$, budgets $\varepsilon,\delta>0$, and cap $\bar\alpha\in(0,1)$, with $1/0=+\infty$.
The cost model is $\log(1/\alpha_k)/G_{e_k}$, and the direct expected error cost is at most $\Delta_{e_k}\alpha_k$. Consider
\begin{equation}
\min_{0<\alpha_k\le\bar\alpha}
\sum_{k=1}^m\frac{\log(1/\alpha_k)}{G_{e_k}}
\quad\text{subject to}\quad
\sum_{k=1}^m\Delta_{e_k}\alpha_k\le\varepsilon.
\label{eq:ranking_obj}
\end{equation}
This is a direct-error surrogate, not a constraint on final ranking quality: a wrong comparison can change both other relative orders and the subsequent comparison path. Its unique solution is
\begin{equation}
\alpha_k=\min\left\{\bar\alpha,\frac1{\xi\Delta_{e_k}G_{e_k}}\right\},
\label{eq:ranking-allocation}
\end{equation}
where a unique $\xi>0$ satisfies $\sum_{k=1}^m\Delta_{e_k}\alpha_k=\varepsilon$ when $\varepsilon<\bar\alpha\sum_{k=1}^m\Delta_{e_k}$, and $\alpha_k=\bar\alpha$ for all $k$ (e.g., $\xi=0$) when $\varepsilon\ge\bar\alpha\sum_{k=1}^m\Delta_{e_k}$. Adding the global constraint $\sum_k\alpha_k\le\delta$ gives the unique solution
\begin{equation}
\alpha_k=\min\left\{\bar\alpha,
\frac1{G_{e_k}(\xi\Delta_{e_k}+\zeta)}\right\},
\qquad\xi,\zeta\ge0,
\label{eq:ranking-certified-allocation}
\end{equation}
with the multipliers chosen to satisfy complementary slackness.
\end{proposition}
Derivation below establishes both rules using convex optimality conditions~\citep{vandenberghe2004convex}. The global constraint, not the direct-error constraint alone, supplies the error spending required by Theorem~\ref{thm:ranking-guarantees}.

\begin{proof}[Proof of Proposition~\ref{prop:call-list-allocation}]
We index the allocation by calls rather than by distinct pairs. Fix the list $e_1,\ldots,e_m$ with nonnegative gaps $\Delta_{e_k}\ge0$ and positive rates $G_{e_k}>0$. Each call has its own tolerance $\alpha_k$, even when a pair repeats. Linearity of expectation bounds the expected direct comparison-error cost by $\sum_k\Delta_{e_k}\alpha_k$, without requiring independent errors. As above, this is a surrogate cost, not the final ranking-quality loss.\\[-0.5em]

\noindent\textbf{Step 1: A Single direct-error budget.}
The objective in Eq.~\ref{eq:ranking_obj} is separable across coordinates, decreasing in every tolerance, and strictly convex, since its off-diagonal second derivatives vanish and its diagonal second derivatives are
\[
\frac{1}{G_{e_k}\alpha_k^2}>0.
\]
It diverges when any tolerance tends to zero (and for $f(\alpha):=\sum_{k=1}^mG_{e_k}^{-1}\log(1/\alpha_k)$ and any $M_0\in\mathbb R$, since $\alpha_j\le\bar\alpha$ lower-bounds each other summand by $G_{e_j}^{-1}\log(1/\bar\alpha)$, any nonempty sublevel set $\{f\le M_0\}$ is closed in the compact box $\prod_{k=1}^m[\underline\alpha_k,\bar\alpha]\subset(0,\bar\alpha]^m$ with $\underline\alpha_k:=\exp\{-G_{e_k}(M_0-\sum_{j\ne k}G_{e_j}^{-1}\log(1/\bar\alpha))\}>0$), so an optimum is attained with all tolerances positive. Positive budgets admit strictly feasible tolerances by taking them sufficiently small. The convex Lagrange-multiplier conditions are therefore necessary and sufficient~\citep{vandenberghe2004convex}.

If $\varepsilon\ge\bar\alpha\sum_k\Delta_{e_k}$, every tolerance can equal its cap, and monotonicity makes this optimal. Otherwise the direct-error budget is active (so $\{k:\Delta_{e_k}>0\}$ is nonempty). Introduce its multiplier $\xi>0$ and minimize the Lagrangian separately over $0<\alpha_k\le\bar\alpha$. The part depending on call $k$ is
\[
\phi_k(\alpha_k):={}\frac{\log(1/\alpha_k)}{G_{e_k}}+\xi\Delta_{e_k}\alpha_k.
\]
Its derivative is $\phi_k'(\alpha_k)={}-1/(G_{e_k}\alpha_k)+\xi\Delta_{e_k}$. When $\Delta_{e_k}=0$, $\phi_k'(\alpha_k)<0$ on $(0,\bar\alpha]$, so $\alpha_k=\bar\alpha=\min\{\bar\alpha,1/0\}$. When $\Delta_{e_k}>0$, the unconstrained minimizer is $\alpha_k^{\mathrm{unc}}={}1/(\xi\Delta_{e_k}G_{e_k})$; since $\phi_k$ is strictly decreasing on $(0,\alpha_k^{\mathrm{unc}}]$ and strictly increasing on $[\alpha_k^{\mathrm{unc}},\infty)$, imposing the upper cap gives
\[
\alpha_k=\min\!\left\{\bar\alpha,\frac{1}{\xi\Delta_{e_k}G_{e_k}}\right\}
\]
(equivalently, the box-constraint KKT multiplier for $\alpha_k-\bar\alpha\le0$ is $\eta_k=\max\{0,(G_{e_k}\bar\alpha)^{-1}-\xi\Delta_{e_k}\}\ge0$).
This proves Eq.~\ref{eq:ranking-allocation} without introducing a separate multiplier for each cap.

To choose $\xi$, sum $\Delta_{e_k}\alpha_k$ over calls. This sum is continuous in $\xi$, equals the capped budget $\bar\alpha\sum_k\Delta_{e_k}>\varepsilon$ for sufficiently small $\xi$ ($0<\xi\le\xi_{\min}:=\min_{k:\Delta_{e_k}>0}(\bar\alpha\Delta_{e_k}G_{e_k})^{-1}$), and strictly decreases once any tolerance leaves its cap ($\xi>\xi_{\min}$). It tends to zero as $\xi\to\infty$ (since $\sum_k\Delta_{e_k}\alpha_k\le\xi^{-1}\sum_{k:\Delta_{e_k}>0}G_{e_k}^{-1}$). There is therefore a unique $\xi >\xi_{\min}$ exhausting an active budget. Strict convexity gives a unique allocation.\\[-0.5em]

\noindent\textbf{Step 2: Add a global comparison-error budget.}
Assign a multiplier $\zeta\ge0$ to $\sum_k\alpha_k\le\delta$. The per-call derivative becomes
\[
-\frac1{G_{e_k}\alpha_k}+\xi\Delta_{e_k}+\zeta.
\]
Minimizing over the same capped domain (with box multiplier $\eta_k=\max\{0,(G_{e_k}\bar\alpha)^{-1}-(\xi\Delta_{e_k}+\zeta)\}\ge0$) gives Eq.~\ref{eq:ranking-certified-allocation}. The multipliers also satisfy
\[
\xi\!\left(\sum_{k=1}^m\Delta_{e_k}\alpha_k-\varepsilon\right)=0,
\qquad
\zeta\!\left(\sum_{k=1}^m\alpha_k-\delta\right)=0.
\]
The same strict-feasibility argument makes these conditions sufficient. The allocation remains unique, although the multipliers need not be. If the denominator $\xi\Delta_{e_k}+\zeta$ vanishes (e.g., $\xi=\zeta=0$, or $\zeta=0$ with $\Delta_{e_k}=0$), the decreasing objective places $\alpha_k$ at its cap; this is also the limiting interpretation of the displayed allocation formula under $1/0=+\infty$.
\end{proof}

\begin{algorithm}[t]
\caption{Ranking with allocated pairwise evaluations}
\label{alg:ranking}
\begin{algorithmic}[1]
\Require Ranker $\mathcal A$, candidate pairs $\mathcal E$, budgets $\varepsilon,\delta$, cap $\bar\alpha$
\State Initialize an empty cache and run $\mathcal A$.
\While{$\mathcal A$ requests a pair $(M_i,M_j)$}
  \State Estimate gaps and rates for pair $(M_i,M_j)$; set tolerances by Eq.~\ref{eq:allocation-policy}. 
  \If{the unordered pair is not cached}
    \State Run \texttt{PairEval}$(M_i,M_j,\alpha_{ij})$ with new judgments; save its winner.
  \EndIf
  \State Return the winner to $\mathcal A$ in the requested orientation.
\EndWhile
\State \Return the ordering produced by $\mathcal A$.
\end{algorithmic}
\label{alg:ranking_algo}
\end{algorithm}

\noindent\textbf{What the optimization does and does not establish.}
Additive expected costs independent of the tolerances do not change this optimization for a fixed call list. The pairwise analysis motivates its leading logarithmic objective; the rule is not claimed to minimize the exact stopping cost or a high-probability quantile. For an adaptive sorter, the relevant spending constraint must hold along every possible execution, as in Theorem~\ref{thm:ranking-guarantees}. In particular, a tolerance assigned once to a pair cannot be spent again without accounting for each fresh call.

\clearpage
\section{Prompts}\label{appdx:prompt}

In this section, we present the prompt used for prediction using LLM estimator. We construct the \texttt{History} section by randomly selecting 10 questions along with the label indicating the winning model, among the observed questions. We provide all the unlabeled questions in the \texttt{Questions} section. Each  question in both sections is truncated to 500 characters.

\begin{user_example}[frametitle={LLM Estimator Prompt}]
You are helping evaluate a pairwise comparison of two AI models (Model A vs. Model B).\\

Your task: from the list of remaining questions below, identify the top \{\texttt{K}\} questions that are most likely to \textbf{discriminate} between the two models. For each question, estimate ``$p_{\mathrm{win}}$'': the probability that \textbf{Model A} produces a clearly better response than \textbf{Model B}.\\

- $p_{\mathrm{win}}$ $\approx$ 0.0 $\rightarrow$ you are confident \textbf{Model B} will clearly win

- $p_{\mathrm{win}}$ $\approx$ 0.5 $\rightarrow$ you expect a \textbf{tie} (both models perform similarly)

- $p_{\mathrm{win}}$ $\approx$ 1.0 $\rightarrow$ you are confident \textbf{Model A} will clearly win\\

Prefer questions where p\_win is far from 0.5 (strongly discriminative in either direction).

\{\texttt{History}\} \\

Here are the remaining questions (indexed 0 to \{\texttt{N-1}\}):

\{\texttt{Questions}\}\\

Respond with a JSON object containing a ``candidates'' list. Each item in the list must have:\\

  - ``index'': integer index of the question
  
  - ``$p_{\mathrm{win}}$'': float between 0 and 1 (probability that Model A wins)\\

Return exactly \{$K$\} candidates. Do not include any other text.
\end{user_example}

\clearpage
\section{Dataset Visualization}

\begin{figure}[h!]
    \centering
    \includegraphics[width=0.9\linewidth, keepaspectratio]{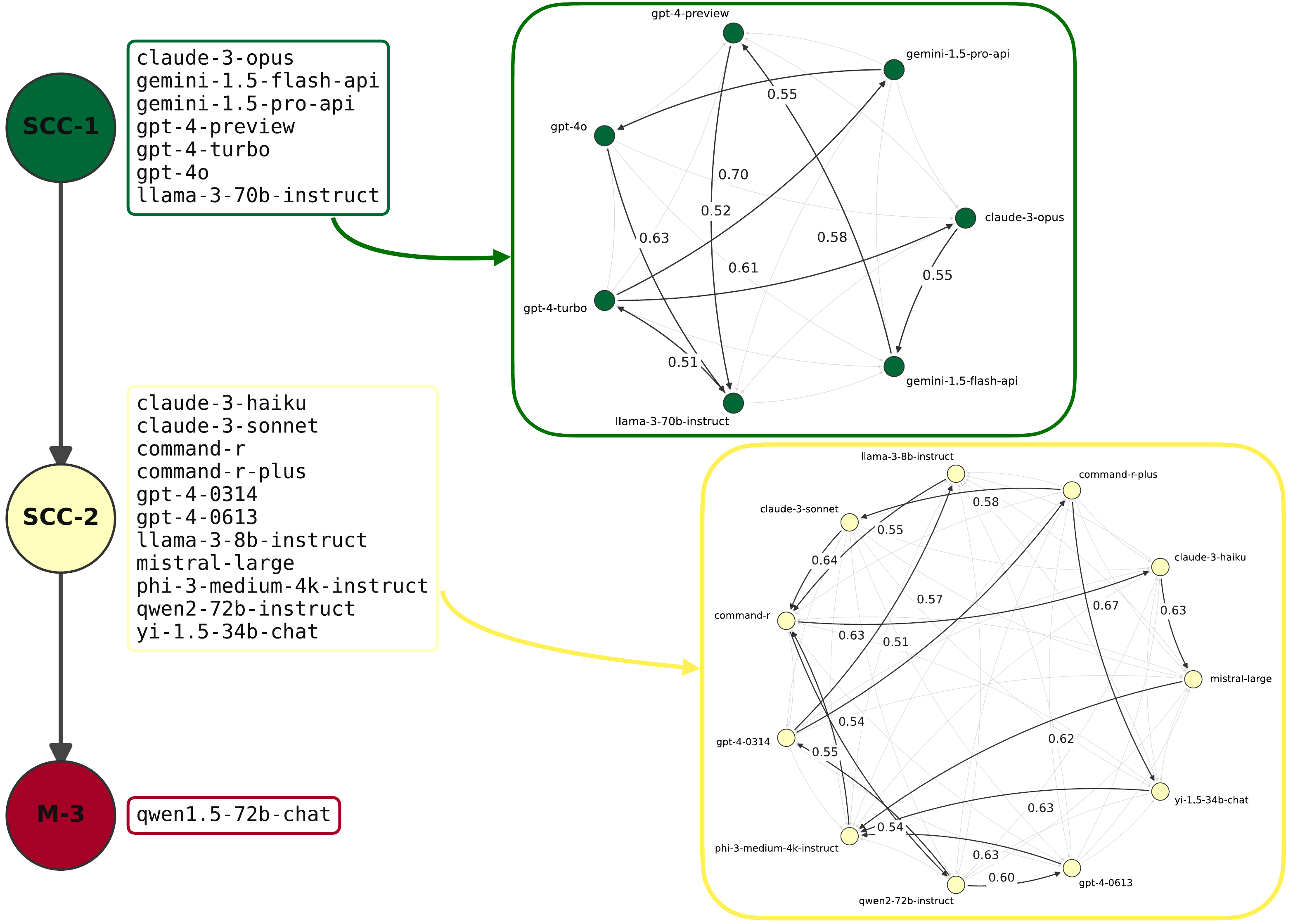}
    \includegraphics[width=0.95\linewidth, keepaspectratio]{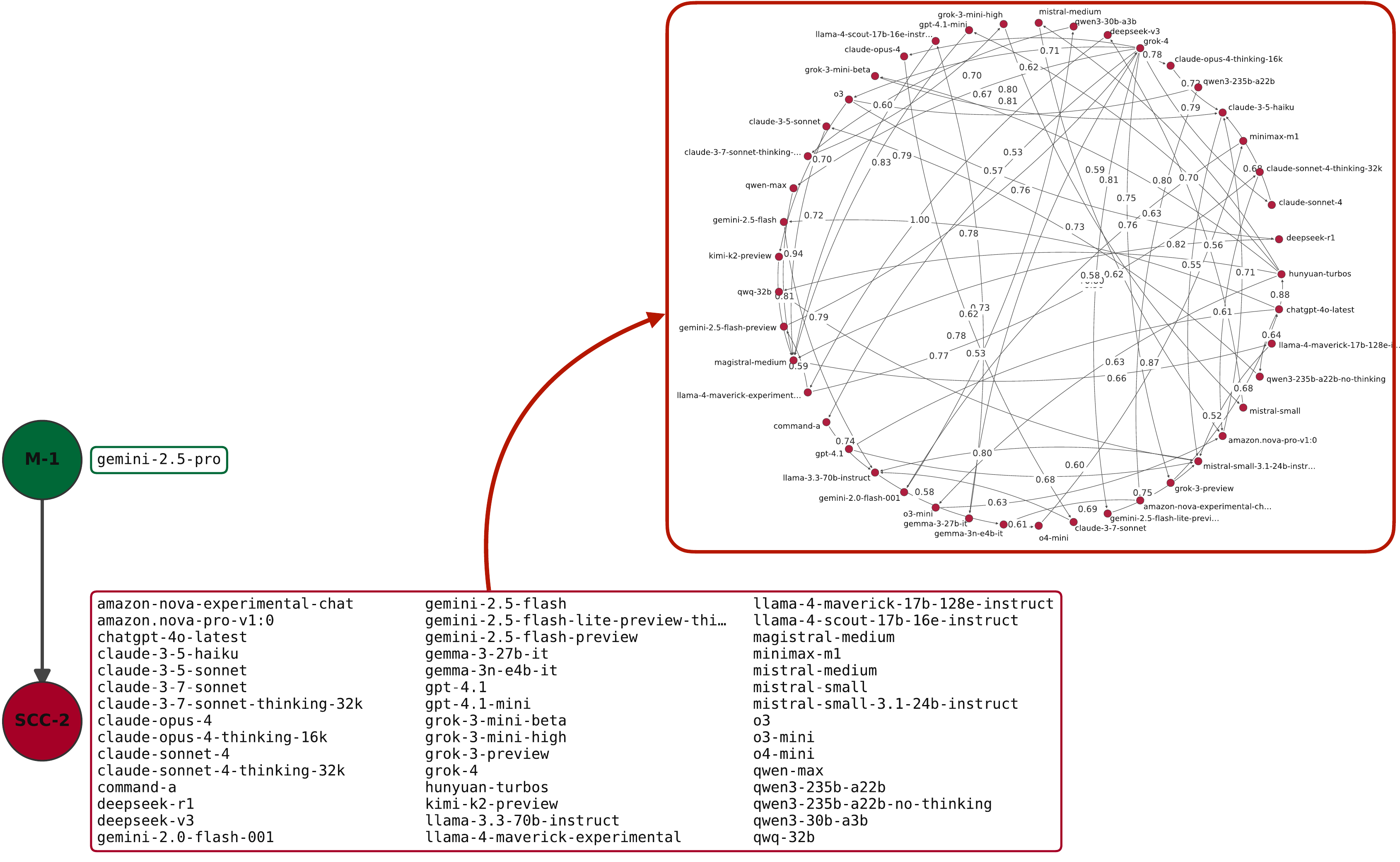}
    \caption{Directed preference graph for the PPE (top) and LM Arena (bottom) datasets using LLM as nodes and human preference as edge direction. The resultant graph shows that there are many strongly connected components and an optimal linear ordering of LLMs doesn't exist. }
    \label{fig:dag1}
\end{figure}
\begin{figure}
    \centering
    \includegraphics[width=0.8\linewidth]{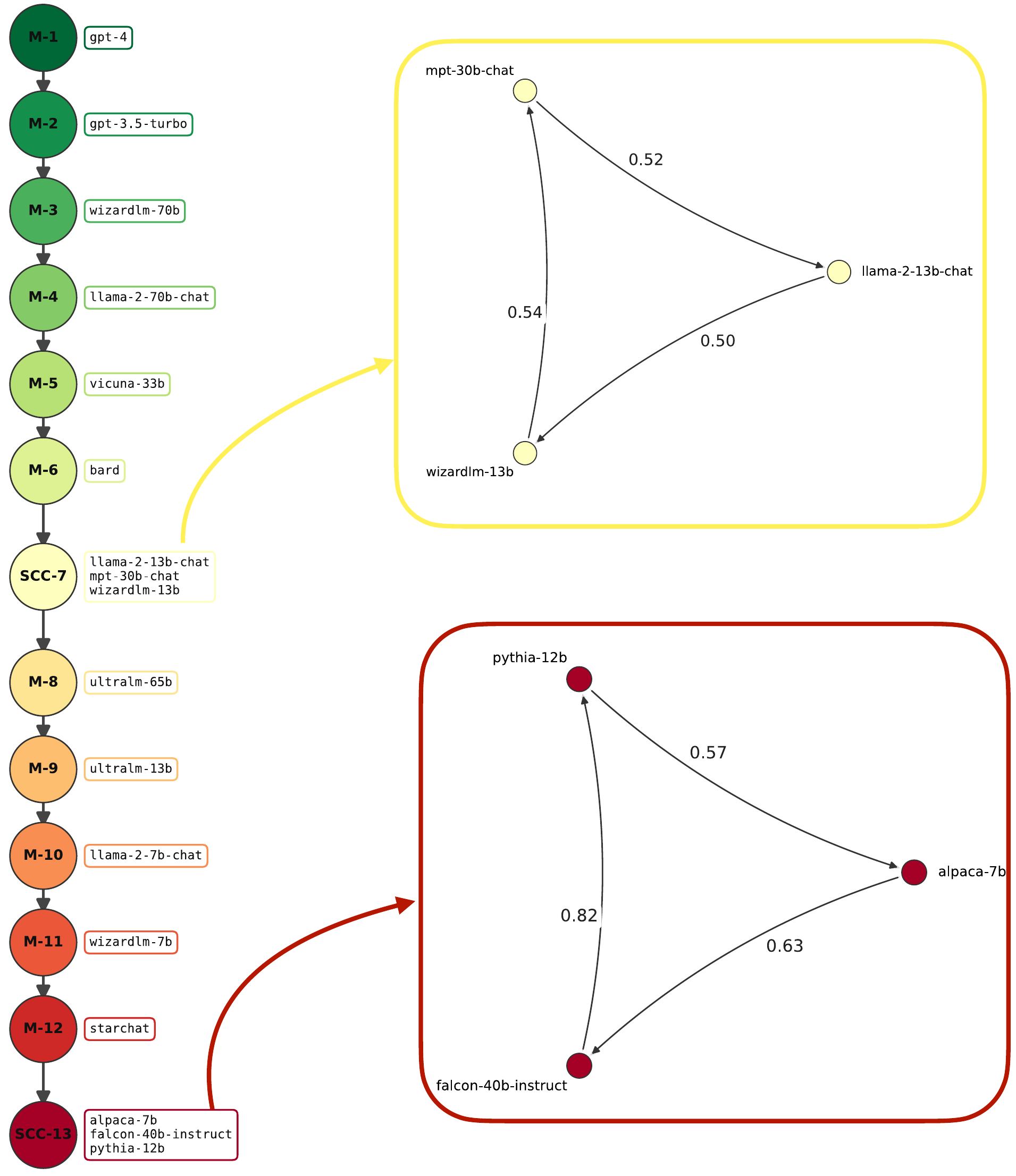}
    \caption{Directed preference graph for the Ultrafeedback dataset. The resultant graph appears much more linear than PPE and LM Arena graphs. This allows the ranking algorithms to converge faster using only 8\% of the total evaluation samples.}
    \label{fig:dag2}
\end{figure}

\newpage
\section{Experiments}

\subsection{Implementation Details}
\label{sec:impl}

In this section, we provide the implementation details for different baselines and variants of our proposed algorithm reported in the paper. For the quick sort (doubling $\alpha$) experiments, we start with $\alpha=0.01$. We use $\bar \alpha=0.99$. In adaptive $\alpha$ methods, we use 20 samples for the pilot phase to estimate $\hat \Delta_{ij}$ and $\hat G_{ij}$ and these samples are accounted for in the overall sample used. For baselines that use $\epsilon$-greedy sampling, we set $\epsilon=0.5$. For LLM-based preference probability prediction, we ask the LLM to return $K=10$ candidates and their corresponding $\pi(x)$ scores, we set $\kappa=0.001$ and use uniform probability over the remaining prompts.

\subsection{Additional Experiments}
\label{sec:addl_exps}

\noindent\textbf{Ablations with different LLMs}.  In this experiment, we replace Gemini 3.6 Flash with an open-source LLM, Gemma 4 26B, for prompt selection. As shown in Table~\ref{tab:ranking-results-gemma}, the results are comparable to those in Table~\ref{tab:ranking-results} in both ranking quality and sample usage, with Gemma 4 26B even outperforming Gemini 3.6 Flash on LMArena. This demonstrates that the choice of prompt-selection LLM has minimal impact on ranking performance, and that even smaller open-source models can effectively select prompts for efficient evaluation.\\[-0.5em]

\begin{table}[t!]
\centering
\caption{We report the quality of ranking (agreement with pairwise data) while using an open-source Gemma 4 26B model for prompt selection. We observe
that the overall results are very close to the fraction of samples used when using Gemini-based LLM in Table~\ref{tab:ranking-results}.}
\resizebox{\linewidth}{!}{
\begin{tabular}{lcc cc cc}
\toprule
 & \multicolumn{2}{c}{\textbf{LMArena}} & \multicolumn{2}{c}{\textbf{PPE}} & \multicolumn{2}{c}{\textbf{Ultrafeedback}} \\
 \cmidrule(lr){2-3}\cmidrule(lr){4-5}\cmidrule(lr){6-7}
 \textbf{Method} & Quality $(\uparrow)$ & Samples $(\downarrow)$ & Quality $(\uparrow)$ & Samples $(\downarrow)$ & Quality $(\uparrow)$ & Samples $(\downarrow)$\\
\midrule
\texttt{Active Ranking} \textbf{\textcolor{ar}{[AR]}} & 59.8\footnotesize{\color{gray} $\pm$0.5} & 72.9\% & 57.4\footnotesize{\color{gray} $\pm$0.8} & 51.8\% & 65.4\footnotesize{\color{gray} $\pm$3.3} & 64.8\% \\
\texttt{Active Testing} (w/ QS) \textbf{\textcolor{atq}{[ATQ]}} & 60.6\footnotesize{\color{gray} $\pm$0.5} & 19.0\% & 59.2\footnotesize{\color{gray} $\pm$0.1} & 30.9\% & 75.0\footnotesize{\color{gray} $\pm$0.1} & 19.9\% \\
\texttt{Active Testing} (w/ MS) \textbf{\textcolor{atm}{[ATM]}} & 61.0\footnotesize{\color{gray} $\pm$0.2} & 16.8\% & 59.0\footnotesize{\color{gray} $\pm$0.5} & 29.8\% & \textbf{75.1}\footnotesize{\color{gray} $\pm$0.0} & 19.9\% \\
\texttt{Quick Sort} (const.\ $\alpha$) \textbf{\textcolor{blqs}{[QS]}} & \textbf{61.5}\footnotesize{\color{gray} $\pm$0.3} & 15.5\% & \textbf{59.4}\footnotesize{\color{gray} $\pm$0.3}
& 29.6\% & \textbf{75.1}\footnotesize{\color{gray} $\pm$0.0} & 14.4\% \\
\texttt{Quick Sort} (doubling $\alpha$) \textbf{\textcolor{blqsd}{[QSD]}} & 61.4\footnotesize{\color{gray} $\pm$0.3} & 15.1\% & 58.8\footnotesize{\color{gray} $\pm$0.4} & 32.7\% & 74.8\footnotesize{\color{gray} $\pm$0.2} & 16.2\% \\
\texttt{Merge Sort} (const.\ $\alpha$) \textbf{\textcolor{blms}{[MS]}} & 61.4\footnotesize{\color{gray} $\pm$0.4} & 14.5\% & 59.2\footnotesize{\color{gray} $\pm$0.4} & 29.1\% &
\textbf{75.1}\footnotesize{\color{gray} $\pm$0.0} & 13.2\% \\
\texttt{Quick Sort} (adaptive $\alpha$) \textbf{\textcolor{blqsa}{[QSA]}} (Ours) & 61.2\footnotesize{\color{gray} $\pm$0.2} & 10.8\% & 58.2\footnotesize{\color{gray} $\pm$0.8} &
\textbf{21.9\%} & 74.8\footnotesize{\color{gray} $\pm$0.1} & 9.5\% \\
\texttt{Merge Sort} (adaptive $\alpha$) \textbf{\textcolor{blmsa}{[MSA]}} (Ours) & 61.4\footnotesize{\color{gray} $\pm$0.1} & ~~\textbf{9.4\%} & 58.6\footnotesize{\color{gray} $\pm$0.3} &
22.4\% & 74.5\footnotesize{\color{gray} $\pm$0.3} & \textbf{9.1\%} \\
\midrule
\textcolor{gray}{\texttt{Expected (MC)} \textbf{\textcolor{blex}{[EX]}}} & \textcolor{gray}{50.0\footnotesize{ $\pm$0.0}} & \textcolor{gray}{-} &
\textcolor{gray}{50.0\footnotesize{ $\pm$0.0}} & \textcolor{gray}{-} & \textcolor{gray}{50.1\footnotesize{ $\pm$0.1}} & \textcolor{gray}{-} \\
\textcolor{gray}{\texttt{Bradley-Terry} \textbf{\textcolor{blbt}{[BT]}}} & \textcolor{gray}{62.6\footnotesize{ $\pm$0.0}} & \textcolor{gray}{100\%} & \textcolor{gray}{59.8\footnotesize{ $\pm$0.0}} & \textcolor{gray}{100\%} & \textcolor{gray}{75.0\footnotesize{ $\pm$0.0}} & \textcolor{gray}{100\%} \\
\textcolor{gray}{\texttt{HodgeRank} \textbf{\textcolor{blhg}{[HG]}}} & \textcolor{gray}{62.5\footnotesize{ $\pm$0.0}} & \textcolor{gray}{100\%} & \textcolor{gray}{59.8\footnotesize{
$\pm$0.0}} & \textcolor{gray}{100\%} & \textcolor{gray}{75.0\footnotesize{ $\pm$0.0}} & \textcolor{gray}{100\%} \\
\textcolor{gray}{\texttt{Oracle} \textbf{\textcolor{bloe}{[OE]}}} & \textcolor{gray}{63.6\footnotesize{ $\pm$0.0}} & \textcolor{gray}{100\%} & \textcolor{gray}{60.8\footnotesize{
$\pm$0.0}} & \textcolor{gray}{100\%} & \textcolor{gray}{75.2\footnotesize{ $\pm$0.0}} & \textcolor{gray}{100\%} \\
\bottomrule
\end{tabular}
}
\label{tab:ranking-results-gemma}
\end{table}

\begin{figure}
    \centering
    \includegraphics[width=0.95\linewidth]{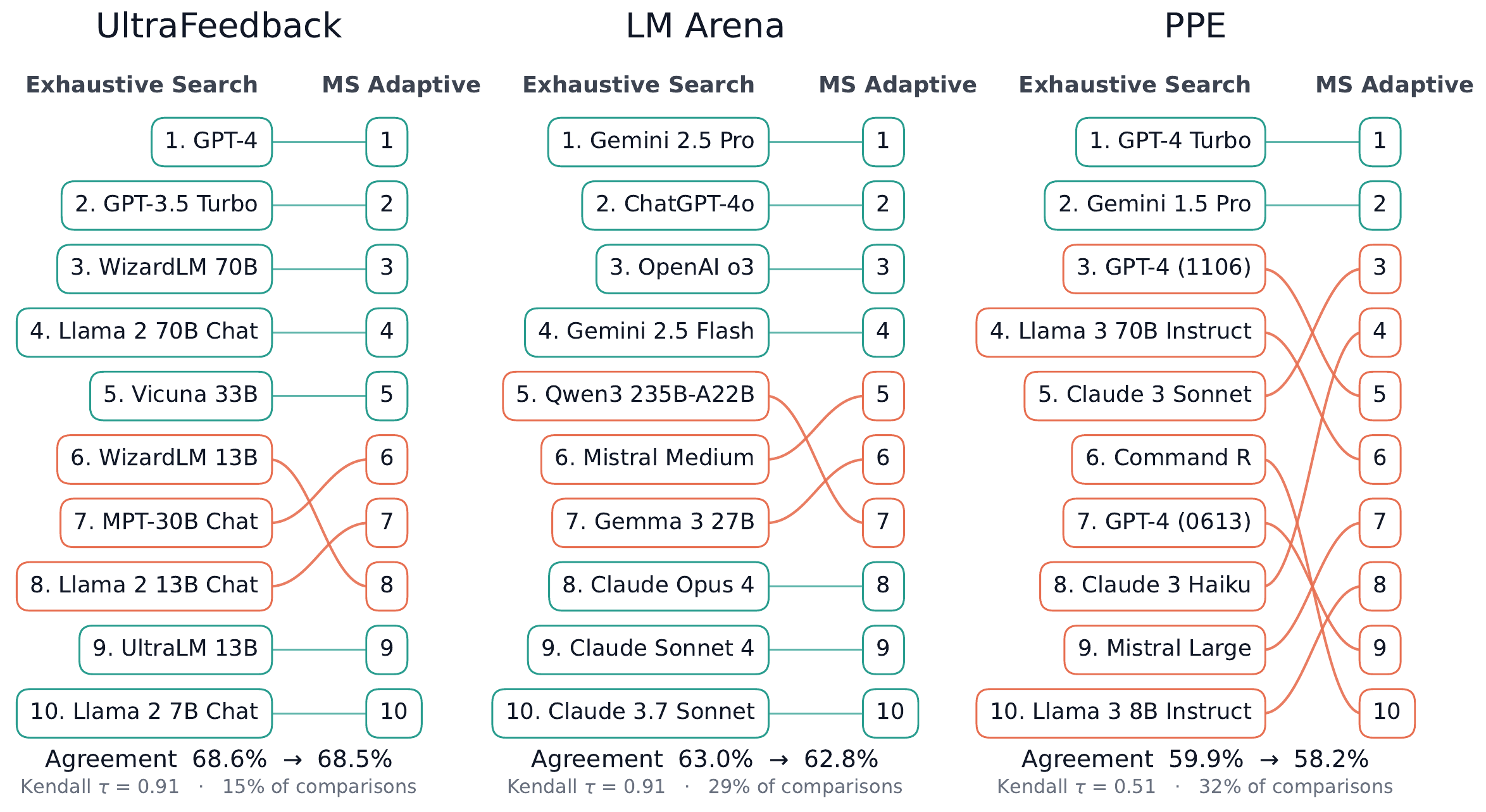}
    \caption{Visualization of the ranking obtained by exhaustive search and adaptive merge sort (MS adaptive) across Ultrafeedback, LM Arena, and PPE datasets.}
    \label{fig:ranking-viz}
\end{figure}

\noindent\textbf{Ranking Visualization}. In this experiment, we select the top 10 models (to keep computation tractable) from each dataset and perform exhaustive search over 10! permutations to find the linear ordering. In Figure~\ref{fig:ranking-viz}, we compare the ranking produced by our method (adaptive merge sort) with the ranking obtained from exhaustive search. On Ultrafeedback and LMArena datasets, we observe that the ranking generated closely matches the one generated by exhaustive search achieving a Kendall's $\tau=0.91$. On Ultrafeedback, we observe that the errors concentrate around ranks 6-8, which is expected as a transitive preference among these models is not supported by the data as shown in Figure~\ref{fig:dag2}.

Among the three datasets, PPE is where our ranking deviates most from the exhaustive optimum (Kendall $\tau$ = 0.51).  PPE also has the fewest annotations per model pair ($\sim$ 78), which suggests the deviations are likely due to the limited number of samples available rather than the ranking algorithm itself. To investigate this hypothesis, we tested every pair against the null hypothesis that its win rate is 0.5, using an exact two-sided binomial test. 
Only 22 out of 45 model pairs (for 10 models) reject the null at $\alpha = 0.05$. Since a ranking outputs a linear ordering, the remaining 23 pairs must still be ordered, with no evidence favouring either direction. Our method (adaptive merge sort) produces a linear ordering of 10 models (shown in Figure~\ref{fig:ranking-viz} (right)), which implies a pairwise decision for 45 possible pairs. We evaluate these pairwise decisions against the ground truth in PPE, and find 10 wrong decisions. However, 9 out of the 10 are among those that even the offline fixed budget test failed to distinguish the winner. Overall, this shows that the apparent deviation from the ranking derived from exhaustive search is an artifact of the limited data available in the PPE dataset and not a limitation of the proposed  algorithm.\\[-0.5em]

\noindent\textbf{Synthetic Results}.  We evaluate our approach in a synthetic setting to exhaustively simulate performance across different win rate gaps ($\Delta$). For a given gap, we generate preference probabilities
from a normal distribution $\pi(x) \sim \mathcal{N}(\mu, \sigma)$, where $\mu = (1+\Delta)/2$, for 10K samples. Using these probabilities, we sample pairwise evaluation labels $y \sim
\text{Bern}(\pi(x))$ and run our proposed e-process-based pairwise evaluation with both random and active sampling, recording the stopping time (number of samples required to determine
a winner). Across 50 trials per gap, active sampling uses $\sim$27\% fewer samples compared to random sampling, with significant gains especially when win rate gap is small where distinguishing
between models is hardest. These results showcase that active sampling provides the greatest benefit precisely in the regime where sample efficiency matters most.\\[-0.5em]

\begin{table}[t!]
\centering
\caption{Quality of ranking (agreement with pairwise data) and fraction of total samples used, with uniform random vs.\ LLM-guided sampling inside each pairwise e-process, with
standard error over 3 runs. A higher quality score and lower fraction of samples used is better.}
\label{tab:ranking-abl}
\resizebox{\linewidth}{!}{
\begin{tabular}{ll cc cc cc}
\toprule
 & & \multicolumn{2}{c}{\textbf{LMArena}} & \multicolumn{2}{c}{\textbf{PPE}} & \multicolumn{2}{c}{\textbf{Ultrafeedback}} \\
 \cmidrule(lr){3-4}\cmidrule(lr){5-6}\cmidrule(lr){7-8}
 \textbf{Method} & \textbf{Sampling} & Quality $(\uparrow)$ & Samples $(\downarrow)$ & Quality $(\uparrow)$ & Samples $(\downarrow)$ & Quality $(\uparrow)$ & Samples $(\downarrow)$\\
\midrule
\multirow{2}{*}{\texttt{Quick Sort} (const.\ $\alpha$) \textbf{\textcolor{blqs}{[QS]}}}
 & Random & 62.0\footnotesize{\color{gray} $\pm$0.3} & 15.6\%\footnotesize{\color{gray} $\pm$0.6} & 58.7\footnotesize{\color{gray} $\pm$0.5} & 28.4\%\footnotesize{\color{gray} $\pm$0.6} & \textbf{75.0}\footnotesize{\color{gray} $\pm$0.0} & 14.5\%\footnotesize{\color{gray} $\pm$0.8} \\
 & LLM & 61.9\footnotesize{\color{gray} $\pm$0.1} & 15.0\%\footnotesize{\color{gray} $\pm$0.7} & 58.4\footnotesize{\color{gray} $\pm$0.3} & 29.7\%\footnotesize{\color{gray} $\pm$0.1}
& 74.9\footnotesize{\color{gray} $\pm$0.1} & 11.5\%\footnotesize{\color{gray} $\pm$0.8} \\
\midrule
\multirow{2}{*}{\texttt{Quick Sort} (doubling $\alpha$) \textbf{\textcolor{blqsd}{[QSD]}}}
 & Random & 62.0\footnotesize{\color{gray} $\pm$0.3} & 13.6\%\footnotesize{\color{gray} $\pm$1.3} & 57.7\footnotesize{\color{gray} $\pm$0.4} & 30.7\%\footnotesize{\color{gray} $\pm$1.7} & 74.7\footnotesize{\color{gray} $\pm$0.3} & 14.5\%\footnotesize{\color{gray} $\pm$0.6} \\
 & LLM & 61.3\footnotesize{\color{gray} $\pm$0.1} & 15.9\%\footnotesize{\color{gray} $\pm$0.4} & 58.4\footnotesize{\color{gray} $\pm$0.2} & 31.3\%\footnotesize{\color{gray} $\pm$1.1}
& \textbf{75.0}\footnotesize{\color{gray} $\pm$0.1} & 15.4\%\footnotesize{\color{gray} $\pm$0.2} \\
\midrule
\multirow{2}{*}{\texttt{Merge Sort} (const.\ $\alpha$) \textbf{\textcolor{blms}{[MS]}}}
 & Random & \textbf{62.1}\footnotesize{\color{gray} $\pm$0.1} & 15.2\%\footnotesize{\color{gray} $\pm$0.6} & 57.9\footnotesize{\color{gray} $\pm$0.5} & 28.5\%\footnotesize{\color{gray} $\pm$1.5} & 74.9\footnotesize{\color{gray} $\pm$0.1} & 14.6\%\footnotesize{\color{gray} $\pm$0.6} \\
 & LLM & 60.8\footnotesize{\color{gray} $\pm$0.2} & 14.4\%\footnotesize{\color{gray} $\pm$0.5} & 59.2\footnotesize{\color{gray} $\pm$0.2} & 28.1\%\footnotesize{\color{gray} $\pm$0.8}
& \textbf{75.0}\footnotesize{\color{gray} $\pm$0.1} & 12.2\%\footnotesize{\color{gray} $\pm$0.8} \\
\midrule
\multirow{2}{*}{\texttt{Quick Sort} (adaptive $\alpha$) \textbf{\textcolor{blqsa}{[QSA]}}}
 & Random & 61.3\footnotesize{\color{gray} $\pm$0.2} & 13.4\%\footnotesize{\color{gray} $\pm$0.8} & 58.6\footnotesize{\color{gray} $\pm$0.4} & 27.1\%\footnotesize{\color{gray} $\pm$0.5} & 74.8\footnotesize{\color{gray} $\pm$0.2} & 11.9\%\footnotesize{\color{gray} $\pm$0.7} \\
 & LLM & 61.0\footnotesize{\color{gray} $\pm$0.3} & \textbf{9.7\%}\footnotesize{\color{gray} $\pm$0.8} & \textbf{59.3}\footnotesize{\color{gray} $\pm$0.1} & 21.8\%\footnotesize{\color{gray} $\pm$1.3} & 74.7\footnotesize{\color{gray} $\pm$0.2} & \textbf{6.8\%}\footnotesize{\color{gray} $\pm$0.5} \\
\midrule
\multirow{2}{*}{\texttt{Merge Sort} (adaptive $\alpha$) \textbf{\textcolor{blmsa}{[MSA]}}}
 & Random & 60.6\footnotesize{\color{gray} $\pm$0.3} & 12.9\%\footnotesize{\color{gray} $\pm$0.3} & 58.8\footnotesize{\color{gray} $\pm$0.3} & 25.9\%\footnotesize{\color{gray} $\pm$0.6} & 74.8\footnotesize{\color{gray} $\pm$0.1} & 11.7\%\footnotesize{\color{gray} $\pm$0.6} \\
 & LLM & 60.1\footnotesize{\color{gray} $\pm$0.5} & \textbf{9.7\%}\footnotesize{\color{gray} $\pm$0.8} & 58.3\footnotesize{\color{gray} $\pm$0.4} & \textbf{21.7\%}\footnotesize{\color{gray} $\pm$1.3} & 74.5\footnotesize{\color{gray} $\pm$0.3} & 7.6\%\footnotesize{\color{gray} $\pm$0.9} \\
\bottomrule
\end{tabular}
}
\end{table}

\noindent \textbf{Sampling Ablations}. In this experiment, we perform ablations of all sorting based ranking methods by using either random or LLM-based sampling. We report the results in Table~\ref{tab:ranking-abl}. We observe that LLM-based sampling provides gains in the number of samples in the majority of settings, with the gains being most pronounced in adaptive $\alpha$ methods. We hypothesize that these gains are more prominent because for higher $\alpha$ we need fewer samples to conclude hypothesis testing and the LLM-based prediction is effective in identifying discriminative examples. We also observe that adaptive $\alpha$ methods showcase sample count gains irrespective of the underlying sampling method. This implies that the gains are an artifact of the derived error allocation strategy (Section~\ref{sec:ranking}) that are further amplified using the LLM-based sampling method.\\

\begin{table}[t]
\centering
\caption{Detailed statistics of the LLM expenses used in prompt selection using Gemini 3.6 Flash.}
\label{tab:llm_cost}
\resizebox{0.8\linewidth}{!}{
\begin{tabular}{lrrr}
\toprule
 & LMArena & PPE & UltraFeedback \\
\midrule
Avg.\ input tokens per call & $\sim$4.7k & $\sim$4.9k & $\sim$43k \\
Cost per call & $\sim$\$0.008 & $\sim$\$0.008 & $\sim$\$0.065 \\
LLM calls per ranking & $\sim$9.1k & $\sim$2.6k & $\sim$4.4k \\
LLM cost per ranking & $\sim$\$70 & $\sim$\$22 & $\sim$\$280 \\
Human labels saved (w.r.t.\ uniform sampling) & $\sim$3,200 & $\sim$580 & $\sim$2,800 \\
LLM cost per human label saved & $\sim$\$0.02 & $\sim$\$0.04 & $\sim$\$0.10 \\
\bottomrule
\end{tabular}}
\end{table}

\noindent \textbf{Cost Analysis}. In our framework, we use a single LLM call per human label collected. The LLM call selects the prompt, then the models generate their corresponding responses, which are annotated by the human. In Table~\ref{tab:llm_cost}, we report the detailed token and cost statistics for Gemini 3.6 Flash used in our experiments. We observe that LLM cost to save a human label is only \$0.02-\$0.1, which is very small compared to human costs required to judge complex LLM trajectories.\\

\begin{table}[t!]
\centering
\caption{We compare our adaptive error allocation with constant-$\alpha$ baselines whose $\alpha$ is matched to the average $\alpha$ of our adaptive allocation.}
\resizebox{\linewidth}{!}{
\begin{tabular}{lcc cc cc}
\toprule
 & \multicolumn{2}{c}{\textbf{LMArena}} & \multicolumn{2}{c}{\textbf{PPE}} & \multicolumn{2}{c}{\textbf{Ultrafeedback}} \\
 \cmidrule(lr){2-3}\cmidrule(lr){4-5}\cmidrule(lr){6-7}
 \textbf{Method} & Quality $(\uparrow)$ & Samples $(\downarrow)$ & Quality $(\uparrow)$ & Samples $(\downarrow)$ & Quality $(\uparrow)$ & Samples $(\downarrow)$\\
\midrule
\texttt{Quick Sort} (matched const.\ $\alpha$) \textbf{\textcolor{blqs}{[QS]}} & \textbf{61.0}\footnotesize{\color{gray} $\pm$0.6} & 11.1\%\footnotesize{\color{gray} $\pm$0.4} & 58.6\footnotesize{\color{gray} $\pm$0.3} & 22.3\%\footnotesize{\color{gray} $\pm$1.7} & 75.0\footnotesize{\color{gray} $\pm$0.1} & 11.9\%\footnotesize{\color{gray} $\pm$0.6} \\
\texttt{Quick Sort} (adaptive $\alpha$) \textbf{\textcolor{blqsa}{[QSA]}} (Ours) & \textbf{61.0}\footnotesize{\color{gray} $\pm$0.3} & \textbf{9.7\%}\footnotesize{\color{gray} $\pm$0.8} & \textbf{59.3}\footnotesize{\color{gray} $\pm$0.1} & \textbf{21.8\%}\footnotesize{\color{gray} $\pm$1.3} & 74.7\footnotesize{\color{gray} $\pm$0.2} & \textbf{6.8\%}\footnotesize{\color{gray} $\pm$0.5} \\
\midrule
\texttt{Merge Sort} (matched const.\ $\alpha$) \textbf{\textcolor{blms}{[MS]}} & \textbf{61.0}\footnotesize{\color{gray} $\pm$0.3} & 11.1\%\footnotesize{\color{gray} $\pm$0.3} & 59.1\footnotesize{\color{gray} $\pm$0.2} & 22.3\%\footnotesize{\color{gray} $\pm$1.9} & \textbf{75.1}\footnotesize{\color{gray} $\pm$0.0} & 12.6\%\footnotesize{\color{gray} $\pm$0.3} \\
\texttt{Merge Sort} (adaptive $\alpha$) \textbf{\textcolor{blmsa}{[MSA]}} (Ours) & 60.1\footnotesize{\color{gray} $\pm$0.5} & \textbf{9.7\%}\footnotesize{\color{gray} $\pm$0.8} & 58.3\footnotesize{\color{gray} $\pm$0.4} & ~\textbf{21.7\%}\footnotesize{\color{gray} $\pm$1.3} & 74.5\footnotesize{\color{gray} $\pm$0.3} & \textbf{7.6\%}\footnotesize{\color{gray} $\pm$0.9} \\
\bottomrule
\end{tabular}}
\label{tab:matched-alpha}
\end{table}

\noindent \textbf{Matched Error Allocation}. In this experiment, we conduct experiments to investigate the performance of sorting-based e-process algorithm with constant-$\alpha$ that matches the average $\alpha$ produced by the adaptive allocation. We report the results in Table~\ref{tab:matched-alpha}. Even with $\alpha$ matched, our methods still use ~13\% fewer labels on LMArena and ~40\% fewer on UltraFeedback, while being comparable on PPE. The matched constant $\alpha$ is also an oracle hyperparameter choice, it is only known after running the adaptive method. Our allocation strategy instead sets per-pair error tolerances in a principled way without tuning $\alpha$ using additional heuristics.\\

\begin{table}[t!]
\centering
\caption{Fraction of LLM pairs resolved by SERPANT when run with the sample budget required by quick sort adaptive $\alpha$ (QSA) to produce a full ranking.}
\label{tab:serpant}
\resizebox{0.5\linewidth}{!}{
\begin{tabular}{lrr}
\toprule
\textbf{Dataset} & Samples ($\downarrow$)  & Pairs resolved $(\uparrow)$ \\
\midrule
{LMArena}       & 9.7\%  & 0.0\%\footnotesize{\color{gray} $\pm$0.0} \\
{PPE}           & 21.8\% & 0.6\%\footnotesize{\color{gray} $\pm$0.3} \\
{Ultrafeedback} & 6.8\%  & 36.6\%\footnotesize{\color{gray} $\pm$1.1} \\
\bottomrule
\end{tabular}}
\end{table}

\noindent\textbf{Comparison with SERPANT~\citep{gu2026anytimevalid}}. In this experiment, we compare with SERPANT, which also uses e-processes to resolve LLM pairs. It maintains a set of unresolved model pairs and, in each round, labels randomly selected examples for every unresolved pair, updating that pair's e-process. With a finite budget, only some pairs are resolved, so SERPANT returns a partial ordering rather than a full ranking and cannot be directly compared with our method. For a fair comparison, we run SERPANT with the same sample budget our method (QSA) needs to produce a full ranking, and report the fraction of pairs it resolves in Table~\ref{tab:serpant}. At this budget, SERPANT resolves almost no pairs on LMArena and PPE, and only about a third on UltraFeedback. Because it spreads labels uniformly over all $O(n^2)$ model pairs, evidence for each pair builds up slowly. Our method instead evaluates only the $O(n \log n)$ comparisons requested by the sorting algorithm, concentrates labels on one pair at a time, and selects informative prompts, which lets it return a complete ranking within the same budget.\\

\begin{table}[t!]
\centering
\caption{Comparison of sampling distributions when both use the same LLM predictions. Win rate is the fraction of LLM pairs where a method requires fewer labels than uniform sampling.}
\resizebox{0.9\linewidth}{!}{
\begin{tabular}{l cc cc cc}
\toprule
 & \multicolumn{2}{c}{\textbf{LMArena}} & \multicolumn{2}{c}{\textbf{PPE}} & \multicolumn{2}{c}{\textbf{Ultrafeedback}} \\
\cmidrule(lr){2-3}\cmidrule(lr){4-5}\cmidrule(lr){6-7}
\textbf{Method} & Win rate $(\uparrow)$ & Acc. $(\uparrow)$ & Win rate $(\uparrow)$ & Acc. $(\uparrow)$ & Win rate $(\uparrow)$ & Acc. $(\uparrow)$ \\
\midrule
Active Testing (w/ LLM) & 35.37\% & {92.68\%} & 41.03\% & {85.87\%} & {67.80\%} & 95.42\% \\
\texttt{LLM} {\textcolor{bla}{\textbf{[LM]}}} (Ours) & {57.32\%} & 91.71\% & {56.79\%} & 85.65\% & 59.32\% & {97.46\%} \\
\bottomrule
\end{tabular}
}
\label{tab:same-llm}
\end{table}

\noindent \textbf{Prediction Ablations}. In this experiment, we investigate the gains obtained due to the LLM's predictions versus the gains due to our sampling distribution. To achieve this, we equip Active Testing's proposal distribution with the same LLM predictions. In Table~\ref{tab:same-llm}, Active Testing with LLM predictions is substantially less sample-efficient than our method on LMArena and PPE, at comparable accuracy. On UltraFeedback, it is more sample-efficient but loses 2\% accuracy. Since both methods use identical LLM predictions, this experiment shows that the efficiency gains come from our proposed sampling distribution, not the LLM alone.

\end{document}